\pdfoutput=1

\documentclass[10pt,twocolumn,letterpaper]{article}

\usepackage[pagenumbers]{cvpr}

\usepackage[OT1]{fontenc}

\usepackage{caption} 
\DeclareCaptionJustification{overviewfull}{\parfillskip=0pt}
\usepackage{array}
\usepackage{multirow}
\usepackage{colortbl}
\usepackage{placeins}
\usepackage{needspace}
\providecommand{\nolinenumbers}{}
\usepackage{cuted}
\usepackage{acro}
\usepackage{listings}
\usepackage[most]{tcolorbox}

\definecolor{UcCodeInk}{HTML}{151515}
\definecolor{UcCodeKeyword}{HTML}{2F6593}
\definecolor{UcCodeComment}{HTML}{367F62}
\definecolor{UcCodeNumber}{HTML}{627D94}
\definecolor{UcCodeString}{HTML}{633D25}
\lstdefinestyle{unicraft}{
    language=Python,
    basicstyle=\fontfamily{zi4}\mdseries\upshape\fontsize{8}{9}\selectfont\color{UcCodeInk},
    keywordstyle=\color{UcCodeKeyword}\bfseries,
    commentstyle=\color{UcCodeComment},
    stringstyle=\color{UcCodeString},
    numbers=left,numberstyle=\fontfamily{zi4}\fontsize{6}{7}\selectfont\color{UcCodeNumber},
    numbersep=5pt,stepnumber=1,
    columns=fullflexible,keepspaces=true,showstringspaces=false,
    breaklines=true,breakatwhitespace=false,tabsize=4,
    aboveskip=0pt,belowskip=0pt
}

\DeclareAcronym{llm}{short = LLM, long = large language model}
\DeclareAcronym{vlm}{short = VLM, long = vision-language model}
\DeclareAcronym{vla}{short = VLA, long = vision-language-action}
\DeclareAcronym{cap}{short = CaP, long = Code as Policies}
\DeclareAcronym{eef}{short = EEF, long = end-effector}
\newcommand{\red}[1]{#1}

\newcommand{\method}{\textsc{UniCraft}\xspace}

\newcommand{\ModelGPTAstra}{GPT-6 Astra}
\newcommand{\ModelGPTSol}{GPT-6.1 Sol}
\newcommand{\ModelGPTTerra}{GPT-6.1 Terra}
\newcommand{\ModelGeminiFlash}{Gemini 3.8 Flash}
\newcommand{\ModelDeepSeekFlash}{DeepSeek-V4.1-Flash}
\newcommand{\ModelMiniMaxFlash}{MiniMax M3.1 Flash}

\DeclareMathAlphabet{\mathprogram}{OT1}{LinuxBiolinumT-LF}{m}{n}
\newcommand{\FunctionName}[1]{\ensuremath{\mathprogram{#1}}}

\definecolor{BudgetTrain}{HTML}{EDF2F6}
\definecolor{BudgetInk}{HTML}{293845}

\newcommand{\ExperimentBenchmarkCount}{\red{5}}
\newcommand{\ExperimentBackboneCount}{\red{3}}

\newcommand{\TrainingTaskCount}{\red{180}}
\newcommand{\TrainingPairCount}{\red{300}}
\newcommand{\TrainingEmbodimentCount}{\red{6}}
\newcommand{\TrainingEpochCount}{\red{3}}
\newcommand{\TrainingCaseCount}{\red{900}}
\newcommand{\TrainingRolloutCount}{\red{2{,}700}}
\newcommand{\TrainingBatchCount}{\red{6}}
\newcommand{\TrainingBatchSize}{\red{50}}
\newcommand{\TrainingRolloutsPerBatch}{\red{150}}
\newcommand{\TrainingRepeatCount}{\red{3}}
\newcommand{\TestRepeatCount}{\red{5}}
\newcommand{\TrainingHorizonMultiple}{\red{2}}
\newcommand{\LiberoTaskCount}{\red{90}}
\newcommand{\RoboTwinTaskCount}{\red{40}}
\newcommand{\RoboTwinTotalTaskCount}{\red{50}}
\newcommand{\RoboTwinBodyCount}{\red{4}}
\newcommand{\RoboCasaTaskCount}{\red{50}}

\definecolor{linkblue}{rgb}{0.21,0.49,0.74}
\usepackage[pagebackref,breaklinks,colorlinks,allcolors=linkblue]{hyperref}
\hypersetup{
  pdftitle={Alignment Drives Reuse: Learning Generalizable Robot-Use Agents from Heterogeneous Experience},
  pdfauthor={Ziyue Wang; Yaopeng Yang; Liyi Zhou}
}

\title{Alignment Drives Reuse:\\ Learning Generalizable Robot-Use Agents from Heterogeneous Experience}

\author{%
Ziyue Wang$^{1}$\qquad Yaopeng Yang$^{2}$\qquad Liyi Zhou$^{1}$\\
$^{1}$The University of Sydney\qquad $^{2}$Nanjing University
}

\begin{document}
\maketitle
\begingroup
\renewcommand{\thefootnote}{}
\makeatletter
\renewcommand{\@makefntext}[1]{\noindent#1}
\makeatother
\footnotetext{\itshape\textbf{Author contributions:}
Ziyue Wang led the research and carried out the core implementation and experiments.
Yaopeng Yang refined the implementation and conducted the DeepSeek experiments.
Liyi Zhou helped refine the research ideas and provided research guidance.
All authors contributed to writing the paper.}
\endgroup
\raggedbottom
\begin{figure*}[t]
    \captionsetup{hypcap=false,font=small,labelfont=normalfont,
      position=bottom,justification=overviewfull}
    \centering
    \includegraphics[page=1,width=\textwidth]{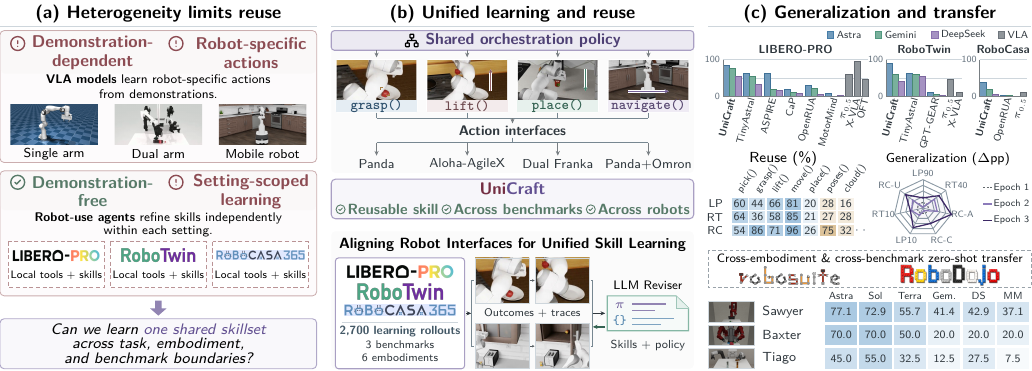}\par
    \caption{\textbf{Alignment drives shared learning and reuse in UniCraft.}
    (a) Robot-specific actions and setting-scoped learning motivate shared
    skills across task, embodiment, and benchmark
    boundaries (\S\ref{sec:introduction}).
    (b) UniCraft aligns representations and execution interfaces
    (\S\ref{sec:alignment}) to jointly learn and refine shared
    skills and their orchestration policy from multi-episode feedback
    (\S\ref{sec:offline-learning}). Training mixes 2,700 rollouts per model
    across 3 benchmarks and 6 embodiments (\S\ref{sec:exp-evolution}).
    The same skills compose across robots via native adapters
    (\S\ref{sec:online-composition}).
    (c) We comprehensively evaluate UniCraft across 5 benchmarks and
    10 embodiments, assessing performance against baselines, skill reuse,
    and zero-shot transfer of the same skills and policy to 4 new robots
    and 2 unseen benchmarks (\S\ref{sec:experiments}).
    Transfer cells show robosuite at 2,000 simulation steps (Gem.: Gemini; DS: DeepSeek;
    MM: MiniMax).}
    \label{fig:overview}
\end{figure*}
\acresetall
\begin{abstract}
\Acp{vlm} demonstrate broad generalization across tasks and domains.
However, comparable generalization in robot manipulation remains challenging.
\textbf{Heterogeneity} in robot embodiments, task formulations, geometric
conventions, and action interfaces fragments interaction experience across
benchmarks and hardware.
We rethink robot skill learning through alignment: \textbf{Can heterogeneous
interactions across benchmarks and embodiments jointly improve one shared
skillset for robot manipulation?}
We propose \method, a framework for learning robot manipulation
skills by aligning heterogeneous settings.
Specifically, \method represents observed scene geometry using consistent
coordinate conventions and aligns sensing and control interfaces across robots.
It organizes tasks from different benchmarks in a canonical format, allowing
interaction feedback to jointly refine shared skills and the orchestration
policy that guides their selection and composition, \textbf{without any}
demonstrations or model weight updates.
We interleave tasks from 3 benchmarks and 6 embodiments, collecting 2,700
learning rollouts per agent model.
During training, \method improves pooled task success on unseen scenarios and
tasks by \mbox{9--28} percentage points on LIBERO-PRO, \mbox{10.8--13.3} on
RoboTwin 2.0, and \mbox{3.1--14.4} on RoboCasa365 (Epochs~\mbox{1--3}).
After training, with the learned skills frozen, \method demonstrates zero-shot
transfer across 4 new robot embodiments and 5 additional agent models on
robosuite and RoboDojo.
\end{abstract}
\acresetall
\section{Introduction}
\label{sec:introduction}

\Acp{llm} and \acp{vlm} have
demonstrated broad generalization across language, image, and video
tasks~\cite{brown2020language,alayrac2022flamingo}.
Natural language and images provide a unified interface for representing
information and specifying tasks across domains.
Aligning heterogeneous data through this shared interface supports training
at scale on diverse sources, allowing models to acquire knowledge that
extends to new concepts, instructions, and reasoning
tasks~\cite{chung2024scaling,grattafiori2024llama3}.
At inference, the same interface enables zero-shot transfer to new tasks without
task-specific examples or parameter updates~\cite{wei2022finetuned,sanh2022multitask,li2023blip2}.

However, comparable generalization in robotics remains challenging.
One of the main reasons is \textbf{heterogeneity} (Figure~\ref{fig:overview}(a)).
Robot embodiments differ in their kinematics, observation
spaces, and action representations; even the same numerical action can
produce different physical effects depending on the robot and its
controller~\cite{openx2023,yuan2026qwenrobotmanip}.
These embodiment differences also shape how interaction experience is
collected and represented, while datasets introduce further differences in
coordinate conventions and annotation practices~\cite{kim2025openvla,yuan2026qwenrobotmanip}.
These differences fragment experience across platforms, making it costly to
collect and difficult to transfer to new environments and embodiments.
These challenges also affect \ac{vla} models, which extend pretrained
\acp{vlm} to robot control by learning to predict actions from task-specific
demonstrations~\cite{zitkovich2023rt2,kim2025openvla,black2025pi0,black2025pi05,zheng2025xvla}.
Their improvements on manipulation benchmarks do not consistently translate
into generalization beyond training conditions: changes in scenes,
instructions, or embodiments can cause sharp performance
drops~\cite{fei2025liberoplus,yuan2026qwenrobotmanip,zhou2025liberopro}.
Qwen-RobotManip advocates aligning heterogeneous manipulation data in
representation, motion, and behavior before \acs{vla}
training~\cite{yuan2026qwenrobotmanip}.

Another line of work learns and evolves robot control policies through
iterative interaction with the environment, reducing reliance on
task-specific trajectory
demonstrations~\cite{chen2026gap,he2026skill2real,liang2026robofoundry,lu2026aspire,zhang2026playful}.
In this setting, \acp{llm} or \acp{vlm} write executable policy code and
continually revise it using interaction
feedback~\cite{liang2023code,lu2026aspire,zhang2026playful}.
Successful executions and validated repairs are distilled into programmatic skills, 
accumulating cross-task knowledge and enabling agents to solve diverse tasks through skill composition.~\cite{he2026skill2real,lu2026aspire,zhang2026playful}.
This line of work represents a pragmatic middle ground between learning
from demonstrations in predefined settings and open-ended continual learning.
Its demonstrated scope, however, remains bounded by task definitions,
hardware configurations, and benchmark suites.
For example, GaP formulates policy evolution for variational automation
tasks, where a specified task is repeated within a fixed workcell with
bounded shape and pose variations~\cite{chen2026gap}.
SHAPER evaluates evolved skills under fixed execution interfaces, leaving
cross-embodiment transfer underexplored~\cite{wang2026shaper}.
Skill2Real learns programmatic skills separately on LIBERO-90 and
robosuite~\cite{he2026skill2real}.
These boundaries and misalignments constrain the experience generalization across tasks, scenarios and embodiments.
Naturally, they motivate three research questions.
\textit{(i) Can alignment enable unified learning of manipulation skills from
interaction across heterogeneous tasks, benchmarks, and embodiments?
(ii) If so, does such learning improve generalization to unseen tasks and scenarios?
(iii) Can the learned skills also transfer to unseen benchmarks and new embodiments?}

Inspired by the \emph{alignment first, then scale}
principle~\cite{yuan2026qwenrobotmanip}, we present \method, a unified framework
designed to align heterogeneous settings (e.g., embodiments, environments,
and benchmark cases) for learning robot manipulation skills through
iterative interactions without trajectory demonstrations or updates to model weights.
We realize this principle by introducing
(i) a unified environment representation that describes observed scene
entities as typed \emph{Objects} and exposes robot and camera capabilities
through a shared Python package;
(ii) composable skills that use reusable library functions
and declare the action interfaces required for execution, with adapters
providing their hardware-specific implementations;
(iii) a learnable orchestration policy that guides the agent model to write
and execute programs in response to observations and rollout outcomes; and
(iv) a canonical case representation that consistently specifies task
instructions and success conditions, embodiments, scenes, and initial
states across various benchmarks.
During the skill learning stage, mixed rollouts of these cases provide
aggregated feedback for jointly optimizing the shared programmatic
components and the orchestration policy.
This alignment enables each agent model to learn one shared repository from
180 tasks across LIBERO-PRO, RoboTwin 2.0, and RoboCasa365, covering six
robot embodiments. Learning interleaves these tasks and embodiments,
yielding 2,700 learning rollouts per agent model
(Figure~\ref{fig:overview}(b)). UniCraft demonstrates improved generalization
across 66 unseen-scenario cases and 36 held-out tasks with continued refinement,
as well as zero-shot transfer capability to four
new robot embodiments and the unseen robosuite~\cite{zhu2020robosuite} and
RoboDojo~\cite{chen2026robodojo} benchmarks. Our main contributions are:
\begin{itemize}[leftmargin=*,labelindent=0pt]
    \item \textbf{Methodology.} We introduce \method, which aligns execution
    interfaces and learning cases, enabling joint learning and refinement of
    one shared skillset and its orchestration policy across benchmarks and embodiments.
    \item \textbf{Interleaved training.} We interleave interaction experience
    from 300 task--embodiment pairs spanning 3 benchmarks and 6 embodiments
    to learn one shared skillset per agent model for reuse across heterogeneous settings.
    \item \textbf{Generalization.} We demonstrate improved generalization
    across 3 agent models over 3 consecutive training epochs. From the first
    to the third epoch, pooled task success on unseen scenarios and tasks
    increases by \mbox{9--28} percentage points on LIBERO-PRO, 10.8--13.3 on RoboTwin,
    and 3.1--14.4 on RoboCasa365.
    \item \textbf{Zero-shot transfer.} We show that \method can transfer
    zero-shot across 4 new robot embodiments, 2 unseen benchmarks, and 5
    additional agent models.
\end{itemize}
\flushbottom
\section{Related Work}
\label{sec:related_work}

\noindent\textbf{Cross-embodiment training.}
\ac{vla} models learn to predict actions from robot
demonstrations~\cite{zitkovich2023rt2,kim2025openvla,black2025pi05}.
Cross-embodiment training combines demonstrations across robotic platforms.
Open X-Embodiment pools data from 22 robot embodiments and demonstrates
positive transfer across robots~\cite{openx2023}.
X-VLA uses source-specific soft prompts to accommodate heterogeneous
datasets~\cite{zheng2025xvla}.
Qwen-RobotManip further aligns state-action representations, motion, and
behavior for large-scale training~\cite{yuan2026qwenrobotmanip}.
These approaches address heterogeneity within shared action-prediction
models. Our work examines alignment at the level of interfaces for reusable
manipulation skills.\par
\noindent\textbf{Code-generated robot policies.}
\ac{cap} turns language instructions into Python policies that call APIs
for perception and control, allowing explicit branches and feedback
loops~\cite{liang2023code}.
Instruct2Act grounds generated programs with visual foundation
models~\cite{huang2023instruct2act}, while VoxPoser writes code that
composes 3D value maps for motion planning~\cite{huang2023voxposer}.
RoboScript improves deployability through a simulation-to-real interface,
syntax checks, and Gazebo validation~\cite{chen2024roboscript}.
CaP-Agent0, introduced in CaP-X, synthesizes nine task-agnostic helper
functions from successful rollouts and uses execution feedback to recover
failures~\cite{fu2026capx}.
These helpers provide reusable components for subsequent program generation.
RECENT extends reuse across embodiments by preserving skill semantics
while adapting execution bindings through code refactoring~\cite{choi2026recent}.\par
\noindent\textbf{Robotic manipulation skill learning.}
Interaction-based skill learning retains executable procedures and
validated repairs for reuse across tasks~\cite{chen2026gap,lu2026aspire,zhang2026playful,liang2026robofoundry,he2026skill2real}.
GaP refines computation graphs built from modular skills through
simulation rehearsal for variational automation in fixed workcells~\cite{chen2026gap}.
ASPIRE transfers guidance distilled from validated repairs to support
further debugging on a target robot~\cite{lu2026aspire}.
Skill2Real learns hierarchical skills in simulation and transfers the frozen
skills through a shared API, while its LIBERO-90 and robosuite skill
collections are trained independently~\cite{he2026skill2real}.
These studies differ in how experience is retained and what adaptation
is needed in a new setting.
Our focus is on how aligned interfaces let experience from heterogeneous tasks,
benchmarks, and embodiments jointly refine one shared skillset.
\section{Method}
\label{sec:method}

Here we describe alignment, joint refinement of skills and their orchestration policy, and online skill composition.

\subsection{Aligning Heterogeneous Manipulation Settings}
\label{sec:alignment}

\noindent\textbf{Aligned geometry.}
Translating high-level cognitive plans into real-world manipulation requires
grounding abstract semantic relations in metric physical space.
Although \acp{vlm} can identify interaction candidates and reason about
spatial relationships, their predictions may require additional grounding
in metric geometry and the robot's current state for reliable closed-loop
execution~\cite{chen2024spatialvlm,yang2025thinking,ma2025threedsrbench}.
We therefore build on object-centric
manipulation~\cite{pan2025omnimanip,huang2024copa,huang2025rekep} and
represent observed scene entities as typed \emph{Objects} carrying image
bounding boxes and the measured position and extent of their visible
regions, with every measurement, target, and tolerance expressed in one
episode-local frame. Each episode fixes this frame by its origin and a
fixed upward $z$-axis, and the camera adapter reports intrinsics $K$ together with the capture-time rotation $R_t$ and
translation $\mathbf t_t$. A pixel $(u,v)$ with measured depth $d_t(u,v)$
is back-projected into this frame as
\begin{equation}
  \mathbf p^{\,w}_t
  = R_t\!\left(d_t(u,v)\,K^{-1}[u,v,1]^{\top}\right) + \mathbf t_t,
  \label{eq:back-projection}
\end{equation}
so that a region selected in an image becomes a point a skill can reason
about in meters.

A shared metric frame is necessary but not sufficient, because two devices
can agree on where a target is and still disagree on how to touch it. We
therefore fix the orientation convention of the tool frame as well.
Writing $\hat{\mathbf z}_a$ for the approach axis and $\hat{\mathbf x}_g$
for the jaw-closing axis, we define
\begin{equation}
  R^{\mathrm{tcp}} = \big[\,\hat{\mathbf x}_g,\;
  \hat{\mathbf y}_g,\; \hat{\mathbf z}_a\,\big],
  \qquad \hat{\mathbf y}_g
  = \frac{\hat{\mathbf z}_a\times\hat{\mathbf x}_g}
         {\lVert \hat{\mathbf z}_a\times\hat{\mathbf x}_g\rVert},
  \label{eq:tcp-frame}
\end{equation}
so that the approach and grasp directions are carried by the target itself
rather than implied by the executing device. The convention is inherited where a
benchmark already defines a tool site, and otherwise recovered from the
device's finger geometry at assembly, not supplied as a per-device
rotation matrix. Scene objects thereby provide geometric arguments to
shared skills; Figure~\ref{fig:alignment}(a) shows these representations and skill calls,
in which a camera observation supplies a target the agent passes to
\FunctionName{robot.pick(target)}.

\newcommand{\AlignmentCaption}{%
Unified environment representation and shared skill execution.
(a) Camera and Robot expose data and callable skills; Scene Objects carry
measured geometry.
(b) Shared skills invoke Actions through their native adapter bindings.
Dashed arrows bind the highlighted \FunctionName{actions.set\_gripper()}
call to directional commands in LIBERO-PRO and position targets in RoboTwin.}
\begin{figure}[t]
    \centering
    \includegraphics[page=2,width=\linewidth]{figures.pdf}
    \caption{\AlignmentCaption}
    \label{fig:alignment}
\end{figure}

\Needspace{4\baselineskip}
\noindent\textbf{Execution contracts.}
Deploying manipulation policies across heterogeneous robotic platforms
poses a trade-off: high-level operational behaviors such as grasping and
placing are largely embodiment-agnostic, while their physical
instantiation is governed by disparate kinematics, coordinate frames, and
native low-level controllers. We resolve this tension by decoupling
procedural task logic from embodiment-dependent execution through a
modular, capability-based interface, in which the execution hierarchy
comprises four components:
(i) a skill $s\in\mathcal S$ is an executable procedure whose control flow
combines library computations, action calls, and outcome checks, without
internal \acs{llm}/\acs{vlm} calls;
(ii) a library function $\ell\in\mathcal L$ performs reusable computation
on supplied observations and device parameters without issuing device
commands;
(iii) an action interface $a\in\mathcal A$ specifies a sensing or control
operation; and
(iv) an adapter $\alpha$ binds each supported interface $a$ to a native
implementation $\alpha(a)$. An interface
$a=\big(\sigma_a, F_a, \varphi_a\big)$ consists of a signature
$\sigma_a$ of arguments and results, a frame and unit convention $F_a$
for those arguments, and a completion predicate $\varphi_a$ evaluated on
the device's own measurements. A Cartesian motion completes when the
measured tool pose lies within a position and orientation tolerance, and
fails otherwise. For embodiment $r$, let $\mathcal S_r\subseteq\mathcal S$ denote
its assembled skills and $\mathcal A_s\subseteq\mathcal A$ the interfaces
required by skill $s$. An adapter must implement their combined
requirements,
$\mathcal A_{\mathrm{req}}(r)=\bigcup_{s\in\mathcal S_r}\mathcal A_s$,
and each binding must preserve the frame and the predicate together with
the signature, reporting the measured outcome $\hat{\mathbf y}$ of the call
so that $\varphi_a(\hat{\mathbf y})$ decides success. Replacing a measured
value with a commanded one violates the contract, and a motion that was
issued but not achieved is a failure. Device parameters declare available
effectors and their gripper limits, so shared procedures can select an
effector and reject geometrically infeasible targets before acting.

\Needspace{4\baselineskip}
\noindent\textbf{Embodiment bindings.}
Between a shared target and a native command lies a chain that no skill
should see. For a Cartesian target $\mathbf x^{\,w}$, the adapter composes
the frame convention $F_a$, inverse kinematics, and its native controller
into the chain
$\mathbf x^{\,w}\!\to\mathbf x^{\,n}\!\to\mathbf u\!\to\hat{\mathbf y}$,
where $\mathbf x^{\,n}$ is the target in the device's native tool frame,
$\mathbf u$ the joint- or task-level input its controller accepts, and
$\hat{\mathbf y}$ the measured result returned to the skill. Coordinate
conversion, inverse kinematics, and controller conventions are thus
resolved deterministically inside the adapter, and the skill text contains
no frame, unit, or kinematic constant. Figure~\ref{fig:alignment}(b)
illustrates this organization. Fixed Panda and Panda+Omron share grasping
and placing modules, with navigation added for Panda+Omron; Aloha-AgileX
combines the same per-arm capabilities with coordinated lifting. Within
grasping, the shared library function \FunctionName{grasp\_poses()}
computes poses from observed geometry, the selected effector's state, and
its gripper limits. The highlighted \FunctionName{actions.set\_gripper()} call
shows one Action bound to two native mechanisms: LIBERO-PRO repeatedly
sends directional commands, while RoboTwin supplies normalized opening
targets through its native joint-position interface. Both report the
measured gripper state, so the shared skill applies the same predicate and
reaches the same verdict. Mixed-benchmark rollouts therefore refine shared
procedures used by different capability combinations.

\subsection{Offline Learning and Evolution}
\label{sec:offline-learning}

\begin{figure}[t]
    \centering
    \includegraphics[page=3,width=\linewidth]{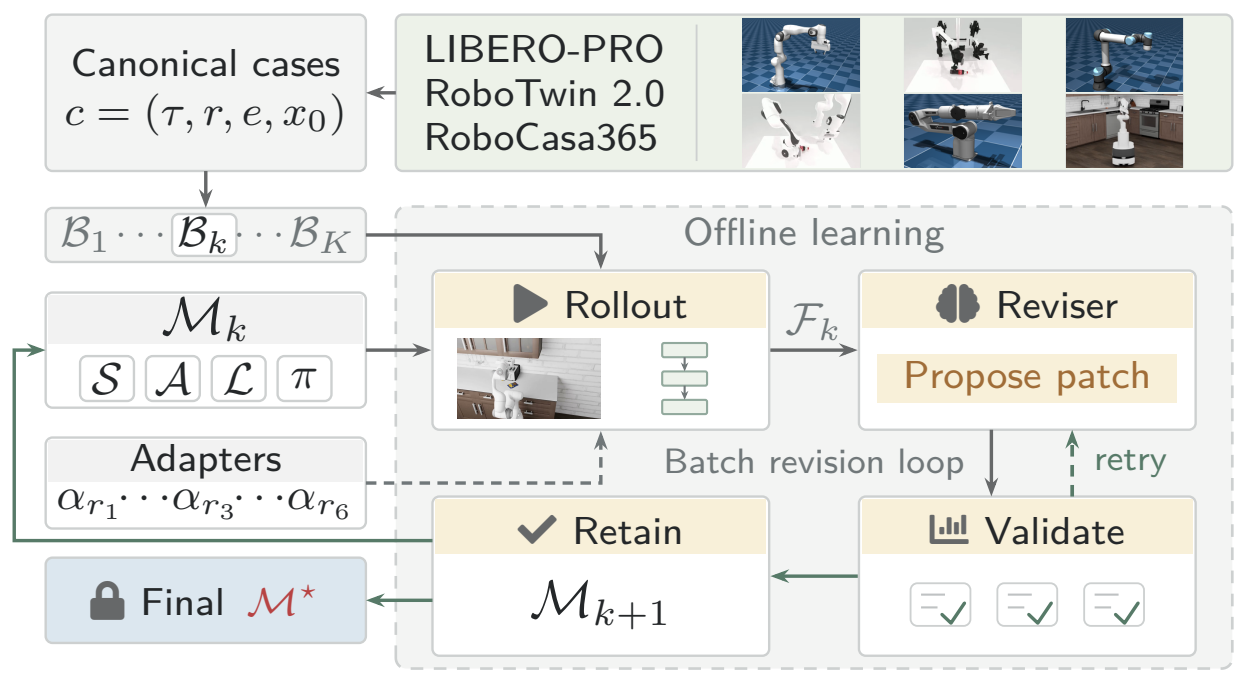}
    \caption{Learning across three source benchmarks and embodiments.
    Canonical cases are processed in batches $\mathcal B_1,\ldots,\mathcal B_K$.
    For each batch $\mathcal B_k$, rollouts under the shared configuration
    $\mathcal M_k$ aggregate multi-round execution feedback
    $\mathcal F_k$. The reviser diagnoses feedback and proposes a patch,
    evaluated through repeated rollouts on selected same-batch cases. Retained updates
    form $\mathcal M_{k+1}$ and carry forward to the next batch along the outer
    loop. After learning, the retained implementation is frozen for deployment.}
    \label{fig:offline-learning}
\end{figure}

Figure~\ref{fig:offline-learning} summarizes our offline learning loop.
We execute aligned cases in batches, revise shared skills and their
orchestration policy using rollout feedback, and carry validated updates
forward to subsequent batches.

\noindent\textbf{Aligning robot learning cases across benchmarks.}
To support shared skill learning across benchmarks, we align heterogeneous learning cases for unified rollout and evaluation across LIBERO-PRO~\cite{liu2023libero,zhou2025liberopro},
RoboTwin 2.0~\cite{chen2025robotwin2},
RoboCasa365~\cite{nasiriany2026robocasa365}, and
RoboDojo~\cite{chen2026robodojo}.
These benchmarks differ in their embodiments and in how a case specifies
its task, scene, and initial state (e.g., declarative descriptions in LIBERO-PRO vs. configurable procedural classes in RoboCasa365). We define a canonical learning case as $c=(\tau,r,e,x_0)$.
Here, $\tau$ specifies the instruction and the success criteria. The embodiment $r$ identifies the robot. The scene configuration $e$ specifies
the layout, object and fixture assets. The initial state $x_0$ determines the starting robot configuration, object
poses, and articulated-fixture states. These fields describe the task
independently of each benchmark's \mbox{configuration} format. For execution,
each benchmark integration selects the native task and robot, configures
the scene, and initializes the starting state through the benchmark's existing setup routines. The native success check
supplies the rollout outcome. The learning pipeline can therefore execute
heterogeneous cases and compare their outcomes.
The learning loop processes these cases in batches
$\mathcal B_1,\ldots,\mathcal B_K$
(Figure~\ref{fig:offline-learning}, \emph{Canonical cases}).

\noindent\textbf{Shared configuration and rollouts.}
We jointly refine the shared programmatic repository
$(\mathcal S,\mathcal A,\mathcal L)$ and manipulation
policy using feedback from aligned cases. At learning
iteration $k$, the retained configuration is
\begin{equation}
    \mathcal M_k=(\mathcal S_k,\mathcal A_k,\mathcal L_k,\pi_k).
    \label{eq:component-configuration}
\end{equation}
Here, $\pi_k$ denotes the manipulation policy guiding the agent's use of
the programmatic components. Execution cost is measured in environment
steps at each benchmark's native control frequency. For case $c$, let
$d_c$ denote its reference execution length and assign a rollout horizon
$H_c=\lceil 2d_c\rceil$. The additional steps provide a controlled margin
for correcting execution errors within the rollout horizon. Reference
trajectory demonstrations are not provided.
For batch $\mathcal B_k$, rollouts under $\mathcal M_k$ produce feedback
$\mathcal F_k$, including native task outcomes, execution traces, and
step counts.

\noindent\textbf{Feedback-driven revision.}
The \ac{llm}-based Reviser (\mbox{Figure~\ref{fig:offline-learning}}) diagnoses
shared failure mechanisms and acquires or refines
$\mathcal S,\mathcal A,\mathcal L$, or updates
$\pi$ when failures arise from skill use or feedback handling.
Candidate revisions are assessed through case-wise comparisons
(Figure~\ref{fig:offline-learning}, \emph{Validate}).
We consider two optimization objectives: \textit{task success rate} and
\textit{execution efficiency}. For each verification case
$c\in\mathcal V$, let $\hat p_c(\mathcal M)$ denote empirical native
success under matched rollout conditions. The improvement from candidate
$\widetilde{\mathcal M}$ is
$\Delta_c=\hat p_c(\widetilde{\mathcal M})-\hat p_c(\mathcal M)$.
We seek improvements subject to the non-regression constraint
$\Delta_c\geq0$ for every checked case, with $\Delta_c>0$ for at least
one case. This prevents improvements on one task from compensating for
degradation on another. When the case-wise success profile is unchanged,
efficiency is compared on matched successful rollout pairs $\mathcal P$.
Let $n_i(\mathcal M)$ be the consumed environment steps and $d_i$ the
corresponding reference length. The paired cost difference is
$\Delta_{\mathrm{step}}=|\mathcal P|^{-1}\sum_{i\in\mathcal P}
[n_i(\widetilde{\mathcal M})-n_i(\mathcal M)]/d_i$,
with $\Delta_{\mathrm{step}}<0$ favoring the candidate.
Candidates passing these checks on same-batch target failures, related cases,
and previously successful controls form $\mathcal M_{k+1}$; the final
configuration $\mathcal M^\star$ is frozen for deployment.

\subsection{Online Manipulation and Skill Transfer}
\label{sec:online-composition}

\noindent\textbf{Programmatic online manipulation.}
During online manipulation, action sequences adapt to the evolving scene
and execution outcomes. With $\mathcal M^\star$ fixed, the manipulation
policy $\pi^\star$ guides the \acs{llm}/\acs{vlm} in synthesizing a Python
program $C_t$ conditioned on the task instruction $g$ and accumulated
interaction history $h_t$. Online execution combines two processes:
(i) intra-program control flow, where variables, branches, and loops
compose skill calls and condition subsequent operations on their returned
results, and (ii) inter-program decision-making, where the
\acs{llm}/\acs{vlm} uses program outputs, exceptions, and explicitly
requested visual observations to synthesize $C_{t+1}$ within the same
rollout. For example, a program can withhold placement when grasp
verification fails, while the subsequent model decision uses a new
observation to plan a corrective retry. This dual-level hierarchy enables dynamic runtime adaptation of action sequences and parameters.

\noindent\textbf{Cross-embodiment transfer.}
A target embodiment assembles compatible modules from the frozen shared
skills and supplies device parameters and adapter bindings satisfying
their action requirements. The learned implementations and manipulation
policy remain unchanged. For example, dual ARX-X5 can reuse grasping and
placing modules learned on Aloha-AgileX through its native action
bindings. Panda+Omron can combine manipulation modules shared with a
fixed Panda with navigation. An online program then determines the call
order and arguments for these modules.

\begingroup
\makeatletter
\setlength{\@fptop}{0pt}
\setlength{\@fpsep}{16pt}
\setlength{\@fpbot}{0pt plus 1fil}
\makeatother
\FloatBarrier
\endgroup
\section{Experiments}
\label{sec:experiments}

\begin{figure*}[t]
    \centering
    \begingroup\color{black}
    \captionsetup{font={small,color=black},labelfont={bf,color=black},hypcap=false}
    \includegraphics[page=4,width=\linewidth]{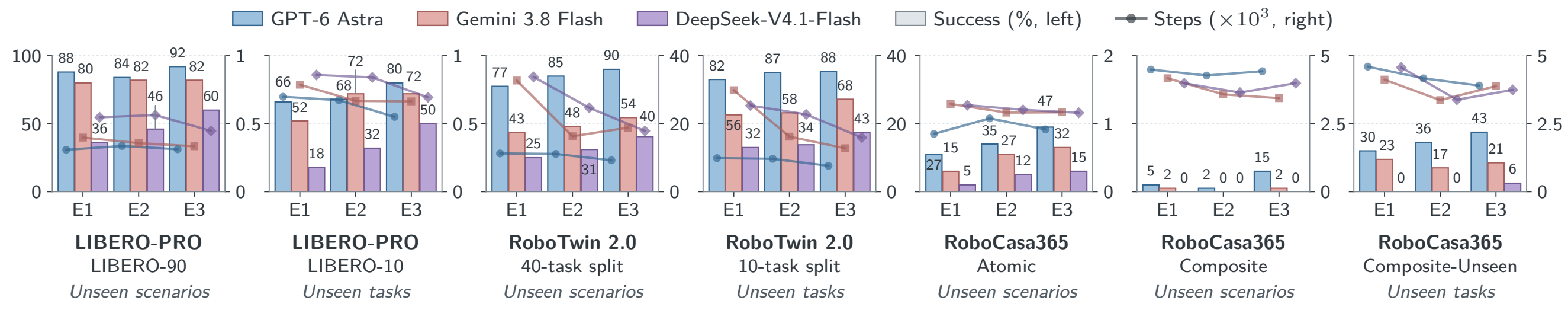}
    \caption{Epoch-wise generalization.
    Bars show full-task success rates and lines show mean execution steps over
    successful and failed rollouts after initialization. Step scales are benchmark-specific.
    Labels truncate percentages to integers. Full results:
    Appendix~\ref{app:additional-results}, Table~\ref{tab:generalization}.}
    \label{fig:generalization-trajectories}
    \endgroup
\end{figure*}

\begingroup
\color{black}
\captionsetup{font={small,color=black},labelfont={bf,color=black},hypcap=false}

\subsection{Experimental Setup}
\label{sec:experimental-setup}

\begin{table}[t]\centering
\captionsetup{font={small,color=black},labelfont={bf,color=black},hypcap=false}
\caption{Learning and evaluation budgets (learning/generalization per agent model). Transfer uses six models for RoboTwin/robosuite and Astra only for RoboDojo; horizon ablations are excluded. B/E/S/T denote benchmark, embodiment, scenario, and task. $\checkmark$/$-$ mark seen/unseen dimensions, and $\approx$ unseen embodiments kinematically similar to seen ones. \mbox{AA/DF/DP/UW} denote Aloha-AgileX and dual Franka/Piper/UR5+WSG.}
\label{tab:benchmark-comparisons}
\fontsize{6.0}{6.8}\selectfont\setlength{\tabcolsep}{0pt}
\renewcommand{\arraystretch}{1.00}
\setlength{\aboverulesep}{.25pt}\setlength{\belowrulesep}{.25pt}
{\color{BudgetInk}
\begin{tabular*}{\linewidth}{@{}
  >{\raggedright\arraybackslash}m{.15\linewidth}
  >{\raggedright\arraybackslash}m{.27\linewidth}
  >{\raggedright\arraybackslash}m{.20\linewidth}
  >{\centering\arraybackslash}m{.075\linewidth}
  >{\centering\arraybackslash}m{.075\linewidth}
  >{\centering\arraybackslash}m{\dimexpr.23\linewidth-24pt\relax}
  *{4}{w{c}{6pt}}@{}}
\toprule[.5pt]
\multirow{2}{*}[-.40pt]{\textbf{Benchmark}} & \multirow{2}{*}[-.40pt]{\textbf{Source}} & \multirow{2}{*}[-.40pt]{\textbf{Embodiment}} & \multirow{2}{*}[-.40pt]{\textbf{Task}} & \multirow{2}{*}[-.40pt]{\textbf{Case}} & \multirow{2}{*}[-.40pt]{\textbf{Rollout}} & \multicolumn{4}{c@{}}{\textbf{Setting}}\\
\cmidrule[.3pt]{7-10}
 & & & & & & \textbf{B} & \textbf{E} & \textbf{S} & \textbf{T}\\
\midrule[.3pt]
\rowcolor{BudgetTrain}\multicolumn{10}{@{}l}{\S\ref{sec:exp-evolution} Training, new cases each epoch {(rollout $=$ task $\times$ embodiment $\times$ 3 epochs $\times$ 3 repeats)}}\\
LIBERO-PRO & \hspace*{4pt}{LIBERO-90} & Fixed Panda & \red{90} & \red{270} & \red{810} & $\checkmark$ & $\checkmark$ & $\checkmark$ & $\checkmark$\\
RoboCasa365 & \hspace*{4pt}{pretrain50} & Panda+Omron & \red{50} & \red{150} & \red{450} & $\checkmark$ & $\checkmark$ & $\checkmark$ & $\checkmark$\\
RoboTwin 2.0 & \hspace*{4pt}{40-task split} & AA DF DP UW & \red{40} & \red{480} & \red{1{,}440} & $\checkmark$ & $\checkmark$ & $\checkmark$ & $\checkmark$\\
\midrule[.3pt]
\rowcolor{BudgetTrain}\multicolumn{10}{@{}l}{\S\ref{sec:exp-generalization} Generalization, one fixed case set {(rollout $=$ task $\times$ embodiment $\times$ 3 epochs $\times$ 5 repeats)}}\\
LIBERO-PRO & \hspace*{4pt}{LIBERO-90} & Fixed Panda & \red{10} & \red{10} & \red{150} & $\checkmark$ & $\checkmark$ & $-$ & $\checkmark$\\
RoboCasa365 & \hspace*{4pt}{Atomic + Composite} & Panda+Omron & \red{16} & \red{16} & \red{240} & $\checkmark$ & $\checkmark$ & $-$ & $\checkmark$\\
RoboTwin 2.0 & \hspace*{4pt}{40-task split} & AA DF DP UW & \red{10} & \red{40} & \red{600} & $\checkmark$ & $\checkmark$ & $-$ & $\checkmark$\\
\cmidrule[.3pt]{1-10}
LIBERO-PRO & \hspace*{4pt}{LIBERO-10} & Fixed Panda & \red{10} & \red{10} & \red{150} & $\checkmark$ & $\checkmark$ & $-$ & $-$\\
RoboCasa365 & \hspace*{4pt}{Composite-Unseen} & Panda+Omron & \red{16} & \red{16} & \red{240} & $\checkmark$ & $\checkmark$ & $-$ & $-$\\
RoboTwin 2.0 & \hspace*{4pt}{10-task split} & AA DF DP UW & \red{10} & \red{40} & \red{600} & $\checkmark$ & $\checkmark$ & $-$ & $-$\\
\midrule[.3pt]
\rowcolor{BudgetTrain}\multicolumn{10}{@{}l}{\S\ref{sec:exp-transfer} Transfer, final epoch {(rollout $=$ case $\times$ models $\times$ 5 repeats)}}\\
RoboTwin 2.0 & \hspace*{4pt}{10-task split} & dual ARX-X5 & \red{10} & \red{10} & \red{300} & $\checkmark$ & $\approx$ & $-$ & $-$\\
\cmidrule[.3pt]{1-10}
RoboDojo & \hspace*{4pt}{all tasks (Astra only)} & dual ARX-X5 & \red{42} & \red{42} & \red{210} & $-$ & $\approx$ & $-$ & $-$\\
\cmidrule[.3pt]{1-10}
robosuite & \hspace*{4pt}{single-arm tasks} & Sawyer & \red{7} & \red{14} & \red{420} & $-$ & $-$ & $-$ & $-$\\
robosuite & \hspace*{4pt}{two-arm tasks} & Baxter / Tiago & \red{4} & \red{16} & \red{480} & $-$ & $-$ & $-$ & $-$\\
\bottomrule[.5pt]\end{tabular*}}
\end{table}

\noindent\textbf{Benchmarks.}
Our evaluation protocol covers \ExperimentBenchmarkCount{} diverse
benchmarks: LIBERO-PRO~\cite{liu2023libero,zhou2025liberopro},
RoboTwin 2.0~\cite{chen2025robotwin2},
RoboCasa365~\cite{nasiriany2026robocasa365},
RoboDojo~\cite{chen2026robodojo}, and
robosuite~\cite{zhu2020robosuite}.
Table~\ref{tab:benchmark-comparisons} summarizes the learning and evaluation budget.

\noindent\textbf{Baselines.}
Table~\ref{tab:baseline-survey} groups baselines by their main control or
refinement mechanism:
(i) VLA baselines use benchmark-adapted checkpoints of
$\pi_{0.5}$~\cite{pi05liberocheckpoint,motuspi05robotwincheckpoint,pi05robocasa365checkpoint},
X-VLA~\cite{zheng2025xvla}, and OpenVLA-OFT~\cite{kim2025oft}.
(ii) Direct EEF baselines use MotorMind~\cite{li2026motormind} on LIBERO-PRO and
GPT-GEAR~\cite{zhou2026gptgear} on RoboTwin 2.0. MotorMind uses a frozen VLM
to propose bounded translations, rotations, and gripper commands; its
deterministic controller executes them with measured feedback. GPT-GEAR
emits end-effector pose deltas through an observe--act interface. Both
methods use their released benchmark control interfaces.
(iii) Code-as-policy agents generate task programs over robot APIs.
They include CaP-Agent0~\cite{fu2026capx} on LIBERO-PRO and
OpenRUA~\cite{chu2026openrua}, which uses native ROS~2 interfaces.
(iv) Agent recursive self-improvement (RSI) methods iteratively refine
reusable skills or agent harnesses from rollout feedback.
TinyAstral~\cite{gong2026tinyastral} and ASPIRE v1.0~\cite{lu2026aspire}
provide benchmark- or embodiment-specific variants.
All methods use their released interfaces without additional training.

\noindent\textbf{Agent harness.}
We use Codex as the underlying agent harness~\cite{openai2026codexappserver}
(Appendix~\ref{app:case-study}).
In each rollout, a frozen snapshot of the system exposes skills and device APIs as a Python package.
The agent writes scripts using this package to dispatch commands to the embodiment via a device server, iteratively acting on observations and execution feedback.

\subsection{Learning from Heterogeneous Experience}
\label{sec:exp-evolution}

\noindent\textbf{Agent models and learning cases.}
We use \ExperimentBackboneCount{} agent models: the closed-source
\ModelGPTAstra{} and \ModelGeminiFlash{}, and the open-source
\ModelDeepSeekFlash{} (1017). All three use \textit{high} reasoning effort.
The runs share a pool of \TrainingTaskCount{} native tasks from
three sources: the \LiberoTaskCount{} LIBERO-90 tasks, \RoboTwinTaskCount{} of
the \RoboTwinTotalTaskCount{} RoboTwin 2.0 tasks under its \textit{hard} setting, and the
\RoboCasaTaskCount{} RoboCasa365 \textit{pretraining} tasks (Table~\ref{tab:benchmark-comparisons}). Cases are randomly shuffled across the task pool, so each epoch interleaves benchmarks and embodiments. The pool covers a fixed-base
arm, \RoboTwinBodyCount{} dual-arm platforms, and a mobile manipulator across
short-horizon tabletop, bimanual, and long-horizon kitchen tasks. All three sources provide native task definitions and success checks,
which the learning loop uses in place of demonstrations.
These assignments give \TrainingPairCount{} task--embodiment pairs ($90 + 50 + 40 \times 4$) over
\TrainingEmbodimentCount{} embodiments, and Appendix
Table~\ref{tab:task-inventory} lists the tasks. We reserve dual ARX-X5, Sawyer,
Baxter, and Tiago, as well as RoboDojo and robosuite, for the transfer experiments
in Section~\ref{sec:exp-transfer}. Each run starts empty,
evolving independently in its worktree under matched budgets.

\noindent\textbf{Epochs, batches, revisions, and budgets.}
Over \TrainingEpochCount{} epochs, each of the \TrainingPairCount{} task--embodiment pairs receives a fresh configuration and initial state every epoch, yielding \TrainingCaseCount{} distinct cases and \TrainingRolloutCount{} base rollouts per agent model at \TrainingRepeatCount{} executions per case. Each epoch is split into \TrainingBatchCount{} batches. All agent models see the same cases in the same order, and one harness revision follows each batch. Each case runs under a horizon of \TrainingHorizonMultiple$\times$ the pinned native evaluation budget of its task, counted in native steps and identical across agent models, configurations, and embodiments. For example, a 400-step LIBERO-90 task receives an 800-step horizon. The steps provide a margin for correcting errors.

\begingroup\color{black}
\noindent\textbf{Evolution results.}
Across three epochs, each with six batches, GPT-6 Astra retained
38.7\% (12/31) of candidate revisions, compared with 40\% (12/30)
for Gemini 3.8 Flash and 39.3\% (11/28) for DeepSeek-V4.1-Flash.
These revisions, detailed in Appendix~\ref{app:revision-history}, cluster
around grasp acquisition and recovery, motion planning and collision avoidance,
visual perception and geometric measurement, and feedback-driven manipulation
and skill orchestration.
\endgroup

\endgroup

\begingroup\color{black}
\frenchspacing
\subsection{Generalization to Unseen Scenarios and Tasks}
\label{sec:exp-seen}
\label{sec:exp-generalization}

\Needspace{5\baselineskip}
\noindent\textbf{Unseen scenarios.}\label{sec:exp-unseen-scenarios}
\looseness=-1
Refinement improves performance on learning tasks under held-out configurations
and initial states (Figure~\ref{fig:generalization-trajectories}). From E1 to E3,
all three agent models improve on RoboTwin~\cite{chen2025robotwin2} and
RoboCasa Atomic~\cite{nasiriany2026robocasa365}. Astra rises from 77.5\% to 90.0\%
and from 27.5\% to 47.5\%, respectively, while reaching 92.0\% on
LIBERO-PRO~\cite{liu2023libero,zhou2025liberopro}. These gains do not extend
uniformly to composite tasks: final success on RoboCasa Composite-Seen remains
only 0--15\% across agent models. The results indicate improved robustness to scene
variation, alongside persistent difficulty with composite manipulation
(see Appendix~\ref{app:composite-task-failures}).

\Needspace{5\baselineskip}
\noindent\textbf{Unseen tasks.}\label{sec:exp-unseen-tasks}
\looseness=-1
We test native task identities excluded from learning. From E1 to E3,
success increases by 14--32 percentage points on LIBERO-10~\cite{liu2023libero}
and 6--11.5 points on RoboTwin~\cite{chen2025robotwin2} across all three agent models.
Mean execution steps also fall in all six model--benchmark comparisons.
For Astra on RoboTwin, success rises from 82.5\% to 88.5\%, while steps
 decrease from 9,864 to 7,609 (22.9\%). RoboCasa compositions~\cite{nasiriany2026robocasa365} show less
consistent gains: Astra improves from 30.0\% to 43.75\%, whereas Gemini
falls from 23.75\% to 21.25\% on these tasks.

\begin{table}[!t]
\centering
\begingroup\color{black}
\definecolor{BaselineAstra}{HTML}{DFECF7}
\definecolor{BaselineGemini}{HTML}{F6E1DE}
\definecolor{BaselineDeepSeek}{HTML}{EBE1F5}
\newcommand{\BaselineModelKey}[2]{%
 {\setlength{\fboxsep}{.5pt}\colorbox{#1}{\strut#2}}}
\newcommand{\BaselineResultsCaption}{%
Full-task success (\%) on unseen scenarios and tasks.
Each VLA baseline uses a separate released checkpoint for each evaluated
benchmark.
Agent columns repeat left to right:
\BaselineModelKey{BaselineAstra}{GPT-6 Astra},
\BaselineModelKey{BaselineGemini}{Gemini 3.8 Flash},
\BaselineModelKey{BaselineDeepSeek}{DeepSeek-V4.1-Flash}.
Task/Rollout gives baseline budgets. A dash means the original release lacks
a checkpoint, integration adapter, or learned skill library needed for that
benchmark. Bold marks the highest success rate across methods and agent
models per benchmark.}
\newcommand{\BaselineVLARows}{%
$\pi_{0.5}$~\cite{pi05liberocheckpoint,motuspi05robotwincheckpoint,pi05robocasa365checkpoint} & \multicolumn{3}{c}{59.00} & \multicolumn{3}{c}{46.00} & \multicolumn{3}{c}{10.63} \\
X-VLA~\cite{xvlaliberocheckpoint,xvlarobotwincheckpoint} & \multicolumn{3}{c}{\textbf{96.00}} & \multicolumn{3}{c}{10.00} & \multicolumn{3}{c}{\BaselineMissing} \\
OpenVLA-OFT~\cite{kim2025oft} & \multicolumn{3}{c}{47.00} & \multicolumn{3}{c}{\BaselineMissing} & \multicolumn{3}{c}{\BaselineMissing} \\
}
\newcommand{\BaselineAgentRows}{%
\BaselineGroup{Direct EEF: VLMs directly propose end-effector commands}
MotorMind~\cite{li2026motormind} & 9.00 & 1.00 & 3.00 & \BaselineMissing & \BaselineMissing & \BaselineMissing & \BaselineMissing & \BaselineMissing & \BaselineMissing \\
GPT-GEAR~\cite{zhou2026gptgear} & \BaselineMissing & \BaselineMissing & \BaselineMissing & 10.00 & 6.00 & 3.00 & \BaselineMissing & \BaselineMissing & \BaselineMissing \\
\BaselineGroup{Code-as-policy agents: generate task programs over robot APIs}
CaP-Agent0~\cite{fu2026capx} & 18.00 & 10.00 & 7.00 & \BaselineMissing & \BaselineMissing & \BaselineMissing & \BaselineMissing & \BaselineMissing & \BaselineMissing \\
OpenRUA~\cite{chu2026openrua} & 30.00 & 22.00 & 6.00 & \BaselineMissing & \BaselineMissing & \BaselineMissing & 3.75 & 1.88 & 1.25 \\
\BaselineGroup{Agent RSI: iteratively refine reusable skills or agent harnesses}
TinyAstral~\cite{gong2026tinyastral} & 62.00 & 56.00 & 34.00 & 64.00 & 60.00 & 56.00 & \BaselineMissing & \BaselineMissing & \BaselineMissing \\
ASPIRE~\cite{lu2026aspire} & 62.00 & 18.00 & 16.00 & \BaselineMissing & \BaselineMissing & \BaselineMissing & \BaselineMissing & \BaselineMissing & \BaselineMissing \\
\textbf{UniCraft (ours)} & 86.00 & 77.00 & 55.00 & \textbf{89.25} & 61.25 & 42.00 & \textbf{37.50} & 19.38 & 6.88 \\
}
\captionsetup{font={small,color=black},labelfont={bf,color=black},hypcap=false,
 position=top,aboveskip=0pt,belowskip=4pt}
\caption{\BaselineResultsCaption}
\label{tab:baseline-survey}
\label{tab:baseline-results}
\definecolor{BaselineGroupFill}{HTML}{EDF2F6}
\definecolor{BaselineMuted}{HTML}{64717D}
\newcommand{\BaselineMissing}{\textcolor{BaselineMuted}{\textemdash}}
\newcommand{\BaselineHeader}[1]{{\fontsize{6.6}{8.2}\selectfont\bfseries #1}}
\newcommand{\BaselineGroup}[1]{%
 \rowcolor{BaselineGroupFill}\multicolumn{10}{@{}l@{}}{\strut #1}\\}
\newlength{\BaselineMethodWidth}\setlength{\BaselineMethodWidth}{.28\linewidth}
\newlength{\BaselineResultWidth}
\setlength{\BaselineResultWidth}{\dimexpr(\linewidth-\BaselineMethodWidth-6pt)/9\relax}
\fontsize{7.0}{8.2}\selectfont
\setlength{\tabcolsep}{0pt}
\renewcommand{\arraystretch}{1.06}
\setlength{\aboverulesep}{1pt}\setlength{\belowrulesep}{1pt}
\begin{tabular}{@{}
 >{\raggedright\arraybackslash}m{\BaselineMethodWidth}
 >{\columncolor{BaselineAstra}\centering\arraybackslash}m{\BaselineResultWidth}
 >{\columncolor{BaselineGemini}\centering\arraybackslash}m{\BaselineResultWidth}
 >{\columncolor{BaselineDeepSeek}\centering\arraybackslash}m{\BaselineResultWidth}
 @{\hspace{3pt}}
 >{\columncolor{BaselineAstra}\centering\arraybackslash}m{\BaselineResultWidth}
 >{\columncolor{BaselineGemini}\centering\arraybackslash}m{\BaselineResultWidth}
 >{\columncolor{BaselineDeepSeek}\centering\arraybackslash}m{\BaselineResultWidth}
 @{\hspace{3pt}}
 >{\columncolor{BaselineAstra}\centering\arraybackslash}m{\BaselineResultWidth}
 >{\columncolor{BaselineGemini}\centering\arraybackslash}m{\BaselineResultWidth}
 >{\columncolor{BaselineDeepSeek}\centering\arraybackslash}m{\BaselineResultWidth}@{}}
\toprule[.5pt]
\BaselineHeader{Method} &
\multicolumn{3}{c}{\BaselineHeader{LIBERO-PRO}} &
\multicolumn{3}{c}{\BaselineHeader{RoboTwin 2.0}} &
\multicolumn{3}{c@{}}{\BaselineHeader{RoboCasa365}}\\
Task / Rollout & \multicolumn{3}{c}{20 / 100} &
\multicolumn{3}{c}{20 / 100} & \multicolumn{3}{c@{}}{32 / 160}\\
\midrule[.3pt]
\BaselineGroup{VLA: learned action prediction from vision and language}
\BaselineVLARows
\BaselineAgentRows
\bottomrule[.5pt]
\end{tabular}
\endgroup
\end{table}
\newcommand{\SkillReuseCaption}{%
Shared component refinement, reuse, and backend cost for GPT-6 Astra.
(a) Reconstructed changes to skills (28) and library functions (7), normalized separately.
(b) Estimated percentage of rollouts invoking each entry at least once.
(c) Backend call shares (hollow) and time shares (filled):
\mbox{$100N_k/\sum_{j=1}^{5}N_j$} and
\mbox{$100T_k/\sum_{j=1}^{5}T_j$},
where $N_k$ and $T_k$ are matched call count and cumulative backend time
for group $k$ in each benchmark.
Timing excludes LLM waiting and unrecorded client-side computation.}
\begin{figure}[t]
    \centering
    \begingroup\color{black}
    \captionsetup{font={small,color=black},labelfont={bf,color=black},hypcap=false}
    \includegraphics[page=5,width=\linewidth]{figures.pdf}
    \caption{\SkillReuseCaption}
    \label{fig:skill-revision-reuse}
    \endgroup
\end{figure}
\begin{figure*}[t]
  \centering
  \includegraphics[page=6,width=\linewidth]{figures.pdf}
  \captionsetup{font={small,color=black},labelfont={bf,color=black},hypcap=false}
  \caption{\textbf{Execution budgets on robosuite.}
  Agent models (left to right): \ModelGPTAstra{}, \ModelGPTSol{},
  \ModelGPTTerra{}, \ModelGeminiFlash{}, \ModelDeepSeekFlash{}, and \ModelMiniMaxFlash{}.
  (a) Stacked bars show outcome percentages at 1,000--4,000 environment steps,
  pooling 30 cases with five repeats (150 rollouts) per condition.
  (b) Split violins show model-request counts, including retries, for
  success (left) and failure (right).
  Dotted/dashed lines mark quartiles/medians; solid ticks mark the 200-request cap.}
  \label{fig:robosuite-budget-ablation}
\end{figure*}

\begingroup
\setlength{\parfillskip}{0pt}
\frenchspacing
\noindent\textbf{Baseline comparison.}
Table~\ref{tab:baseline-survey} pools both evaluation panels per benchmark.
The evaluated VLA baselines use benchmark-specific checkpoints, while the RSI
baselines retain setting-specific skill variants. In contrast, UniCraft jointly
refines one shared skillset and orchestration policy per agent model across all
three learning benchmarks and reuses the same skill implementations across
embodiments.
X-VLA leads on LIBERO-PRO, achieving 96\% success versus 86\% for UniCraft with Astra.
However, its RoboTwin-specific checkpoint achieves only 10\%, while UniCraft
reaches 89.25\% across four evaluated embodiments.
UniCraft also achieves 37.50\% on RoboCasa365, compared with 10.63\% for $\pi_{0.5}$.
With Gemini, UniCraft achieves 77\% on LIBERO-PRO, exceeding TinyAstral's
56\%~\cite{gong2026tinyastral} and ASPIRE's 18\%~\cite{lu2026aspire}.
Together, these results show that a shared programmatic repository supports
competitive performance across heterogeneous manipulation settings.
\par\endgroup

\noindent\textbf{Component reuse and execution cost.}
Figure~\ref{fig:skill-revision-reuse} summarizes reconstructed changes,
estimated component use, and matched backend cost for Astra.
Panel (a) reconstructs 28 changes to skills and 7 to library functions; motion and grasp
proposals account for half of the skill changes. Panel (b) estimates use across
benchmarks: \FunctionName{pick()} appears in 54--64\% of rollouts and
\FunctionName{move\_to()} in 81--96\%. Panel (c) compares call and time shares across
five backend operation groups. On RoboCasa, execution accounts for 27.3\% of
matched calls but 70.0\% of backend time, whereas state reads account for
43.9\% and 0.16\%, respectively.
\endgroup

\begingroup\color{black}
\captionsetup{font={small,color=black},labelfont={bf,color=black},hypcap=false}

\Needspace{4\baselineskip}
\begingroup\color{black}
\subsection{Zero-Shot Transfer across Unseen Embodiments, Benchmarks, and Agent Models}
\label{sec:exp-transfer}
\label{sec:exp-benchmark-transfer}
\label{sec:exp-backbone-transfer}

We freeze the final repository and orchestration policy learned by
\ModelGPTAstra{} and reuse them with five other open- and closed-source
agent models: \ModelGPTSol{}, \ModelGPTTerra{},
\ModelGeminiFlash{}, \ModelDeepSeekFlash{}, and \ModelMiniMaxFlash{}.
\ModelGPTAstra{} is the source-model reference.
We run five rollouts per case with doubled native step budgets for
RoboTwin and robosuite, as in Section~\ref{sec:exp-evolution}.
\par\endgroup

\begingroup\color{black}
\Needspace{5\baselineskip}
\noindent\textbf{New embodiments and benchmarks.}
We evaluate the ten held-out RoboTwin tasks (split10) on dual ARX-X5.
It shares the dual 6-DoF arm structure of the learning-time Aloha-AgileX
platform. We configure coordinate transforms and gripper limits from
simulator-provided parameters and implement native action bindings.
These bindings are established without tuning on target-task feedback.
Astra achieves 74.0\% success
(Table~\ref{tab:transfer-results}), compared with 88.5\% on the same task
split averaged over the four learning-time embodiments, retaining 83.6\%
of reference performance. This demonstrates substantial reuse across
embodiments. The remaining 14.5-point gap is consistent with reachability
and contact geometry differences that interface alignment alone cannot remove.

We further transfer to the unseen robosuite benchmark~\cite{zhu2020robosuite},
covering all 11 core task families: seven single-arm tasks on
Sawyer~\cite{rethinkroboticssawyer} and four bimanual tasks on both
Baxter~\cite{rethinkroboticsbaxter} and Tiago~\cite{palroboticstiago}.
These embodiments span
single-arm, fixed-base bimanual, and mobile bimanual platforms with torso
lifting. We reuse the frozen repository through the same task-independent
adapter setup. At 2K steps, success pooled across all
cases reaches 66.7\% for Astra, 67.3\% for Sol, 48.0\% for Terra,
28.0\% for Gemini, 32.7\% for DeepSeek, and 24.7\% for MiniMax.
Sol performs comparably to the source model without model-specific skill
refinement, supporting reuse with other agent models.
The wider spread across agent models indicates that portable
skills remain dependent on the execution model's ability to use them
effectively.
\par\endgroup

\begin{table}[!t]
\centering
\begingroup\color{black}
\captionsetup{font={small,color=black},labelfont={bf,color=black},hypcap=false,position=top,aboveskip=0pt,belowskip=4pt}
\caption{Zero-shot transfer with frozen \ModelGPTAstra{}
skills and policy. Success (\%); robosuite gives min--max success over
1,000--4,000 environment steps.
Task/Rollout lists tasks and rollouts per setting.
Figure~\ref{fig:robosuite-budget-ablation} reports outcomes and request statistics by budget.}
\label{tab:transfer-results}
\label{tab:benchmark-transfer-results}
\fontsize{8}{10}\selectfont
\setlength{\tabcolsep}{0pt}
\setlength{\aboverulesep}{1.5pt}\setlength{\belowrulesep}{1.5pt}
\renewcommand{\arraystretch}{1.12}
\newlength{\TransferModelWidth}\setlength{\TransferModelWidth}{.33\linewidth}
\newlength{\TransferArxWidth}\setlength{\TransferArxWidth}{.19\linewidth}
\newlength{\TransferMetricWidth}
\setlength{\TransferMetricWidth}{\dimexpr(\linewidth-\TransferModelWidth-\TransferArxWidth-6pt)/3\relax}
\newcommand{\TransferHeader}[1]{{\fontsize{7.2}{8.6}\selectfont\bfseries #1}}
\begin{tabular}{@{}
  >{\raggedright\arraybackslash}m{\TransferModelWidth}
  >{\centering\arraybackslash}m{\TransferArxWidth}
  @{\hspace{6pt}}
  *{3}{>{\centering\arraybackslash}m{\TransferMetricWidth}}
  @{}}
\toprule[.5pt]
\TransferHeader{Benchmark} &
  \makebox[\TransferArxWidth][c]{\TransferHeader{RoboTwin 2.0}~\cite{chen2025robotwin2}} &
  \multicolumn{3}{c@{}}{\TransferHeader{robosuite}~\cite{zhu2020robosuite}}\\
\cmidrule[.3pt]{3-5}
\TransferHeader{Embodiment} &
  \TransferHeader{dual ARX-X5} &
  \TransferHeader{Sawyer} & \TransferHeader{Baxter} & \TransferHeader{Tiago}\\
\TransferHeader{Task / Rollout} &
  10 / 50 & 7 / 70 & 4 / 40 & 4 / 40\\
\midrule[.35pt]
\ModelGPTAstra{} & 74.0 & 60.0--88.6 & 45.0--92.5 & 32.5--75.0\\
\ModelGPTSol{} & 66.0 & 55.7--82.9 & 45.0--70.0 & 25.0--67.5\\
\ModelGPTTerra{} & 50.0 & 51.4--68.6 & 40.0--70.0 & 25.0--45.0\\
\ModelGeminiFlash{} & 50.0 & 28.6--51.4 & 15.0--35.0 & 10.0--25.0\\
\ModelDeepSeekFlash{} & 34.0 & 30.0--51.4 & 10.0--25.0 & 15.0--27.5\\
\ModelMiniMaxFlash{} & 26.0 & 37.1--51.4 & 20.0--25.0 & 5.0--22.5\\
\bottomrule[.5pt]
\end{tabular}
\endgroup
\end{table}

\begingroup\color{black}
\Needspace{5\baselineskip}
\noindent\textbf{Failure reason analysis and ablation study.}
We examine whether failures primarily reflect budget constraints or
ineffective online orchestration. At the standard 2K horizon,
environment-step exhaustion dominates failures for five of six models,
motivating horizon ablation with the 200-request cap fixed. We evaluate
1K, 3K, and 4K horizons alongside the 2K reference.
Figure~\ref{fig:robosuite-budget-ablation} reports outcome breakdowns in
(a) and API-request distributions for successful and failed rollouts in
(b). Increasing the horizon from 2K to 4K improves Astra from 66.7\% to
83.3\% and Sol from 67.3\% to 75.3\%, with neither model exhausting the
request budget at any horizon. Terra also improves from 48.0\% to 62.0\%,
although its request-cap terminations rise from 0.7\% to 10.7\%. At 4K,
successful rollouts across these three models use median request counts
of just 23.5--33, suggesting efficient use of model calls. MiniMax
exhibits an atypical failure pattern: request exhaustion accounts for
40.7\% and 56.0\% of rollouts at 2K and 4K, while success
remains at 24.7\%. More horizons benefit most models, but cannot compensate for failure to act.
\par\endgroup

\begin{table}[!t]
\centering
\captionsetup{font={small,color=black},labelfont={bf,color=black},hypcap=false,position=top,aboveskip=0pt,belowskip=4pt}
\caption{Per-task zero-shot success on RoboDojo~\cite{chen2026robodojo}
(successful rollouts out of five). Types are Generalization (G),
Memory (M), Precision (P), Long (L),
and Open (O).}
\label{tab:robodojo-results}
\fontsize{7.5}{8.2}\selectfont
\setlength{\tabcolsep}{0pt}
\setlength{\aboverulesep}{1pt}
\setlength{\belowrulesep}{1pt}
\newlength{\DojoNameWidth}
\setlength{\DojoNameWidth}{\dimexpr(\columnwidth-6.3pt)/2-42pt\relax}
\newcommand{\DojoHeader}[1]{{\fontsize{7}{8}\selectfont\bfseries #1}}
\begin{tabular}{@{}
  W{l}{\DojoNameWidth}@{\hspace{2pt}}W{c}{15pt}@{\hspace{2pt}}W{r}{23pt}
  @{\hspace{3pt}\vrule width .3pt\hspace{3pt}}
  W{l}{\DojoNameWidth}@{\hspace{2pt}}W{c}{15pt}@{\hspace{2pt}}W{r}{23pt}@{}}
\toprule[.5pt]
\DojoHeader{Name (1--21)} & \DojoHeader{Type} & \DojoHeader{Success} &
\DojoHeader{Name (22--42)} & \DojoHeader{Type} & \DojoHeader{Success}\\
\midrule[.3pt]
Stack bowls & G & 5/5 & Pour balls into vase & P & 4/5\\
Push T & G & 2/5 & Play xylophone & P & 5/5\\
Pack objects into box & G & 1/5 & Deposit coin & P & 1/5\\
Fold clothes & G & 1/5 & Insert key & P & 1/5\\
Hang mugs & G & 1/5 & Build tower & P & 4/5\\
Sweep blocks & G & 3/5 & Put bottles into dustbin & L & 2/5\\
Pour liquid into cup & G & 3/5 & Fill pen holder & L & 3/5\\
Make toast & G & 0/5 & Classify objects & L & 2/5\\
Arrange largest number & G & 4/5 & Play tic tac toe & L & 3/5\\
Sort nesting dolls\textellipsis{} & G & 2/5 & Fill egg holder & L & 1/5\\
Store laptop\textellipsis{} & G & 1/5 & Organize table & L & 2/5\\
Stack blocks & G & 5/5 & Make kong & L & 2/5\\
Cover blocks & M & 4/5 & Play stacking toy & L & 2/5\\
Match and pick\textellipsis{} & M & 4/5 & Align blocks & O & 1/5\\
Swap blocks & M & 3/5 & General pickup & O & 5/5\\
Swap T & M & 3/5 & Stack blocks by\textellipsis{} & O & 4/5\\
Press by number & M & 5/5 & Solve equation & O & 4/5\\
Imitate sorting\textellipsis{} & M & 4/5 & Classify objects by\textellipsis{} & O & 4/5\\
Fasten screws & P & 4/5 & Pick from conveyor\textellipsis{} & O & 3/5\\
Plug in charger & P & 4/5 & Store tools in toolbox & O & 0/5\\
Insert tubes & P & 2/5 & Pour by language & O & 1/5\\
\bottomrule[.5pt]
\end{tabular}
\end{table}

\Needspace{5\baselineskip}
\noindent\textcolor{black}{\textbf{Zero-shot evaluation on RoboDojo.}}
We reuse frozen skills on RoboDojo, implementing only the adapter and
calibrating the interface for dual ARX-X5. Using \ModelGPTAstra{} with high
reasoning effort, we fix seed and layout ID to 0 and run five rollouts per
task across 42 tasks (210 total). Overall success is 54.8\%
(115/210; Table~\ref{tab:robodojo-results}). Memory performs best (76.7\%),
followed by Precision (62.5\%) and Open (55.0\%). The Generalization
and Long-Horizon task groups achieve 46.7\% and 42.5\%, respectively.
This fixed-layout evaluation demonstrates reuse across all five RoboDojo
task groups without further skill refinement.

\endgroup
\section{Discussion}
\label{sec:discussion}

\noindent\underline{\textit{How does UniCraft execute tasks with frozen skills?}}
We present trajectories, programs using frozen skills, and outcomes for
five benchmarks in Appendix~\ref{app:case-study}.

\noindent\underline{\textit{Why do some cases remain unsolved after revisions?}}
We conduct failure reason analysis on cases that fail across revisions
and repeated rollouts. Specifically, on RoboCasa365, \mbox{\ModelGPTAstra{}} achieves
15\% success on Composite-Seen at Epoch~3, with persistent failures on
\FunctionName{RinseSinkBasin} and \FunctionName{SteamInMicrowave}.
We identify four major failure reasons (Appendix~\ref{app:persistent-failures}).
(i) Revision may leave required capabilities unavailable
(Figs.~\ref{fig:failure-capability-wash}--\ref{fig:failure-capability-pack}).
(ii) Later actions can invalidate completed subgoals
(Figs.~\ref{fig:failure-composition}--\ref{fig:failure-steam}).
(iii) Agents struggle to infer how movable parts respond to their actions
and whether the intended effect occurs
(Figs.~\ref{fig:failure-mechanism}--\ref{fig:failure-door}).
(iv) Fine manipulation remains unreliable when success depends on
precise alignment and stable contact
(Figs.~\ref{fig:failure-contact}--\ref{fig:failure-toolhang}).

\noindent\underline{\textit{Can validated revisions cause regressions, and why?}}
Yes. For example, \ModelGeminiFlash{} scores 23.75\%, 17.50\%, and 21.25\% on
RoboCasa365 Composite-Unseen across Epochs~1--3
(Table~\ref{tab:generalization}).
Reviser design involves a coverage--context trade-off. Specifically,
broader diagnosis increases time, computation, and evidence volume,
straining the Reviser's finite context for root-cause analysis.
On one hand, the Reviser reviews
150 rollouts per batch and can consult epoch-level execution and revision
histories for a broader perspective. On the other, to reduce context load,
we diagnose rollouts concurrently and have the Reviser read case summaries
before full traces. We analyze this trade-off in detail in
Appendix~\ref{app:revision-tradeoff}.
\section{Limitations}
\label{sec:limitations}
\noindent\textbf{Comparison fairness.} Differences in auxiliary tools
and pretrained components limit comparability across methods. \method uses
SAM2.1 for prompted foreground segmentation, whereas ASPIRE's released
stack provides SAM3 segmentation, Contact-GraspNet grasp proposals, and
Molmo-based pixel grounding as a fallback~\cite{lu2026aspire}.
We control other sources of variation through three measures.
\textit{Model inference:} For each agent model, compared methods use the
same API provider and \textit{high} reasoning-effort setting.
\textit{Execution budgets:} Within each evaluation setting, agent methods
share the same per-case limits on simulation steps and model requests.
\textit{Hardware configuration:} All methods are reproduced on the same
machine and hardware configuration. These controls standardize inference
and execution conditions.\par

\noindent\textbf{Evaluation budget.} We evaluate six agent models at
\textit{high} reasoning effort and track learning for
\ExperimentBackboneCount{} of them over \TrainingEpochCount{} epochs.
Due to the cost of model inference and simulation rollouts, we have not
assessed other reasoning budgets or longer learning schedules.\par

\noindent\textbf{Empirical design choices.} We fix the revision
interval at \TrainingRolloutsPerBatch{} scheduled rollouts, comprising
\TrainingBatchSize{} cases with \TrainingRepeatCount{} training executions
each. We use \TrainingBatchCount{} batches per epoch and \TestRepeatCount{}
evaluation rollouts per case. These settings were not systematically
optimized or assessed for sensitivity to batch size, revision frequency,
and repetition count.\par

\noindent\textbf{Real-world validation.} All evaluations are conducted
in simulation. Future work will evaluate transfer to physical robots under
real sensor noise, calibration errors, and differences in contact dynamics.\par
\section{Conclusion}
\label{sec:conclusion}
We presented \method, aligning cases and execution interfaces to refine
shared manipulation skills and their orchestration policy from heterogeneous
feedback. Learning spans three benchmarks and six embodiments without
trajectory demonstrations or model weight updates. Results show improved
generalization to unseen tasks and scenarios, and zero-shot reuse on
new robots and agent models. Alignment offers a basis for reusable skills.\par

\clearpage
\appendix
\nolinenumbers
\raggedbottom
\makeatletter
\setlength{\@dblfptop}{0pt}
\makeatother

\begin{table*}[p]
\section{Task Inventory}
\label{app:task-inventory}
\begingroup\nolinenumbers

\definecolor{InvTrain}{HTML}{EDF2F6}
\definecolor{InvTest}{HTML}{EEF4F2}
\definecolor{InvTransfer}{HTML}{F6F2EC}
\definecolor{InvTarget}{HTML}{F2EFF5}
\definecolor{InvStripe}{HTML}{F7F8FA}
\definecolor{InvRule}{HTML}{98A6B3}
\definecolor{InvInk}{HTML}{293845}
\newlength{\InvGap}\setlength{\InvGap}{3pt}
\newlength{\InvPad}\setlength{\InvPad}{2pt}
\newlength{\InvLane}\setlength{\InvLane}{\dimexpr(\linewidth-3\InvGap)/4\relax}
\newlength{\InvEmbWidth}\setlength{\InvEmbWidth}{40pt}
\newlength{\InvTaskWidth}\setlength{\InvTaskWidth}{\dimexpr\InvLane-\InvEmbWidth-1.5pt\relax}
\newlength{\InvPanelWidth}\setlength{\InvPanelWidth}{\dimexpr2\InvLane+\InvGap\relax}
\newsavebox{\InvTaskBox}
\newcommand{\InvTask}[1]{%
  \sbox{\InvTaskBox}{#1}%
  \ifdim\wd\InvTaskBox>\dimexpr\InvTaskWidth-\InvPad\relax
    \PackageError{inventory}{Task name exceeds its column: #1}{}%
  \fi
  \makebox[\InvTaskWidth][l]{\hspace{\InvPad}\usebox{\InvTaskBox}}}
\newcommand{\InvEmb}[1]{\makebox[\InvEmbWidth][c]{#1}}
\newcommand{\InvHeading}[2]{\parbox[c][10pt][c]{#1}{\raggedright\fontsize{6.2}{7.2}\selectfont\bfseries\hspace*{\InvPad}#2}}
\newcommand{\InvTotal}[2]{\parbox[c][7.5pt][c]{#1}{\raggedright\fontsize{6.2}{7.2}\selectfont\bfseries\hspace*{\InvPad}Total: #2}}
\newcommand{\InvStageHeading}[2]{\parbox[c][10pt][c]{\linewidth}{\fontsize{6.2}{7.2}\selectfont\bfseries
  \hbox to\linewidth{\hspace{\InvPad}#1\hfil#2\hspace{\InvPad}}}}
\newcommand{\InvBenchmark}[2]{\parbox[c][7.5pt][c]{#1}{\raggedright\bfseries\hspace*{\InvPad}#2}}

\color{InvInk}\setlength{\parindent}{0pt}\setlength{\parskip}{0pt}
\captionsetup{font={small,color=black},labelfont={bf,color=black},position=top,hypcap=false,aboveskip=0pt,belowskip=5pt}
\captionof{table}{Task inventory. K/L/S: kitchen/living-room/study; AA/DF/DP/UW: Aloha-AgileX and dual Franka/Piper/UR5+WSG. Learning/generalization counts are per model and epoch; robosuite uses two initializations per task--embodiment pair.}
\label{tab:task-inventory}
\fontsize{6.0}{6.1}\selectfont\setlength{\tabcolsep}{0pt}\renewcommand{\arraystretch}{1}
\setlength{\aboverulesep}{.3pt}\setlength{\belowrulesep}{.3pt}\arrayrulecolor{InvRule}
\noindent\begin{tabular}{@{}l@{\hspace{1.5pt}}l@{\hspace{\InvGap}}l@{\hspace{1.5pt}}l@{\hspace{\InvGap}}l@{\hspace{1.5pt}}l@{\hspace{\InvGap}}l@{\hspace{1.5pt}}l@{}}
\toprule[.4pt]
\multicolumn{8}{@{}l@{}}{\cellcolor{InvTrain}\InvStageHeading{\S\ref{sec:exp-evolution} Unified Robot Learning across Benchmarks (evolving over \TrainingEpochCount{} epochs)}{}}\\
\makebox[\InvTaskWidth][c]{\textbf{Task}} & \InvEmb{\textbf{Embodiment}} & \makebox[\InvTaskWidth][c]{\textbf{Task}} & \InvEmb{\textbf{Embodiment}} & \makebox[\InvTaskWidth][c]{\textbf{Task}} & \InvEmb{\textbf{Embodiment}} & \makebox[\InvTaskWidth][c]{\textbf{Task}} & \InvEmb{\textbf{Embodiment}}\\
\midrule[.25pt]
\rowcolor{InvStripe}
\multicolumn{2}{@{}c@{\hspace{\InvGap}}}{\cellcolor{InvTrain}\InvBenchmark{\InvLane}{LIBERO-PRO: 90 tasks / 90 cases}} & \InvTask{K10: Pudding in top drw. + shut} & \InvEmb{Fixed Panda} & \InvTask{adjust\_bottle} & \InvEmb{AA DF DP UW} & \InvTask{CloseFridge} & \InvEmb{Panda+Omron}\\
\InvTask{K1: Open btm. drw.} & \InvEmb{Fixed Panda} & \InvTask{L1: Soup in basket} & \InvEmb{Fixed Panda} & \InvTask{beat\_block\_hammer} & \InvEmb{AA DF DP UW} & \InvTask{CloseToasterOvenDoor} & \InvEmb{Panda+Omron}\\
\rowcolor{InvStripe}
\InvTask{K1: Open top drw.} & \InvEmb{Fixed Panda} & \InvTask{L1: Cheese in basket} & \InvEmb{Fixed Panda} & \InvTask{blocks\_ranking\_rgb} & \InvEmb{AA DF DP UW} & \InvTask{CoffeeSetupMug} & \InvEmb{Panda+Omron}\\
\InvTask{K1: Open top drw. + bowl in} & \InvEmb{Fixed Panda} & \InvTask{L1: Ketchup in basket} & \InvEmb{Fixed Panda} & \InvTask{click\_alarmclock} & \InvEmb{AA DF DP UW} & \InvTask{CoolBakedCake} & \InvEmb{Panda+Omron}\\
\rowcolor{InvStripe}
\InvTask{K1: Black bowl on plate} & \InvEmb{Fixed Panda} & \InvTask{L1: Tomato sauce in basket} & \InvEmb{Fixed Panda} & \InvTask{click\_bell} & \InvEmb{AA DF DP UW} & \InvTask{DateNight} & \InvEmb{Panda+Omron}\\
\InvTask{K1: Black bowl on cab.} & \InvEmb{Fixed Panda} & \InvTask{L2: Soup in basket} & \InvEmb{Fixed Panda} & \InvTask{dump\_bin\_bigbin} & \InvEmb{AA DF DP UW} & \InvTask{DeliverStraw} & \InvEmb{Panda+Omron}\\
\rowcolor{InvStripe}
\InvTask{K2: Open top drw.} & \InvEmb{Fixed Panda} & \InvTask{L2: Butter in basket} & \InvEmb{Fixed Panda} & \InvTask{grab\_roller} & \InvEmb{AA DF DP UW} & \InvTask{FilterMicrowavableItem} & \InvEmb{Panda+Omron}\\
\InvTask{K2: Back black bowl on plate} & \InvEmb{Fixed Panda} & \InvTask{L2: Milk in basket} & \InvEmb{Fixed Panda} & \InvTask{handover\_block} & \InvEmb{AA DF DP UW} & \InvTask{GetToastedBread} & \InvEmb{Panda+Omron}\\
\rowcolor{InvStripe}
\InvTask{K2: Front black bowl on plate} & \InvEmb{Fixed Panda} & \InvTask{L2: Orange juice in basket} & \InvEmb{Fixed Panda} & \InvTask{handover\_mic} & \InvEmb{AA DF DP UW} & \InvTask{KettleBoiling} & \InvEmb{Panda+Omron}\\
\InvTask{K2: Middle black bowl on plate} & \InvEmb{Fixed Panda} & \InvTask{L2: Tomato sauce in basket} & \InvEmb{Fixed Panda} & \InvTask{hanging\_mug} & \InvEmb{AA DF DP UW} & \InvTask{LoadDishwasher} & \InvEmb{Panda+Omron}\\
\rowcolor{InvStripe}
\InvTask{K2: Middle black bowl on cab.} & \InvEmb{Fixed Panda} & \InvTask{L3: Soup in tray} & \InvEmb{Fixed Panda} & \InvTask{lift\_pot} & \InvEmb{AA DF DP UW} & \InvTask{MeatSkewerAssembly} & \InvEmb{Panda+Omron}\\
\InvTask{K2: Stack front bowl on middle} & \InvEmb{Fixed Panda} & \InvTask{L3: Butter in tray} & \InvEmb{Fixed Panda} & \InvTask{move\_can\_pot} & \InvEmb{AA DF DP UW} & \InvTask{NavigateKitchen} & \InvEmb{Panda+Omron}\\
\rowcolor{InvStripe}
\InvTask{K2: Stack middle bowl on back} & \InvEmb{Fixed Panda} & \InvTask{L3: Cheese in tray} & \InvEmb{Fixed Panda} & \InvTask{move\_playingcard\_away} & \InvEmb{AA DF DP UW} & \InvTask{OpenCabinet} & \InvEmb{Panda+Omron}\\
\InvTask{K3: Frying pan on stove} & \InvEmb{Fixed Panda} & \InvTask{L3: Ketchup in tray} & \InvEmb{Fixed Panda} & \InvTask{move\_stapler\_pad} & \InvEmb{AA DF DP UW} & \InvTask{OpenDrawer} & \InvEmb{Panda+Omron}\\
\rowcolor{InvStripe}
\InvTask{K3: Moka pot on stove} & \InvEmb{Fixed Panda} & \InvTask{L3: Tomato sauce in tray} & \InvEmb{Fixed Panda} & \InvTask{open\_laptop} & \InvEmb{AA DF DP UW} & \InvTask{OpenStandMixerHead} & \InvEmb{Panda+Omron}\\
\InvTask{K3: Stove on} & \InvEmb{Fixed Panda} & \InvTask{L4: Left black bowl in tray} & \InvEmb{Fixed Panda} & \InvTask{open\_microwave} & \InvEmb{AA DF DP UW} & \InvTask{PackIdenticalLunches} & \InvEmb{Panda+Omron}\\
\rowcolor{InvStripe}
\InvTask{K3: Frypan on lit stove} & \InvEmb{Fixed Panda} & \InvTask{L4: Pudding in tray} & \InvEmb{Fixed Panda} & \InvTask{pick\_diverse\_bottles} & \InvEmb{AA DF DP UW} & \InvTask{PickPlaceCounterToCabinet} & \InvEmb{Panda+Omron}\\
\InvTask{K4: Close btm. drw.} & \InvEmb{Fixed Panda} & \InvTask{L4: Salad dressing in tray} & \InvEmb{Fixed Panda} & \InvTask{pick\_dual\_bottles} & \InvEmb{AA DF DP UW} & \InvTask{PickPlaceCounterToStove} & \InvEmb{Panda+Omron}\\
\rowcolor{InvStripe}
\InvTask{K4: Close btm. drw. + open top} & \InvEmb{Fixed Panda} & \InvTask{L4: L bowl atop R, in tray} & \InvEmb{Fixed Panda} & \InvTask{place\_a2b\_right} & \InvEmb{AA DF DP UW} & \InvTask{PickPlaceDrawerToCounter} & \InvEmb{Panda+Omron}\\
\InvTask{K4: Black bowl in btm. drw.} & \InvEmb{Fixed Panda} & \InvTask{L4: R bowl atop L, in tray} & \InvEmb{Fixed Panda} & \InvTask{place\_bread\_basket} & \InvEmb{AA DF DP UW} & \InvTask{PickPlaceSinkToCounter} & \InvEmb{Panda+Omron}\\
\rowcolor{InvStripe}
\InvTask{K4: Black bowl on cab.} & \InvEmb{Fixed Panda} & \InvTask{L5: Red mug on left plate} & \InvEmb{Fixed Panda} & \InvTask{place\_bread\_skillet} & \InvEmb{AA DF DP UW} & \InvTask{PickPlaceToasterToCounter} & \InvEmb{Panda+Omron}\\
\InvTask{K4: Wine bottle in btm. drw.} & \InvEmb{Fixed Panda} & \InvTask{L5: Red mug on right plate} & \InvEmb{Fixed Panda} & \InvTask{place\_burger\_fries} & \InvEmb{AA DF DP UW} & \InvTask{PlaceVegetablesEvenly} & \InvEmb{Panda+Omron}\\
\rowcolor{InvStripe}
\InvTask{K4: Wine bottle on rack} & \InvEmb{Fixed Panda} & \InvTask{L5: White mug on left plate} & \InvEmb{Fixed Panda} & \InvTask{place\_can\_basket} & \InvEmb{AA DF DP UW} & \InvTask{PreSoakPan} & \InvEmb{Panda+Omron}\\
\InvTask{K5: Close top drw.} & \InvEmb{Fixed Panda} & \InvTask{L5: YW mug on right plate} & \InvEmb{Fixed Panda} & \InvTask{place\_cans\_plasticbox} & \InvEmb{AA DF DP UW} & \InvTask{PrepMarinatingMeat} & \InvEmb{Panda+Omron}\\
\rowcolor{InvStripe}
\InvTask{K5: Black bowl in top drw.} & \InvEmb{Fixed Panda} & \InvTask{L6: Pudding left of plate} & \InvEmb{Fixed Panda} & \InvTask{place\_container\_plate} & \InvEmb{AA DF DP UW} & \InvTask{PrepSinkForCleaning} & \InvEmb{Panda+Omron}\\
\InvTask{K5: Black bowl on plate} & \InvEmb{Fixed Panda} & \InvTask{L6: Pudding right of plate} & \InvEmb{Fixed Panda} & \InvTask{place\_dual\_shoes} & \InvEmb{AA DF DP UW} & \InvTask{PrepareCocktailStation} & \InvEmb{Panda+Omron}\\
\rowcolor{InvStripe}
\InvTask{K5: Black bowl on cab.} & \InvEmb{Fixed Panda} & \InvTask{L6: Red mug on plate} & \InvEmb{Fixed Panda} & \InvTask{place\_object\_basket} & \InvEmb{AA DF DP UW} & \InvTask{PrepareCoffee} & \InvEmb{Panda+Omron}\\
\InvTask{K5: Ketchup in top drw.} & \InvEmb{Fixed Panda} & \InvTask{L6: White mug on plate} & \InvEmb{Fixed Panda} & \InvTask{place\_object\_scale} & \InvEmb{AA DF DP UW} & \InvTask{PrepareSausageCheese} & \InvEmb{Panda+Omron}\\
\rowcolor{InvStripe}
\InvTask{K6: Close microw.} & \InvEmb{Fixed Panda} & \InvTask{S1: Book in caddy (front)} & \InvEmb{Fixed Panda} & \InvTask{place\_object\_stand} & \InvEmb{AA DF DP UW} & \InvTask{RemoveBroiledFish} & \InvEmb{Panda+Omron}\\
\InvTask{K6: YW mug in front of white} & \InvEmb{Fixed Panda} & \InvTask{S1: Book in caddy (left)} & \InvEmb{Fixed Panda} & \InvTask{place\_phone\_stand} & \InvEmb{AA DF DP UW} & \InvTask{RinseCuttingBoard} & \InvEmb{Panda+Omron}\\
\rowcolor{InvStripe}
\InvTask{K7: Open microw.} & \InvEmb{Fixed Panda} & \InvTask{S1: Book in caddy (right)} & \InvEmb{Fixed Panda} & \InvTask{place\_shoe} & \InvEmb{AA DF DP UW} & \InvTask{RinseSinkBasin} & \InvEmb{Panda+Omron}\\
\InvTask{K7: White bowl on plate} & \InvEmb{Fixed Panda} & \InvTask{S1: YW mug right of caddy} & \InvEmb{Fixed Panda} & \InvTask{press\_stapler} & \InvEmb{AA DF DP UW} & \InvTask{ScrubCuttingBoard} & \InvEmb{Panda+Omron}\\
\rowcolor{InvStripe}
\InvTask{K7: White bowl right of plate} & \InvEmb{Fixed Panda} & \InvTask{S2: Book in caddy (back)} & \InvEmb{Fixed Panda} & \InvTask{put\_bottles\_dustbin} & \InvEmb{AA DF DP UW} & \InvTask{SearingMeat} & \InvEmb{Panda+Omron}\\
\InvTask{K8: Right moka pot on stove} & \InvEmb{Fixed Panda} & \InvTask{S2: Book in caddy (front)} & \InvEmb{Fixed Panda} & \InvTask{put\_object\_cabinet} & \InvEmb{AA DF DP UW} & \InvTask{SetUpCuttingStation} & \InvEmb{Panda+Omron}\\
\rowcolor{InvStripe}
\InvTask{K8: Stove off} & \InvEmb{Fixed Panda} & \InvTask{S2: Book in caddy (left)} & \InvEmb{Fixed Panda} & \InvTask{rotate\_qrcode} & \InvEmb{AA DF DP UW} & \InvTask{SizeSorting} & \InvEmb{Panda+Omron}\\
\InvTask{K9: Frying pan on cab. shelf} & \InvEmb{Fixed Panda} & \InvTask{S2: Book in caddy (right)} & \InvEmb{Fixed Panda} & \InvTask{shake\_bottle} & \InvEmb{AA DF DP UW} & \InvTask{SlideDishwasherRack} & \InvEmb{Panda+Omron}\\
\rowcolor{InvStripe}
\InvTask{K9: Frying pan on cab. top} & \InvEmb{Fixed Panda} & \InvTask{S3: Book in caddy (front)} & \InvEmb{Fixed Panda} & \InvTask{shake\_bottle\_horizontally} & \InvEmb{AA DF DP UW} & \InvTask{StackBowlsCabinet} & \InvEmb{Panda+Omron}\\
\InvTask{K9: Frying pan under cab. shelf} & \InvEmb{Fixed Panda} & \InvTask{S3: Book in caddy (left)} & \InvEmb{Fixed Panda} & \InvTask{stack\_blocks\_three} & \InvEmb{AA DF DP UW} & \InvTask{SteamInMicrowave} & \InvEmb{Panda+Omron}\\
\rowcolor{InvStripe}
\InvTask{K9: White bowl on cab.} & \InvEmb{Fixed Panda} & \InvTask{S3: Book in caddy (right)} & \InvEmb{Fixed Panda} & \InvTask{stack\_blocks\_two} & \InvEmb{AA DF DP UW} & \InvTask{StirVegetables} & \InvEmb{Panda+Omron}\\
\InvTask{K9: Stove on} & \InvEmb{Fixed Panda} & \InvTask{S3: Red mug right of caddy} & \InvEmb{Fixed Panda} & \InvTask{stamp\_seal} & \InvEmb{AA DF DP UW} & \InvTask{StoreLeftoversInBowl} & \InvEmb{Panda+Omron}\\
\rowcolor{InvStripe}
\InvTask{K9: Frypan on lit stove} & \InvEmb{Fixed Panda} & \InvTask{S3: White mug right of caddy} & \InvEmb{Fixed Panda} & \multicolumn{2}{@{}c@{\hspace{\InvGap}}}{\cellcolor{InvTrain}\InvBenchmark{\InvLane}{RoboCasa365: 50 tasks / 50 cases}} & \InvTask{TurnOffStove} & \InvEmb{Panda+Omron}\\
\InvTask{K10: Close top drw.} & \InvEmb{Fixed Panda} & \InvTask{S4: Middle book on cab. shelf} & \InvEmb{Fixed Panda} & \InvTask{AddSweetener} & \InvEmb{Panda+Omron} & \InvTask{TurnOnElectricKettle} & \InvEmb{Panda+Omron}\\
\rowcolor{InvStripe}
\InvTask{K10: Close drw. + bowl on cab.} & \InvEmb{Fixed Panda} & \InvTask{S4: Left book on shelf top} & \InvEmb{Fixed Panda} & \InvTask{ArrangeBuffetDessert} & \InvEmb{Panda+Omron} & \InvTask{TurnOnMicrowave} & \InvEmb{Panda+Omron}\\
\InvTask{K10: Black bowl in top drw.} & \InvEmb{Fixed Panda} & \InvTask{S4: Right book on cab. shelf} & \InvEmb{Fixed Panda} & \InvTask{CleanBoard} & \InvEmb{Panda+Omron} & \InvTask{TurnOnSinkFaucet} & \InvEmb{Panda+Omron}\\
\rowcolor{InvStripe}
\InvTask{K10: Back butter in drw.; close} & \InvEmb{Fixed Panda} & \InvTask{S4: Right book under cab. shelf} & \InvEmb{Fixed Panda} & \InvTask{CleanMicrowave} & \InvEmb{Panda+Omron} & \InvTask{WashLettuce} & \InvEmb{Panda+Omron}\\
\InvTask{K10: Front butter in drw.; close} & \InvEmb{Fixed Panda} & \multicolumn{2}{@{}c@{\hspace{\InvGap}}}{\cellcolor{InvTrain}\InvBenchmark{\InvLane}{RoboTwin 2.0: 40 tasks / 160 cases}} & \InvTask{CloseBlenderLid} & \InvEmb{Panda+Omron} & \multicolumn{2}{@{}l@{}}{\InvTotal{\InvLane}{180 tasks / 300 cases}}\\
\midrule[.4pt]
\multicolumn{8}{@{}l@{}}{\cellcolor{InvTest}\InvStageHeading{\S\ref{sec:exp-seen} Generalization to Unseen Scenarios and Tasks (a snapshot per epoch)}{}}\\
\multicolumn{4}{@{}c@{}}{\InvHeading{\InvPanelWidth}{Generalization on unseen scenario}} & \multicolumn{4}{@{}c@{}}{\InvHeading{\InvPanelWidth}{Generalization on unseen task}}\\
\makebox[\InvTaskWidth][c]{\textbf{Task}} & \InvEmb{\textbf{Embodiment}} & \makebox[\InvTaskWidth][c]{\textbf{Task}} & \InvEmb{\textbf{Embodiment}} & \makebox[\InvTaskWidth][c]{\textbf{Task}} & \InvEmb{\textbf{Embodiment}} & \makebox[\InvTaskWidth][c]{\textbf{Task}} & \InvEmb{\textbf{Embodiment}}\\
\midrule[.25pt]
\rowcolor{InvStripe}
\multicolumn{2}{@{}c@{\hspace{\InvGap}}}{\cellcolor{InvTest}\InvBenchmark{\InvLane}{LIBERO-PRO: 10 tasks / 10 cases}} & \InvTask{stack\_blocks\_three} & \InvEmb{AA DF DP UW} & \multicolumn{2}{@{}c@{\hspace{\InvGap}}}{\cellcolor{InvTest}\InvBenchmark{\InvLane}{LIBERO-PRO: 10 tasks / 10 cases}} & \InvTask{stack\_bowls\_two} & \InvEmb{AA DF DP UW}\\
\InvTask{K10: Close drw. + bowl on cab.} & \InvEmb{Fixed Panda} & \InvTask{stack\_blocks\_two} & \InvEmb{AA DF DP UW} & \InvTask{K3: Moka pot on lit stove} & \InvEmb{Fixed Panda} & \InvTask{turn\_switch} & \InvEmb{AA DF DP UW}\\
\rowcolor{InvStripe}
\InvTask{K5: Black bowl in top drw.} & \InvEmb{Fixed Panda} & \multicolumn{2}{@{}c@{\hspace{\InvGap}}}{\cellcolor{InvTest}\InvBenchmark{\InvLane}{RoboCasa365: 16 tasks / 16 cases}} & \InvTask{K4: Bowl in btm. drw. + close} & \InvEmb{Fixed Panda} & \multicolumn{2}{@{}c@{}}{\cellcolor{InvTest}\InvBenchmark{\InvLane}{RoboCasa365: 16 tasks / 16 cases}}\\
\InvTask{K9: Frying pan on cab. shelf} & \InvEmb{Fixed Panda} & \InvTask{CloseBlenderLid} & \InvEmb{Panda+Omron} & \InvTask{K6: YW mug in microw. + shut} & \InvEmb{Fixed Panda} & \InvTask{ArrangeBreadBasket} & \InvEmb{Panda+Omron}\\
\rowcolor{InvStripe}
\InvTask{L1: Ketchup in basket} & \InvEmb{Fixed Panda} & \InvTask{LoadDishwasher} & \InvEmb{Panda+Omron} & \InvTask{K8: Both moka pots on stove} & \InvEmb{Fixed Panda} & \InvTask{ArrangeTea} & \InvEmb{Panda+Omron}\\
\InvTask{L1: Tomato sauce in basket} & \InvEmb{Fixed Panda} & \InvTask{OpenDrawer} & \InvEmb{Panda+Omron} & \InvTask{L1: Soup + cheese in basket} & \InvEmb{Fixed Panda} & \InvTask{BreadSelection} & \InvEmb{Panda+Omron}\\
\rowcolor{InvStripe}
\InvTask{L2: Butter in basket} & \InvEmb{Fixed Panda} & \InvTask{PackIdenticalLunches} & \InvEmb{Panda+Omron} & \InvTask{L2: Soup + tomato in basket} & \InvEmb{Fixed Panda} & \InvTask{CategorizeCondiments} & \InvEmb{Panda+Omron}\\
\InvTask{L6: Pudding left of plate} & \InvEmb{Fixed Panda} & \InvTask{PickPlaceDrawerToCounter} & \InvEmb{Panda+Omron} & \InvTask{L2: Cheese + butter in basket} & \InvEmb{Fixed Panda} & \InvTask{CuttingToolSelection} & \InvEmb{Panda+Omron}\\
\rowcolor{InvStripe}
\InvTask{S1: Book in caddy (left)} & \InvEmb{Fixed Panda} & \InvTask{RinseSinkBasin} & \InvEmb{Panda+Omron} & \InvTask{L5: W/YW mugs on L/R plates} & \InvEmb{Fixed Panda} & \InvTask{GarnishPancake} & \InvEmb{Panda+Omron}\\
\InvTask{S3: Book in caddy (front)} & \InvEmb{Fixed Panda} & \InvTask{SlideDishwasherRack} & \InvEmb{Panda+Omron} & \InvTask{L6: Mug on plate; pudding right} & \InvEmb{Fixed Panda} & \InvTask{GatherTableware} & \InvEmb{Panda+Omron}\\
\rowcolor{InvStripe}
\InvTask{S3: Book in caddy (left)} & \InvEmb{Fixed Panda} & \InvTask{StoreLeftoversInBowl} & \InvEmb{Panda+Omron} & \InvTask{S1: Book in caddy (back)} & \InvEmb{Fixed Panda} & \InvTask{HeatKebabSandwich} & \InvEmb{Panda+Omron}\\
\multicolumn{2}{@{}c@{\hspace{\InvGap}}}{\cellcolor{InvTest}\InvBenchmark{\InvLane}{RoboTwin 2.0: 10 tasks / 40 cases}} & \InvTask{TurnOnElectricKettle} & \InvEmb{Panda+Omron} & \multicolumn{2}{@{}c@{\hspace{\InvGap}}}{\cellcolor{InvTest}\InvBenchmark{\InvLane}{RoboTwin 2.0: 10 tasks / 40 cases}} & \InvTask{MakeIceLemonade} & \InvEmb{Panda+Omron}\\
\rowcolor{InvStripe}
\InvTask{adjust\_bottle} & \InvEmb{AA DF DP UW} & \InvTask{WashLettuce} & \InvEmb{Panda+Omron} & \InvTask{blocks\_ranking\_size} & \InvEmb{AA DF DP UW} & \InvTask{PanTransfer} & \InvEmb{Panda+Omron}\\
\InvTask{move\_playingcard\_away} & \InvEmb{AA DF DP UW} & \InvTask{CoffeeSetupMug} & \InvEmb{Panda+Omron} & \InvTask{move\_pillbottle\_pad} & \InvEmb{AA DF DP UW} & \InvTask{PortionHotDogs} & \InvEmb{Panda+Omron}\\
\rowcolor{InvStripe}
\InvTask{open\_laptop} & \InvEmb{AA DF DP UW} & \InvTask{NavigateKitchen} & \InvEmb{Panda+Omron} & \InvTask{place\_a2b\_left} & \InvEmb{AA DF DP UW} & \InvTask{RecycleBottlesByType} & \InvEmb{Panda+Omron}\\
\InvTask{place\_bread\_skillet} & \InvEmb{AA DF DP UW} & \InvTask{OpenCabinet} & \InvEmb{Panda+Omron} & \InvTask{place\_empty\_cup} & \InvEmb{AA DF DP UW} & \InvTask{SeparateFreezerRack} & \InvEmb{Panda+Omron}\\
\rowcolor{InvStripe}
\InvTask{place\_container\_plate} & \InvEmb{AA DF DP UW} & \InvTask{DeliverStraw} & \InvEmb{Panda+Omron} & \InvTask{place\_fan} & \InvEmb{AA DF DP UW} & \InvTask{WaffleReheat} & \InvEmb{Panda+Omron}\\
\InvTask{place\_dual\_shoes} & \InvEmb{AA DF DP UW} & \InvTask{PrepareCoffee} & \InvEmb{Panda+Omron} & \InvTask{place\_mouse\_pad} & \InvEmb{AA DF DP UW} & \InvTask{WashFruitColander} & \InvEmb{Panda+Omron}\\
\rowcolor{InvStripe}
\InvTask{place\_shoe} & \InvEmb{AA DF DP UW} & \InvTask{SteamInMicrowave} & \InvEmb{Panda+Omron} & \InvTask{scan\_object} & \InvEmb{AA DF DP UW} & \InvTask{WeighIngredients} & \InvEmb{Panda+Omron}\\
\InvTask{rotate\_qrcode} & \InvEmb{AA DF DP UW} & \multicolumn{2}{@{}l@{\hspace{\InvGap}}}{\InvTotal{\InvLane}{36 tasks / 66 cases}} & \InvTask{stack\_bowls\_three} & \InvEmb{AA DF DP UW} & \multicolumn{2}{@{}l@{}}{\InvTotal{\InvLane}{36 tasks / 66 cases}}\\
\midrule[.4pt]
\multicolumn{8}{@{}l@{}}{\cellcolor{InvTransfer}\InvStageHeading{\S\ref{sec:exp-transfer} Transfer across Robots, Benchmarks, and Agent Models (frozen \ModelGPTAstra{} snapshot)}{63 tasks / 82 cases}}\\
\makebox[\InvTaskWidth][c]{\textbf{Task}} & \InvEmb{\textbf{Embodiment}} & \makebox[\InvTaskWidth][c]{\textbf{Task}} & \InvEmb{\textbf{Embodiment}} & \makebox[\InvTaskWidth][c]{\textbf{Task}} & \InvEmb{\textbf{Embodiment}} & \makebox[\InvTaskWidth][c]{\textbf{Task}} & \InvEmb{\textbf{Embodiment}}\\
\midrule[.25pt]
\rowcolor{InvStripe}
\multicolumn{2}{@{}c@{\hspace{\InvGap}}}{\cellcolor{InvTransfer}\InvBenchmark{\InvLane}{RoboTwin 2.0: 10 tasks / 10 cases}} & \InvTask{Wipe} & \InvEmb{Sawyer} & \InvTask{fill\_pen\_holder} & \InvEmb{dual ARX-X5} & \InvTask{pour\_balls\_into\_vase} & \InvEmb{dual ARX-X5}\\
\InvTask{blocks\_ranking\_size} & \InvEmb{dual ARX-X5} & \InvTask{ToolHang} & \InvEmb{Sawyer} & \InvTask{fold\_clothes} & \InvEmb{dual ARX-X5} & \InvTask{pour\_by\_language} & \InvEmb{dual ARX-X5}\\
\rowcolor{InvStripe}
\InvTask{move\_pillbottle\_pad} & \InvEmb{dual ARX-X5} & \multicolumn{2}{@{}c@{\hspace{\InvGap}}}{\cellcolor{InvTransfer}\InvBenchmark{\InvLane}{robosuite two-arm: 4 tasks / 16 cases}} & \InvTask{general\_pickup} & \InvEmb{dual ARX-X5} & \InvTask{pour\_liquid\_into\_cup} & \InvEmb{dual ARX-X5}\\
\InvTask{place\_a2b\_left} & \InvEmb{dual ARX-X5} & \InvTask{TwoArmLift} & \InvEmb{Baxter / Tiago} & \InvTask{hang\_mugs} & \InvEmb{dual ARX-X5} & \InvTask{press\_by\_number} & \InvEmb{dual ARX-X5}\\
\rowcolor{InvStripe}
\InvTask{place\_empty\_cup} & \InvEmb{dual ARX-X5} & \InvTask{TwoArmHandover} & \InvEmb{Baxter / Tiago} & \InvTask{imitate\_sorting\_sequence} & \InvEmb{dual ARX-X5} & \InvTask{push\_T} & \InvEmb{dual ARX-X5}\\
\InvTask{place\_fan} & \InvEmb{dual ARX-X5} & \InvTask{TwoArmTransport} & \InvEmb{Baxter / Tiago} & \InvTask{insert\_key} & \InvEmb{dual ARX-X5} & \InvTask{put\_bottles\_into\_dustbin} & \InvEmb{dual ARX-X5}\\
\rowcolor{InvStripe}
\InvTask{place\_mouse\_pad} & \InvEmb{dual ARX-X5} & \InvTask{TwoArmPegInHole} & \InvEmb{Baxter / Tiago} & \InvTask{insert\_tubes} & \InvEmb{dual ARX-X5} & \InvTask{solve\_equation} & \InvEmb{dual ARX-X5}\\
\InvTask{scan\_object} & \InvEmb{dual ARX-X5} & \multicolumn{2}{@{}c@{\hspace{\InvGap}}}{\cellcolor{InvTransfer}\InvBenchmark{\InvLane}{RoboDojo: 42 tasks / 42 cases}} & \InvTask{make\_kong} & \InvEmb{dual ARX-X5} & \InvTask{sort\_nesting\_dolls\_by\_size} & \InvEmb{dual ARX-X5}\\
\rowcolor{InvStripe}
\InvTask{stack\_bowls\_three} & \InvEmb{dual ARX-X5} & \InvTask{align\_blocks} & \InvEmb{dual ARX-X5} & \InvTask{make\_toast} & \InvEmb{dual ARX-X5} & \InvTask{stack\_blocks} & \InvEmb{dual ARX-X5}\\
\InvTask{stack\_bowls\_two} & \InvEmb{dual ARX-X5} & \InvTask{arrange\_largest\_number} & \InvEmb{dual ARX-X5} & \InvTask{match\_and\_pick\_from\_conveyor} & \InvEmb{dual ARX-X5} & \InvTask{stack\_blocks\_by\_language} & \InvEmb{dual ARX-X5}\\
\rowcolor{InvStripe}
\InvTask{turn\_switch} & \InvEmb{dual ARX-X5} & \InvTask{build\_tower} & \InvEmb{dual ARX-X5} & \InvTask{organize\_table} & \InvEmb{dual ARX-X5} & \InvTask{stack\_bowls} & \InvEmb{dual ARX-X5}\\
\multicolumn{2}{@{}c@{\hspace{\InvGap}}}{\cellcolor{InvTransfer}\InvBenchmark{\InvLane}{robosuite single-arm: 7 tasks / 14 cases}} & \InvTask{classify\_objects} & \InvEmb{dual ARX-X5} & \InvTask{pack\_objects\_into\_box} & \InvEmb{dual ARX-X5} & \InvTask{store\_laptop\_and\_headphones} & \InvEmb{dual ARX-X5}\\
\rowcolor{InvStripe}
\InvTask{Lift} & \InvEmb{Sawyer} & \InvTask{classify\_objects\_by\_language} & \InvEmb{dual ARX-X5} & \InvTask{pick\_from\_conveyor\_by\_image} & \InvEmb{dual ARX-X5} & \InvTask{store\_tools\_in\_toolbox} & \InvEmb{dual ARX-X5}\\
\InvTask{Stack} & \InvEmb{Sawyer} & \InvTask{cover\_blocks} & \InvEmb{dual ARX-X5} & \InvTask{play\_Xylophone} & \InvEmb{dual ARX-X5} & \InvTask{swap\_T} & \InvEmb{dual ARX-X5}\\
\rowcolor{InvStripe}
\InvTask{PickPlace} & \InvEmb{Sawyer} & \InvTask{deposit\_coin} & \InvEmb{dual ARX-X5} & \InvTask{play\_stacking\_toy} & \InvEmb{dual ARX-X5} & \InvTask{swap\_blocks} & \InvEmb{dual ARX-X5}\\
\InvTask{NutAssembly} & \InvEmb{Sawyer} & \InvTask{fasten\_screws} & \InvEmb{dual ARX-X5} & \InvTask{play\_tic\_tac\_toe} & \InvEmb{dual ARX-X5} & \InvTask{sweep\_blocks} & \InvEmb{dual ARX-X5}\\
\rowcolor{InvStripe}
\InvTask{Door} & \InvEmb{Sawyer} & \InvTask{fill\_egg\_holder} & \InvEmb{dual ARX-X5} & \InvTask{plug\_in\_charger} & \InvEmb{dual ARX-X5} & \multicolumn{2}{@{}l@{}}{\InvTotal{\InvLane}{63 tasks / 82 cases}}\\
\bottomrule[.4pt]\end{tabular}
\par\arrayrulecolor{black}\endgroup
\end{table*}

\clearpage
\begin{table*}[p]
\section{Generalization across Epochs}
\label{app:generalization}
\label{app:additional-results}
\noindent\begin{minipage}{\linewidth}\centering
\begingroup\color{black}
\captionsetup{font={small,color=black},labelfont={bf,color=black},
  hypcap=false,position=top,aboveskip=0pt,belowskip=6pt}
\newcommand{\DraftModel}[1]{\mbox{#1}}
\newcommand{\StepCount}[1]{#1}
\definecolor{StepBest}{HTML}{355F7D}
\newcommand{\BestSteps}[1]{\textcolor{StepBest}{\textbf{#1}}}
\newcommand{\EpochCell}[1]{\hspace*{-1.2pt}#1\hspace*{1.2pt}}
\newcommand{\GenHeader}[1]{{\fontsize{7.0}{8.6}\selectfont\bfseries\mbox{#1}}}
\newcommand{\GenSource}[1]{{\fontsize{6.8}{8.4}\selectfont\mbox{#1}}}
\newcommand{\GenMetric}[1]{{\fontsize{7.0}{8.6}\selectfont\mbox{#1}}}
\newlength{\GenModelWidth}
\newlength{\GenEpochWidth}
\newlength{\GenSRWidth}
\newlength{\GenStepsWidth}
\setlength{\GenModelWidth}{.160\linewidth}
\setlength{\GenEpochWidth}{.040\linewidth}
\setlength{\GenSRWidth}{.045\linewidth}
\setlength{\GenStepsWidth}{\dimexpr(\linewidth-\GenModelWidth-\GenEpochWidth)/7-\GenSRWidth\relax}
\captionof{table}{Generalization across epochs (five rollouts/case).
  SR (\%): full-task success. Steps: mean rollout length, including failures
  but excluding initialization and inference/network waits.
  LIBERO-PRO/RC365 (RoboCasa365): 20\,Hz control steps; RoboTwin: 250\,Hz physics steps.
  RoboTwin (40/10): learning/held-out task pools.
  Bold/blue bold: best SR/fewest Steps per model across epochs.}
\label{tab:generalization}
\fontsize{7.4}{9.1}\selectfont
\setlength{\tabcolsep}{0pt}
\renewcommand{\arraystretch}{1.10}
\setlength{\aboverulesep}{1.5pt}\setlength{\belowrulesep}{1.5pt}
\begin{tabular}{@{}
  >{\raggedright\arraybackslash}m{\GenModelWidth}
  >{\centering\arraybackslash}m{\GenEpochWidth}
  *{7}{>{\centering\arraybackslash}m{\GenSRWidth}
        >{\centering\arraybackslash}m{\GenStepsWidth}}@{}}
\toprule[.6pt]
\multicolumn{2}{@{}l}{} &
  \multicolumn{8}{c}{\textbf{(a) Unseen scenarios}} &
  \multicolumn{6}{c@{}}{\textbf{(b) Unseen tasks}}\\
\cmidrule(lr{3pt}){3-10}\cmidrule(l{3pt}){11-16}
\multicolumn{2}{@{}l}{\GenSource{Benchmark}} &
  \multicolumn{2}{c}{\GenHeader{LIBERO-90}} &
  \multicolumn{2}{c}{\GenHeader{RoboTwin (40)}} &
  \multicolumn{2}{c}{\GenHeader{RC365 Atomic}} &
  \multicolumn{2}{c}{\GenHeader{RC365 Composite}} &
  \multicolumn{2}{c}{\GenHeader{LIBERO-10}} &
  \multicolumn{2}{c}{\GenHeader{RoboTwin (10)}} &
  \multicolumn{2}{c@{}}{\GenHeader{RC365 Composite}}\\
\multicolumn{2}{@{}l}{\GenSource{Task / Rollout}} &
  \multicolumn{2}{c}{10 / 50} & \multicolumn{2}{c}{10 / 200} &
  \multicolumn{2}{c}{8 / 40} & \multicolumn{2}{c}{8 / 40} &
  \multicolumn{2}{c}{10 / 50} & \multicolumn{2}{c}{10 / 200} &
  \multicolumn{2}{c@{}}{16 / 80}\\
\cmidrule[.4pt]{1-16}
\GenMetric{Agent Model} & \EpochCell{\GenMetric{Epoch}} &
  \GenMetric{SR} & \GenMetric{Steps} & \GenMetric{SR} & \GenMetric{Steps} &
  \GenMetric{SR} & \GenMetric{Steps} & \GenMetric{SR} & \GenMetric{Steps} &
  \GenMetric{SR} & \GenMetric{Steps} & \GenMetric{SR} & \GenMetric{Steps} &
  \GenMetric{SR} & \GenMetric{Steps}\\
\midrule[.4pt]
\multirow{3}{=}{\DraftModel{GPT-6 Astra}} & \EpochCell{E1} & 88.0 & \BestSteps{308} & 77.5 & \StepCount{11,245} & 27.5 & \BestSteps{850} & 5.0 & \StepCount{4,490} & 66.0 & \StepCount{698} & 82.5 & \StepCount{9,864} & 30.00 & \StepCount{4,599}\\
 & \EpochCell{E2} & 84.0 & \StepCount{336} & 85.0 & \StepCount{11,101} & 35.0 & \StepCount{1,078} & 2.5 & \BestSteps{4,270} & 68.0 & \StepCount{673} & 87.5 & \StepCount{9,665} & 36.25 & \StepCount{4,168}\\
 & \EpochCell{E3} & \textbf{92.0} & \StepCount{312} & \textbf{90.0} & \BestSteps{9,220} & \textbf{47.5} & \StepCount{916} & \textbf{15.0} & \StepCount{4,429} & \textbf{80.0} & \BestSteps{550} & \textbf{88.5} & \BestSteps{7,609} & \textbf{43.75} & \BestSteps{3,902}\\
\midrule[.25pt]
\multirow{3}{=}{\DraftModel{Gemini 3.8 Flash}} & \EpochCell{E1} & 80.0 & \StepCount{398} & 43.5 & \StepCount{32,701} & 15.0 & \StepCount{1,291} & \textbf{2.5} & \StepCount{4,170} & 52.0 & \StepCount{787} & 56.5 & \StepCount{29,852} & \textbf{23.75} & \StepCount{4,121}\\
 & \EpochCell{E2} & \textbf{82.0} & \StepCount{357} & 48.0 & \BestSteps{16,290} & 27.5 & \BestSteps{1,163} & 0.0 & \StepCount{3,579} & \textbf{72.0} & \StepCount{668} & 58.0 & \StepCount{16,187} & 17.50 & \BestSteps{3,365}\\
 & \EpochCell{E3} & \textbf{82.0} & \BestSteps{334} & \textbf{54.5} & \StepCount{18,845} & \textbf{32.5} & \StepCount{1,168} & \textbf{2.5} & \BestSteps{3,439} & \textbf{72.0} & \BestSteps{664} & \textbf{68.0} & \BestSteps{12,772} & 21.25 & \StepCount{3,882}\\
\midrule[.25pt]
\multirow{3}{=}{\DraftModel{DeepSeek-V4.1-Flash}} & \EpochCell{E1} & 36.0 & \StepCount{547.1} & 25.0 & \StepCount{33,769} & 5.0 & \StepCount{1,272.5} & 0.0 & \StepCount{3,986.5} & 18.0 & \StepCount{858.4} & 32.5 & \StepCount{25,329} & 0.00 & \StepCount{4,567.6}\\
 & \EpochCell{E2} & 46.0 & \StepCount{563.6} & 31.0 & \StepCount{24,677} & 12.5 & \StepCount{1,204.0} & 0.0 & \BestSteps{3,643.8} & 32.0 & \StepCount{840.5} & 34.5 & \StepCount{22,732} & 0.00 & \BestSteps{3,380.0}\\
 & \EpochCell{E3} & \textbf{60.0} & \BestSteps{446.6} & \textbf{40.5} & \BestSteps{17,881} & \textbf{15.0} & \BestSteps{1,160.8} & 0.0 & \StepCount{3,990.5} & \textbf{50.0} & \BestSteps{693.6} & \textbf{43.5} & \BestSteps{15,920} & \textbf{6.25} & \StepCount{3,743.1}\\
\bottomrule[.6pt]
\end{tabular}
\endgroup
\end{minipage}
\par
\kern5.73pt
\section{Revision History}
\label{app:revision-history}
\begingroup
\centering
\captionsetup{font={small,color=black},labelfont={bf,color=black},hypcap=false,position=top,aboveskip=0pt,belowskip=6pt}
\captionof{table}{Revision history across three epochs. IDs encode epoch, batch, and within-batch revision. Ticks/crosses mark retained/rejected revisions; scores show before $\to$ after task success on the recorded subset.}
\label{tab:revision-history}
\begingroup
\definecolor{HistoryKeep}{HTML}{26734D}
\definecolor{HistoryDrop}{HTML}{B83232}
\newcommand{\HistoryKeep}{\textcolor{HistoryKeep}{$\checkmark$}}
\newcommand{\HistoryDrop}{\textcolor{HistoryDrop}{$\times$}}
\newcommand{\HistoryScore}[2]{\mbox{#1\kern.1pt$\to$\kern.1pt#2}}
\newcommand{\HistoryHeader}[1]{{\fontsize{5}{6.4}\selectfont\bfseries\mbox{#1}}}
\fontsize{5.2}{6.4}\selectfont
\setlength{\tabcolsep}{0pt}
\setlength{\parskip}{0pt}
\setlength{\aboverulesep}{.4pt}
\setlength{\belowrulesep}{.4pt}
\renewcommand{\arraystretch}{1.02}
\noindent
\begin{tabular}{@{}>{\raggedright\arraybackslash}p{20.5pt}@{\hspace{1pt}}>{\raggedright\arraybackslash}p{\dimexpr(\linewidth-20pt)/3-73.5pt\relax}@{\hspace{.5pt}}>{\raggedleft\arraybackslash}p{31.5pt}@{\hspace{1pt}}>{\centering\arraybackslash}p{19pt}@{\hspace{10pt}}>{\raggedright\arraybackslash}p{20.5pt}@{\hspace{1pt}}>{\raggedright\arraybackslash}p{\dimexpr(\linewidth-20pt)/3-73.5pt\relax}@{\hspace{.5pt}}>{\raggedleft\arraybackslash}p{31.5pt}@{\hspace{1pt}}>{\centering\arraybackslash}p{19pt}@{\hspace{10pt}}>{\raggedright\arraybackslash}p{20.5pt}@{\hspace{1pt}}>{\raggedright\arraybackslash}p{\dimexpr(\linewidth-20pt)/3-73.5pt\relax}@{\hspace{.5pt}}>{\raggedleft\arraybackslash}p{31.5pt}@{\hspace{1pt}}>{\centering\arraybackslash}p{19pt}@{}}
\toprule[.45pt]
\multicolumn{4}{@{}l@{\hspace{10pt}}}{\HistoryHeader{GPT-6 Astra}} & \multicolumn{4}{@{}l@{\hspace{10pt}}}{\HistoryHeader{Gemini 3.8 Flash}} & \multicolumn{4}{@{}l@{}}{\HistoryHeader{DeepSeek-V4.1-Flash}}\\
\midrule[.3pt]
\HistoryHeader{ID} & \HistoryHeader{Modified content} & \HistoryHeader{Success rate} & \HistoryHeader{Retain} & \HistoryHeader{ID} & \HistoryHeader{Modified content} & \HistoryHeader{Success rate} & \HistoryHeader{Retain} & \HistoryHeader{ID} & \HistoryHeader{Modified content} & \HistoryHeader{Success rate} & \HistoryHeader{Retain}\\
\midrule[.3pt]
EP1-1-1 & \mbox{Align approach and jaw axes} & \HistoryScore{6/22}{7/22} & \HistoryKeep & EP1-1-1 & \mbox{Pre-action object disambiguation} & \HistoryScore{8/10}{8/10} & \HistoryDrop & EP1-1-1 & \mbox{Place using held-object geometry} & \HistoryScore{5/8}{5/8} & \HistoryDrop\\
EP1-1-2 & \mbox{Seed an RGB-D mask for the target} & \HistoryScore{4/12}{7/12} & \HistoryKeep & EP1-1-2 & \mbox{Recheck identity after transfer} & \HistoryScore{6/10}{9/10} & \HistoryKeep & EP1-1-2 & \mbox{Stop arm moves on measured progress} & \HistoryScore{4/8}{6/8} & \HistoryKeep\\
\addlinespace[.3pt]
EP1-2-1 & \mbox{Seeded masks, default grasp planner} & \HistoryScore{2/8}{2/8} & \HistoryDrop & EP1-2-1 & \mbox{Measured frontal grasp and pull} & \HistoryScore{2/8}{5/8} & \HistoryKeep & EP1-2-1 & \mbox{Select targets in image coordinates} & \HistoryScore{5/8}{5/8} & \HistoryDrop\\
\addlinespace[.3pt]
EP1-3-1 & \mbox{First whole-body placement attempt} & \HistoryScore{8/16}{8/16} & \HistoryDrop & EP1-2-2 & \mbox{Level carry with holding checks} & \HistoryScore{6/10}{4/10} & \HistoryDrop & EP1-2-2 & \mbox{Align grasp yaw with tool frames} & \HistoryScore{8/8}{8/8} & \HistoryDrop\\
\addlinespace[.3pt]
EP1-3-2 & \mbox{Choose grasp or push at contact} & \HistoryScore{5/8}{5/8} & \HistoryDrop & EP1-3-1 & \mbox{Segmented carry and routed placement} & \HistoryScore{6/10}{7/10} & \HistoryKeep & EP1-3-1 & \mbox{Add carry clearance before release} & \HistoryScore{8/8}{8/8} & \HistoryDrop\\
\addlinespace[.3pt]
EP1-4-1 & \mbox{First-contact following paths} & \HistoryScore{8/16}{9/16} & \HistoryDrop & EP1-4-1 & \mbox{Remove all time-based segmentation} & \HistoryScore{14/23}{9/23} & \HistoryDrop & EP1-3-2 & \mbox{Guard grasp insertion before closure} & \HistoryScore{1/4}{2/4} & \HistoryKeep\\
\addlinespace[.3pt]
EP1-4-2 & \mbox{Transfer along the contact path} & \HistoryScore{7/11}{6/11} & \HistoryDrop & EP1-4-2 & \mbox{Example-heavy instructions} & \HistoryScore{24/42}{23/42} & \HistoryDrop & EP1-4-1 & \mbox{Grip checks owned by each phase} & \HistoryScore{30/54}{26/54} & \HistoryDrop\\
\addlinespace[.3pt]
EP1-5-1 & \mbox{Grasp boxes from rotated XY views} & \HistoryScore{8/15}{8/15} & \HistoryDrop & EP1-5-1 & \mbox{First joint-space path planning} & \HistoryScore{3/3}{3/3} & \HistoryDrop & EP1-5-1 & \mbox{One early-exit rule for every stall} & \HistoryScore{9/18}{12/18} & \HistoryDrop\\
EP1-5-2 & \mbox{Grasp directions from surface axes} & \HistoryScore{5/8}{5/8} & \HistoryDrop & EP1-6-1 & \mbox{Shared execution for both grasp types} & \HistoryScore{23/42}{23/42} & \HistoryKeep & EP1-5-2 & \mbox{Promptable masks for the target} & \HistoryScore{2/6}{3/6} & \HistoryKeep\\
\addlinespace[.3pt]
EP1-6-1 & \mbox{Re-measure the grasp from a near view} & \HistoryScore{9/16}{9/16} & \HistoryDrop &  &  &  &  & EP1-6-1 & \mbox{Rank grasp recovery goals by geometry} & \HistoryScore{3/8}{3/8} & \HistoryDrop\\
\midrule[.25pt]
EP2-1-1 & \mbox{Standoff hold with progress feedback} & \HistoryScore{0/8}{0/8} & \HistoryDrop & EP2-1-1 & \mbox{Seed an RGB-D foreground mask} & \HistoryScore{3/24}{6/24} & \HistoryKeep & EP2-1-1 & \mbox{Separate grasp acquisition from lifting} & \HistoryScore{5/13}{7/13} & \HistoryKeep\\
\addlinespace[.3pt]
EP2-2-1 & \mbox{Align the mobile-base control point} & \HistoryScore{0/11}{0/11} & \HistoryDrop & EP2-2-1 & \mbox{Obstacle-aware base routing} & \HistoryScore{6/24}{8/24} & \HistoryKeep & EP2-2-1 & \mbox{Seed foreground for directional grasps} & \HistoryScore{1/3}{1/3} & \HistoryDrop\\
\addlinespace[.3pt]
EP2-2-2 & \mbox{Heading from observation only} & \HistoryScore{0/8}{0/8} & \HistoryDrop & EP2-2-2 & \mbox{Explicit approach and jaw axes} & \HistoryScore{12/30}{14/30} & \HistoryKeep & EP2-3-1 & \mbox{Refine geometry near foreground seeds} & \HistoryScore{6/9}{6/9} & \HistoryDrop\\
\addlinespace[.3pt]
EP2-3-1 & \mbox{First routes around visible obstacles} & \HistoryScore{0/8}{0/8} & \HistoryDrop & EP2-3-1 & \mbox{Minimum-rotation staged pressing} & \HistoryScore{12/30}{13/30} & \HistoryKeep & EP2-4-1 & \mbox{Pull along estimated hinge paths} & \HistoryScore{4/6}{3/6} & \HistoryDrop\\
EP2-4-1 & \mbox{Obstacle-aware routing with docking} & \HistoryScore{2/8}{2/8} & \HistoryKeep & EP2-3-2 & \mbox{Feedback-adaptive contact pulling} & \HistoryScore{7/10}{4/10} & \HistoryDrop & EP2-4-2 & \mbox{Choose operating stations from reach} & \HistoryScore{4/11}{6/11} & \HistoryDrop\\
\addlinespace[.3pt]
EP2-5-1 & \mbox{Stop arm motion when progress stalls} & \HistoryScore{5/16}{8/16} & \HistoryKeep & EP2-4-1 & \mbox{Refine side-sector stations} & \HistoryScore{5/10}{5/10} & \HistoryDrop & EP2-5-1 & \mbox{Separate planes in RGB-D} & \HistoryScore{4/11}{5/11} & \HistoryKeep\\
EP2-5-2 & \mbox{Adaptive pulling under slow feedback} & \HistoryScore{5/8}{4/8} & \HistoryDrop & EP2-4-2 & \mbox{Close the gripper only at contact} & \HistoryScore{7/12}{7/12} & \HistoryDrop & EP2-5-2 & \mbox{Transport a retained load explicitly} & \HistoryScore{20/36}{14/36} & \HistoryDrop\\
\addlinespace[.3pt]
EP2-6-1 & \mbox{First grip-width load checks} & \HistoryScore{4/8}{4/8} & \HistoryDrop & EP2-5-1 & \mbox{Staged recovery of the approach} & \HistoryScore{3/8}{3/8} & \HistoryDrop & EP2-6-1 & \mbox{Preserve the native arrival result} & \HistoryScore{30/54}{30/54} & \HistoryKeep\\
\addlinespace[.3pt]
EP2-6-2 & \mbox{Joint search on grasp and station} & \HistoryScore{5/10}{3/10} & \HistoryDrop & EP2-6-1 & \mbox{Calibrated grasp proposals} & \HistoryScore{2/8}{4/8} & \HistoryKeep & EP2-6-2 & \mbox{Offer learned and geometric options} & \HistoryScore{23/42}{23/42} & \HistoryKeep\\
 &  &  &  & EP2-6-2 & \mbox{Move the wrist camera closer} & \HistoryScore{6/10}{6/10} & \HistoryDrop &  &  &  & \\
\midrule[.25pt]
EP3-1-1 & \mbox{Grasp candidates from GraspGenX} & \HistoryScore{5/8}{7/8} & \HistoryKeep & EP3-1-1 & \mbox{Body-size and grip-offset placement} & \HistoryScore{12/18}{10/18} & \HistoryDrop & EP3-1-1 & \mbox{Seed an RGB-D foreground mask} & \HistoryScore{8/26}{11/26} & \HistoryKeep\\
\addlinespace[.3pt]
EP3-1-2 & \mbox{Gate motions on IK and clearance} & \HistoryScore{22/42}{25/42} & \HistoryKeep & EP3-1-2 & \mbox{Track the held body through transport} & \HistoryScore{2/7}{4/7} & \HistoryKeep & EP3-2-1 & \mbox{GraspGenX-assisted contact recovery} & \HistoryScore{1/13}{4/13} & \HistoryKeep\\
\addlinespace[.3pt]
EP3-2-1 & \mbox{Feedback-adaptive contact pulling} & \HistoryScore{5/8}{6/8} & \HistoryKeep & EP3-2-1 & \mbox{Use an explicit entry direction} & \HistoryScore{2/12}{5/12} & \HistoryKeep & EP3-2-2 & \mbox{Approach contact along measured rays} & \HistoryScore{3/10}{4/10} & \HistoryDrop\\
\addlinespace[.3pt]
EP3-2-2 & \mbox{Body-aware transfer with load checks} & \HistoryScore{4/8}{5/8} & \HistoryKeep & EP3-2-2 & \mbox{Observed-pivot turning with feedback} & \HistoryScore{12/18}{12/18} & \HistoryDrop & EP3-3-1 & \mbox{SAM2 part measurement and contact} & \HistoryScore{1/14}{3/14} & \HistoryKeep\\
\addlinespace[.3pt]
EP3-3-1 & \mbox{Guarded incremental pushing} & \HistoryScore{6/8}{6/8} & \HistoryDrop & EP3-3-1 & \mbox{Oriented RGB-D regions for grasping} & \HistoryScore{12/18}{10/18} & \HistoryDrop & EP3-3-2 & \mbox{Refresh geometry before grasping} & \HistoryScore{4/10}{3/10} & \HistoryDrop\\
\addlinespace[.3pt]
EP3-3-2 & \mbox{Turn about the observed pivot} & \HistoryScore{8/12}{9/12} & \HistoryKeep & EP3-4-1 & \mbox{Measured grasp enclosure pre-closure} & \HistoryScore{2/9}{4/9} & \HistoryKeep & EP3-4-1 & \mbox{Resume contact from measured feedback} & \HistoryScore{5/14}{5/14} & \HistoryDrop\\
\addlinespace[.3pt]
EP3-4-1 & \mbox{Remove the generic load-abort rule} & \HistoryScore{30/54}{28/54} & \HistoryDrop & EP3-4-2 & \mbox{Gate motions on IK and clearance} & \HistoryScore{5/12}{5/12} & \HistoryDrop & EP3-4-2 & \mbox{Recover explicit-direction pregrasps} & \HistoryScore{5/13}{5/13} & \HistoryDrop\\
\addlinespace[.3pt]
EP3-4-2 & \mbox{Agent-chosen stall feedback} & \HistoryScore{23/42}{24/42} & \HistoryKeep & EP3-5-1 & \mbox{Body-referenced placement, re-observed} & \HistoryScore{4/9}{3/9} & \HistoryDrop & EP3-5-1 & \mbox{Collision check before each motion} & \HistoryScore{23/42}{28/42} & \HistoryKeep\\
\addlinespace[.3pt]
EP3-5-1 & \mbox{Per-phase stopping rules} & \HistoryScore{12/18}{9/18} & \HistoryDrop & EP3-5-2 & \mbox{YOLOE detection with SAM2} & \HistoryScore{4/7}{3/7} & \HistoryDrop & EP3-6-1 & \mbox{Place using carried-body geometry} & \HistoryScore{4/11}{3/11} & \HistoryDrop\\
\addlinespace[.3pt]
EP3-5-2 & \mbox{Calibrate opening with feedback} & \HistoryScore{0/9}{1/9} & \HistoryKeep & EP3-6-1 & \mbox{Explicit-direction pregrasp recovery} & \HistoryScore{7/12}{6/12} & \HistoryDrop &  &  &  & \\
\addlinespace[.3pt]
EP3-6-1 & \mbox{Partial insertion followed by lift} & \HistoryScore{14/33}{14/33} & \HistoryDrop & EP3-6-2 & \mbox{Generator-only grasp variant} & \HistoryScore{27/42}{23/42} & \HistoryDrop &  &  &  & \\
EP3-6-2 & \mbox{Re-inspect the grasp before closing} & \HistoryScore{6/12}{9/12} & \HistoryKeep &  &  &  &  &  &  &  & \\
\bottomrule[.45pt]
\end{tabular}
\endgroup
\par\endgroup
\end{table*}

\clearpage
\section{Persistent Failures across Revisions}
\label{app:persistent-failures}
\label{app:composite-task-failures}

We inspect programs, feedback, and images from failed \ModelGPTAstra{}
rollouts. At Epoch~3, Composite-Seen success is 15\% (6/40).
\FunctionName{RinseSinkBasin} and \FunctionName{SteamInMicrowave} remain
at 0/5 in all three epochs. P$n$ indexes programs.

\begingroup
\definecolor{ETRule}{HTML}{CBD0D4}
\definecolor{ETMuted}{HTML}{49525A}
\definecolor{ETCodeBackground}{HTML}{F4F5F6}
\lstdefinestyle{trajectory}{
  style=unicraft,
  basicstyle=\fontfamily{zi4}\mdseries\upshape\fontsize{7}{8}\selectfont\color{UcCodeInk},
  numbers=none,breaklines=true,breakatwhitespace=true,breakindent=8pt,
  backgroundcolor=\color{ETCodeBackground},
  morekeywords={pick,place,place_beside,grasp,lift,lift_both,navigate,
    adjust_height,align_base,survey,move_to,move_tcps,observe,segment_region,check_motion,set_gripper},
  aboveskip=0pt,belowskip=0pt
}
\newenvironment{ETCase}[3][]{%
  \begin{tcolorbox}[enhanced jigsaw,breakable,frame hidden,boxrule=0pt,
    colback=white,colbacktitle=white,coltitle=black,
    boxsep=0pt,left=0pt,right=0pt,top=0pt,bottom=0pt,
    before skip=0pt,after skip=4pt,pad at break*=1pt,
    fonttitle=\fontsize{8}{9}\selectfont\bfseries,
    title after break={#2 (continued)},#1]
  \fontsize{8}{9}\selectfont
  \captionsetup{font=footnotesize,labelfont=bf,justification=raggedright,
    singlelinecheck=false,hypcap=false,aboveskip=3pt,belowskip=0pt}
}{\end{tcolorbox}}
\newcommand{\ETHeading}[2]{%
  {\color{ETRule}\hrule height .35pt}\vspace{2pt}
  \noindent\begin{minipage}[t]{.81\linewidth}
    \fontsize{9}{10}\selectfont\bfseries #1
  \end{minipage}\hfill
  \begin{minipage}[t]{.17\linewidth}\raggedleft
    \fontsize{8}{9}\selectfont\bfseries\textsc{#2}
  \end{minipage}\par}
\newcommand{\ETMeta}[1]{%
  \vspace{1pt}{\fontsize{7.5}{8.5}\selectfont\color{ETMuted}#1\par}\vspace{2pt}}
\newcommand{\ETView}[3]{%
  \begin{minipage}[b]{#1\linewidth}
  \centering
  {\fontsize{7}{8}\selectfont\color{ETMuted}#2\par}\vspace{1pt}
  \makebox[\linewidth][c]{\includegraphics[page=#3,width=\linewidth,height=47pt,keepaspectratio]{figures.pdf}}%
  \end{minipage}}
\definecolor{PFFindingBackground}{HTML}{EDF4F8}
\definecolor{PFFindingAccent}{HTML}{35677C}
\newtcolorbox{PFFinding}[1]{%
  enhanced,unbreakable,frame hidden,boxrule=0pt,sharp corners,
  colback=PFFindingBackground,
  borderline west={2pt}{0pt}{PFFindingAccent},
  boxsep=0pt,left=7pt,right=7pt,top=5pt,bottom=6pt,
  before skip=5pt,after skip=0pt,
  fontupper=\fontsize{9}{10.5}\selectfont,
  before upper={\textcolor{PFFindingAccent}{\textbf{#1}}\par\vspace{2pt}}}

\noindent\begin{minipage}{\linewidth}
\noindent\textbf{(i) Missing control capabilities.}\par
\noindent Earlier repositories expose arm motion but no body-height control.
In \FunctionName{WashLettuce}, an upward hand movement fails to reach
its target. Raising the arm's support from 0.026 to 0.100~m in a matched
replay restores arrival and reduces position error from 44.19 to 9.95~mm.
\par\smallskip
\noindent\begin{minipage}{\linewidth}
\begin{ETCase}[unbreakable]{A clearance move without body elevation}{Blocked}
\noindent\begin{minipage}{\linewidth}
\ETHeading{A clearance move without body elevation}{Blocked}
\ETMeta{\ModelGPTAstra{} $\cdot$ Panda-Omron $\cdot$ rollout 4}
\noindent \ETView{.316}{Repositioned base}{7}\hfill\ETView{.316}{Revised approach}{8}\hfill\ETView{.316}{Clearance attempt}{9}\par\vspace{1pt}
\end{minipage}\par
\begin{lstlisting}[style=trajectory,xleftmargin=3pt,xrightmargin=3pt]
c = connect(); p = c.robot.state.effectors['arm'].pose
print(c.robot.move_to(Pose((2.65, -2.1, 1.55), p.quaternion)))
\end{lstlisting}
\captionof{figure}{\FunctionName{WashLettuce}. P27 attempts an upward clearance move and reports that the requested hand pose was not reached. These scene views show the original rollout, before the separate height intervention.}\label{fig:failure-capability-wash}
\end{ETCase}
\end{minipage}\par\smallskip
\end{minipage}\par\smallskip

\noindent\begin{minipage}{\linewidth}
\FunctionName{PackIdenticalLunches} requires a fish and a lemon in each
container. Picking the second fish stops at the approach stage.
Raising the arm's support from 0.001 to 0.300~m in a matched replay
restores arrival and reduces error from 14.66 to 9.95~mm. Both pairs keep
the preceding commands, target, budget, and arrival thresholds fixed.
These replays establish recovery of the selected movements only.
\par\smallskip
\noindent\begin{minipage}{\linewidth}
\begin{ETCase}[unbreakable]{Approach stops before grasping the fish}{Blocked}
\noindent\begin{minipage}{\linewidth}
\ETHeading{Approach stops before grasping the fish}{Blocked}
\ETMeta{\ModelGPTAstra{} $\cdot$ Panda-Omron $\cdot$ rollout 3}
\noindent \ETView{.316}{Target view}{10}\hfill\ETView{.316}{Approach stopped}{11}\hfill\ETView{.316}{Scene recheck}{12}\par\vspace{1pt}
\end{minipage}\par
\begin{lstlisting}[style=trajectory,xleftmargin=3pt,xrightmargin=3pt]
c = connect(); cam = c.cameras['scene']
o = cam.segment_region((555, 429, 670, 552), seed=(612, 480),
    save_path='snapshots/fish2_part')
print(o)
print(c.robot.pick(o, clearance=0.06, depth=0.018))
\end{lstlisting}
\captionof{figure}{\FunctionName{PackIdenticalLunches}. P39 stops at the approach stage of \FunctionName{pick()}, before grasping the fish. The source rollout later times out. Views are scene, wrist, and scene, before the separate height intervention.}\label{fig:failure-capability-pack}
\end{ETCase}
\end{minipage}\par\smallskip
\begin{PFFinding}{Finding 1. Delayed capability diagnosis.}
Repeated revisions can refine existing routines while leaving required
skills and controls absent. This pattern suggests delayed diagnosis and
acquisition of missing capabilities, allowing functional gaps to persist.
\end{PFFinding}
\end{minipage}\par\smallskip

\newpage

\noindent\begin{minipage}{\linewidth}
\noindent\textbf{(ii) Later actions undo completed subgoals.}\par
\noindent\FunctionName{StoreLeftoversInBowl} asks the robot to put food in a bowl
and store the filled bowl in the fridge. All five final rollout logs
describe food in the bowl followed by the bowl falling during handling.
Here, the hand reaches its target near the shelf, but the gripper no
longer holds the bowl.
\par\smallskip
\noindent\begin{minipage}{\linewidth}
\begin{ETCase}[unbreakable]{The hand arrives, but the bowl is lost}{Failure}
\noindent\begin{minipage}{\linewidth}
\ETHeading{The hand arrives, but the bowl is lost}{Failure}
\ETMeta{\ModelGPTAstra{} $\cdot$ Panda-Omron $\cdot$ rollout 1}
\noindent \ETView{.316}{Loaded bowl}{13}\hfill\ETView{.316}{Shelf approach}{14}\hfill\ETView{.316}{After loss}{15}\par\vspace{1pt}
\end{minipage}\par
\begin{lstlisting}[style=trajectory,xleftmargin=3pt,xrightmargin=3pt]
c = connect(); r = c.robot
p = r.state.effectors['arm'].pose
print(r.move_to(Pose((4.1, -0.7, 1.18), p.quaternion),
    stop_on_stall=True))
print(r.state)
print(c.cameras['side'].observe(
    save_path='snapshots/shelf_entry_side').status)
print(c.cameras['scene'].observe(
    save_path='snapshots/shelf_entry').status)
\end{lstlisting}
\captionof{figure}{\FunctionName{StoreLeftoversInBowl}. P39 reaches the requested hand pose, then reports that the gripper holds no object. Later images show the bowl and food below the shelf. Views are wrist, scene, and wrist.}\label{fig:failure-composition}
\end{ETCase}
\end{minipage}\par\smallskip
\end{minipage}\par\smallskip

\noindent\begin{minipage}{\linewidth}
\FunctionName{SteamInMicrowave} asks the robot to put an onion in a bowl,
place the bowl in the microwave, then close the door and switch it on.
Three of five final rollouts report placing the onion in the bowl, then
losing the bowl during transfer or insertion. The pictured insertion
separates them: the onion reaches the turntable and the bowl drops to the floor.
\par\smallskip
\noindent\begin{minipage}{\linewidth}
\begin{ETCase}[unbreakable]{The onion reaches the microwave alone}{Failure}
\noindent\begin{minipage}{\linewidth}
\ETHeading{The onion reaches the microwave alone}{Failure}
\ETMeta{\ModelGPTAstra{} $\cdot$ Panda-Omron $\cdot$ rollout 1}
\noindent \ETView{.316}{Loaded bowl}{16}\hfill\ETView{.316}{Onion alone}{17}\hfill\ETView{.316}{Bowl on floor}{18}\par\vspace{1pt}
\end{minipage}\par
\begin{lstlisting}[style=trajectory,xleftmargin=3pt,xrightmargin=3pt]
c = connect(); d = load_artifact('artifacts/insert.json')
print(c.robot.move_to(d['target'], stop_on_stall=True))
print(c.robot.state)
print(c.cameras['side'].observe(
    save_path='snapshots/insert_progress').status)
\end{lstlisting}
\captionof{figure}{\FunctionName{SteamInMicrowave}. Before insertion, the onion is inside the held bowl. P44 loses the grip. The onion ends up on the turntable and the bowl on the floor. Views are wrist, scene, and scene.}\label{fig:failure-steam}
\end{ETCase}
\end{minipage}\par\smallskip
\begin{PFFinding}{Finding 2. Task progress is not monotonic.}
Completed subgoals can be undone by later actions, so task progress is not
monotonic. A hand may reach its target while dropping a filled bowl.
Robot-use agents therefore need to preserve achieved subgoals and restore
them when disrupted, especially in long-horizon tasks involving multiple
subtasks.
\end{PFFinding}
\end{minipage}\par\smallskip

\newpage

\noindent\begin{minipage}{\linewidth}
\noindent\textbf{(iii) Understanding movable parts and action effects.}\par
\noindent\FunctionName{RinseSinkBasin} requires opening the valve and directing
running water across three regions of the sink. In the selected attempt,
the gripper reports holding an object, but the requested spout turn
produces almost no hand movement. Later attempts still leave coverage
incomplete. Grasp and motion feedback alone cannot confirm that water
is running or reaching the required regions.
\par\smallskip
\noindent\begin{minipage}{\linewidth}
\begin{ETCase}[unbreakable]{A requested turn makes little progress}{Failure}
\noindent\begin{minipage}{\linewidth}
\ETHeading{A requested turn makes little progress}{Failure}
\ETMeta{\ModelGPTAstra{} $\cdot$ Panda-Omron $\cdot$ rollout 3}
\noindent \ETView{.316}{Valve attempt}{19}\hfill\ETView{.316}{Valve revisited}{20}\hfill\ETView{.316}{Spout sweep}{21}\par\vspace{1pt}
\end{minipage}\par
\begin{lstlisting}[style=trajectory,xleftmargin=3pt,xrightmargin=3pt]
c = connect(); r = c.robot
print(r.turn_about((3.049, -0.12, 1.1), (0, 0, 1), -0.65,
    step=0.12, rotate_tool=True))
print(c.cameras['wrist'].observe(
    save_path='snapshots/sweep_left').status)
\end{lstlisting}
\captionof{figure}{\FunctionName{RinseSinkBasin}. P58 requests a 0.65-rad spout rotation. The hand\textquotesingle s measured angular travel is $-0.0077$ rad despite detected holding. Later hand movements leave basin coverage incomplete.}\label{fig:failure-mechanism}
\end{ETCase}
\end{minipage}\par\smallskip
\end{minipage}\par\smallskip

\noindent\begin{minipage}{\linewidth}
In the frozen-repository \FunctionName{Door} transfer case, the robot
attempts to turn the latch before pulling the door. The hand completes
most of the requested turn, but whether the latch has actually released
remains uncertain. The following pull advances only 7.2 of 80~mm, and
the door stays closed.
\par\smallskip
\noindent\begin{minipage}{\linewidth}
\begin{ETCase}[unbreakable]{A hand turn leaves the door closed}{Failure}
\noindent\begin{minipage}{\linewidth}
\ETHeading{A hand turn leaves the door closed}{Failure}
\ETMeta{\ModelGPTAstra{} $\cdot$ Sawyer $\cdot$ rollout 1}
\noindent \ETView{.316}{Grasp detected}{22}\hfill\ETView{.316}{After the turn}{23}\hfill\ETView{.316}{After the pull}{24}\par\vspace{1pt}
\end{minipage}\par
\begin{lstlisting}[style=trajectory,xleftmargin=3pt,xrightmargin=3pt]
c = connect()
print(c.robot.pull((0, 1, 0), distance=0.08, step=0.02, axis=(0,
    0, 1), rotate_tool=True))
print(c.robot.state)
print(c.cameras['side'].observe(
    save_path='snapshots/open1').status)
\end{lstlisting}
\captionof{figure}{\FunctionName{Door}. P16 reports 0.428 of 0.450 rad of hand rotation without confirming the latch state. P17 then pulls only 7.2 of 80 mm. The door remains closed in all three side views.}\label{fig:failure-door}
\end{ETCase}
\end{minipage}\par\smallskip
\begin{PFFinding}{Finding 3. Understanding action effects.}
Understanding how movable parts respond remains a bottleneck despite
reusable skills. Grasping or turning feedback alone cannot
confirm an open valve or a released latch.
\end{PFFinding}
\end{minipage}\par\smallskip

\newpage

\noindent\begin{minipage}{\linewidth}
\noindent\textbf{(iv) Fine manipulation under contact constraints.}\par
\noindent The LIBERO microwave task requires placing the mug inside the appliance
and closing the door. The robot lifts the mug successfully, but moving
it through the opening also requires adequate clearance and a stable
grasp. The mug tilts and slips during insertion.
\par\smallskip
\noindent\begin{minipage}{\linewidth}
\begin{ETCase}[unbreakable]{The lifted mug slips during insertion}{Failure}
\noindent\begin{minipage}{\linewidth}
\ETHeading{The lifted mug slips during insertion}{Failure}
\ETMeta{\ModelGPTAstra{} $\cdot$ Panda $\cdot$ rollout 1}
\noindent \ETView{.316}{Lifted mug}{25}\hfill\ETView{.316}{At the opening}{26}\hfill\ETView{.316}{Grip lost}{27}\par\vspace{1pt}
\end{minipage}\par
\begin{lstlisting}[style=trajectory,xleftmargin=3pt,xrightmargin=3pt]
c = connect(); r = c.robot
q = (0, 0.8660254, 0.5, 0)
print(r.move_to(Pose((-0.025, 0.275, 1.012), q),
    stop_on_stall=True))
print(r.state)
\end{lstlisting}
\captionof{figure}{\textbf{LIBERO microwave insertion.} The mug remains held after lifting. P25 reports lost holding contact as the hand enters the opening, and the mug tilts. Later attempts exhaust the horizon.}\label{fig:failure-contact}
\end{ETCase}
\end{minipage}\par\smallskip
\end{minipage}\par\smallskip

\noindent\begin{minipage}{\linewidth}
In the frozen-repository \FunctionName{ToolHang} transfer case, the robot
must assemble a frame and hang a tool from it by its ring.
Bringing the ring near the support is only an approach step. Successful
hanging requires the ring to engage the support and carry the tool after
release. In the recorded attempt, the gripper opens and withdraws, but
the tool stays nearly horizontal against the frame and the task fails.
\par\smallskip
\noindent\begin{minipage}{\linewidth}
\begin{ETCase}[unbreakable]{Release does not produce a hanging tool}{Failure}
\noindent\begin{minipage}{\linewidth}
\ETHeading{Release does not produce a hanging tool}{Failure}
\ETMeta{\ModelGPTAstra{} $\cdot$ Sawyer $\cdot$ rollout 1}
\noindent \ETView{.316}{Frame assembled}{28}\hfill\ETView{.316}{Ring alignment}{29}\hfill\ETView{.316}{After release}{30}\par\vspace{1pt}
\end{minipage}\par
\begin{lstlisting}[style=trajectory,xleftmargin=3pt,xrightmargin=3pt]
c = connect(); r = c.robot
p = r.state.pose
print(r.set_gripper(1))
v = R.from_quat(p.quaternion).apply([0, 0, 1])
print(r.move_to(Pose(tuple(np.array(p.position) - v * 0.12),
    p.quaternion), stop_on_stall=True))
\end{lstlisting}
\captionof{figure}{\FunctionName{ToolHang}. P52 opens the gripper and withdraws from the ring. The tool remains nearly horizontal against the frame, and the native hanging criterion is not met. Views are scene, wrist, and side.}\label{fig:failure-toolhang}
\end{ETCase}
\end{minipage}\par\smallskip
\begin{PFFinding}{Finding 4. Fine manipulation remains unreliable.}
Successful grasping and coarse positioning do not guarantee reliable
fine manipulation. Insertion and release impose tighter geometric and
contact constraints, so skills must preserve the intended grasp or
support throughout execution.
\end{PFFinding}
\end{minipage}

\endgroup
\label{app:persistent-failures-end}

\clearpage
\section{The Coverage--Context Trade-off}
\label{app:revision-tradeoff}

\begingroup
\definecolor{ETRule}{HTML}{CBD0D4}
\definecolor{ETMuted}{HTML}{49525A}
\definecolor{ETCodeBackground}{HTML}{F4F5F6}
\lstdefinestyle{trajectory}{
  style=unicraft,
  basicstyle=\fontfamily{zi4}\mdseries\upshape\fontsize{7}{8}\selectfont\color{UcCodeInk},
  numbers=none,breaklines=true,breakatwhitespace=true,breakindent=8pt,
  backgroundcolor=\color{ETCodeBackground},
  morekeywords={pick,place,place_beside,grasp,lift,lift_both,navigate,
    adjust_height,align_base,survey,move_to,move_tcps,observe,segment_region,check_motion,set_gripper},
  aboveskip=0pt,belowskip=0pt
}
\newenvironment{ETCase}[3][]{%
  \begin{tcolorbox}[enhanced jigsaw,breakable,frame hidden,boxrule=0pt,
    colback=white,colbacktitle=white,coltitle=black,
    boxsep=0pt,left=0pt,right=0pt,top=0pt,bottom=0pt,
    before skip=0pt,after skip=4pt,pad at break*=1pt,
    fonttitle=\fontsize{8}{9}\selectfont\bfseries,
    title after break={#2 (continued)},#1]
  \fontsize{8}{9}\selectfont
  \captionsetup{font=footnotesize,labelfont=bf,justification=raggedright,
    singlelinecheck=false,hypcap=false,aboveskip=3pt,belowskip=0pt}
}{\end{tcolorbox}}
\newcommand{\ETHeading}[2]{%
  {\color{ETRule}\hrule height .35pt}\vspace{2pt}
  \noindent\begin{minipage}[t]{.81\linewidth}
    \fontsize{9}{10}\selectfont\bfseries #1
  \end{minipage}\hfill
  \begin{minipage}[t]{.17\linewidth}\raggedleft
    \fontsize{8}{9}\selectfont\bfseries\textsc{#2}
  \end{minipage}\par}
\newcommand{\ETMeta}[1]{%
  \vspace{1pt}{\fontsize{7.5}{8.5}\selectfont\color{ETMuted}#1\par}\vspace{2pt}}
\newcommand{\ETView}[3]{%
  \begin{minipage}[b]{#1\linewidth}
  \centering
  {\fontsize{7}{8}\selectfont\color{ETMuted}#2\par}\vspace{1pt}
  \makebox[\linewidth][c]{\includegraphics[page=#3,width=\linewidth,height=47pt,keepaspectratio]{figures.pdf}}%
  \end{minipage}}
\definecolor{PFFindingBackground}{HTML}{EDF4F8}
\definecolor{PFFindingAccent}{HTML}{35677C}
\newtcolorbox{PFFinding}[1]{%
  enhanced,unbreakable,frame hidden,boxrule=0pt,sharp corners,
  colback=PFFindingBackground,
  borderline west={2pt}{0pt}{PFFindingAccent},
  boxsep=0pt,left=7pt,right=7pt,top=5pt,bottom=6pt,
  before skip=5pt,after skip=0pt,
  fontupper=\fontsize{9}{10.5}\selectfont,
  before upper={\textcolor{PFFindingAccent}{\textbf{#1}}\par\vspace{2pt}}}

In \FunctionName{CoffeeSetupMug}, the robot must put a mug on the tray
beneath a coffee machine's outlet. A revision added a check intended to
stop movement after the robot loses its grasp: the arm stops when the
gripper no longer detects a held object. This change passed tests on
eight training cases, with no previous success lost. However, the newly
successful rollout never triggered the check.
Figure~\ref{fig:revision-regression-coffee} replays a previously successful
placement program. With the check enabled, descent stops too soon and
the later release leaves the mug outside the tray. Bypassing only this
check lets the mug descend onto the tray and restores task success.
\par\smallskip
\noindent\begin{minipage}{\linewidth}
\begin{ETCase}[unbreakable]{The effect of the stopping check}{Replay}
\ETHeading{The effect of the stopping check}{Replay}
\ETMeta{\ModelGPTAstra{} $\cdot$ Panda-Omron $\cdot$ saved program replay}
\newcommand{\RRView}[2]{%
  \begin{minipage}[b]{.316\linewidth}\centering
  {\fontsize{7}{8}\selectfont\color{ETMuted}#1\par}\vspace{1pt}
  \includegraphics[page=#2,width=\linewidth,height=62pt,keepaspectratio]{figures.pdf}%
  \end{minipage}}
\noindent\textbf{(a) Stopping check enabled}\hfill\textbf{Task fails}\par\vspace{2pt}
\noindent\RRView{Before lowering}{31}\hfill
\RRView{After lowering}{32}\hfill
\RRView{After release}{33}\par\vspace{5pt}
\noindent\textbf{(b) Only the stopping check bypassed}\hfill\textbf{Task succeeds}\par\vspace{2pt}
\noindent\RRView{Before lowering}{34}\hfill
\RRView{After lowering}{35}\hfill
\RRView{After release}{36}\par
\captionof{figure}{Replaying \FunctionName{CoffeeSetupMug} from the same initial state with the same program and early Epoch~3 code.
The two runs differ only in whether the stopping check is bypassed and share their first 51 device calls and results.
All images are from the side camera. Task success in (b) is recorded during a later \FunctionName{return\_home()} action,
after the displayed release frame. Neither replay makes model requests.}\label{fig:revision-regression-coffee}
\end{ETCase}
\end{minipage}\par\smallskip

A higher score on the selected tests therefore did not establish that
earlier successful programs would still work. In our setting, the
Reviser reviews 150 rollouts per batch and can inspect
epoch-level execution and revision histories. Looking across more cases
can reveal regressions and shared failure causes, but reading more
trajectories consumes time, computation, and context. We diagnose
available trajectories while other rollouts continue and write each
case's findings to \texttt{summary.md}. The Reviser reads these summaries
first and opens full sessions when it needs more detail. The prompt
requires testing target failures, related cases, and prior successes
that the patch may affect.

\begin{PFFinding}{Finding. The coverage--context trade-off.}
Reviser design involves a coverage--context trade-off. On one hand,
examining more cases and past revisions can reveal regressions. On the
other, those records compete for finite context during root-cause
analysis. Therefore, the Reviser needs a broad view of prior behavior
while retaining enough detail to explain individual failures.
\end{PFFinding}

\endgroup
\label{app:revision-tradeoff-end}

\newpage
\begingroup
\definecolor{ETRule}{HTML}{CBD0D4}
\definecolor{ETMuted}{HTML}{49525A}
\definecolor{ETCodeBackground}{HTML}{F4F5F6}
\lstdefinestyle{trajectory}{
  style=unicraft,
  basicstyle=\fontfamily{zi4}\mdseries\upshape\fontsize{7}{8}\selectfont\color{UcCodeInk},
  numbers=none,breaklines=true,breakatwhitespace=true,breakindent=8pt,
  backgroundcolor=\color{ETCodeBackground},
  morekeywords={pick,place,place_beside,grasp,lift,lift_both,navigate,
    adjust_height,align_base,survey,move_to,move_tcps,observe,segment_region,check_motion,set_gripper},
  aboveskip=0pt,belowskip=0pt
}
\newenvironment{ETCase}[3][]{%
  \begin{tcolorbox}[enhanced jigsaw,breakable,frame hidden,boxrule=0pt,
    colback=white,colbacktitle=white,coltitle=black,
    boxsep=0pt,left=0pt,right=0pt,top=0pt,bottom=0pt,
    before skip=0pt,after skip=4pt,pad at break*=1pt,
    fonttitle=\fontsize{8}{9}\selectfont\bfseries,
    title after break={#2 (continued)},#1]
  \fontsize{8}{9}\selectfont
  \captionsetup{font=footnotesize,labelfont=bf,justification=raggedright,
    singlelinecheck=false,hypcap=false,aboveskip=3pt,belowskip=0pt}
}{\end{tcolorbox}}
\newcommand{\ETHeading}[2]{%
  {\color{ETRule}\hrule height .35pt}\vspace{2pt}
  \noindent\begin{minipage}[t]{.81\linewidth}
    \fontsize{9}{10}\selectfont\bfseries #1
  \end{minipage}\hfill
  \begin{minipage}[t]{.17\linewidth}\raggedleft
    \fontsize{8}{9}\selectfont\bfseries\textsc{#2}
  \end{minipage}\par}
\newcommand{\ETMeta}[1]{%
  \vspace{1pt}{\fontsize{7.5}{8.5}\selectfont\color{ETMuted}#1\par}\vspace{2pt}}
\newcommand{\ETView}[3]{%
  \begin{minipage}[b]{#1\linewidth}
  \centering
  {\fontsize{7}{8}\selectfont\color{ETMuted}#2\par}\vspace{1pt}
  \makebox[\linewidth][c]{\includegraphics[page=#3,width=\linewidth,height=47pt,keepaspectratio]{figures.pdf}}%
  \end{minipage}}
\section{Execution Trajectories}\label{app:case-study}\label{app:execution-trajectories}

\noindent Each original benchmark includes three native successes and one failure; RoboDojo includes four successes and two failures. Images sample saved-observation quartiles; End may precede termination. Wrist panels show last saved views, absent when unused (Aloha/Baxter) or unavailable (Tiago). P$n$ identifies separate submissions. Excerpts retain target construction and control flow; imports, helper definitions, and intervening stages are omitted.\par\smallskip

\begin{ETCase}{(a) Moka pots on the stove}{Success}
\noindent\begin{minipage}{\linewidth}
\subsection{LIBERO-PRO}\label{app:trajectories-libero}
\ETHeading{(a) Moka pots on the stove}{Success}
\ETMeta{\ModelGPTAstra{} $\cdot$ Panda $\cdot$ rollout 2}
\noindent \ETView{.316}{Start}{37}\hfill\ETView{.316}{$1/4$}{38}\hfill\ETView{.316}{$1/2$}{39}\par\vspace{1pt}
\noindent \ETView{.316}{$3/4$}{40}\hfill\ETView{.316}{End}{41}\hfill\ETView{.316}{Wrist}{42}\par\vspace{1pt}
\end{minipage}\par
\vspace{1pt}\noindent\begin{minipage}{\linewidth}
\begin{lstlisting}[style=trajectory,xleftmargin=3pt,xrightmargin=3pt]
c = connect(); r = c.robot
d = load_artifact('artifacts/held2.json'); p = d['pot']
\end{lstlisting}
\end{minipage}\par\nointerlineskip
\begin{lstlisting}[style=trajectory,xleftmargin=3pt,xrightmargin=3pt]
tcp = d['tcp']; x = -0.12 + tcp.position[0] - p.pose.position[0]
y = -0.2 + tcp.position[1] - p.pose.position[1]
z = 0.932 + tcp.position[2] - (
    p.pose.position[2] - p.extent[2] / 2) + 0.007
for label, pos in [('raise', (tcp.position[0], tcp.position[1],
    1.25)), ('transfer', (x, y, 1.25)), ('lower', (x, y, z))]:
    result = r.move_to(Pose(pos, tcp.quaternion),
        stop_on_stall=True)
    print(label, result)
    if not result.success:
        break
else:
    print(r.set_gripper(1))
\end{lstlisting}
\nointerlineskip\noindent\begin{minipage}{\linewidth}
\begin{lstlisting}[style=trajectory,xleftmargin=3pt,xrightmargin=3pt]
    print(r.move_to(Pose((x, y, z + 0.15), tcp.quaternion),
        stop_on_stall=True))
\end{lstlisting}
\captionof{figure}{Stored pot and tool poses determine the offset-corrected waypoints. The loop stops on motion failure and releases only after successful raise--transfer--lower execution.}\label{fig:trajectory-libero_pro-01}
\label{fig:trajectories-libero}
\end{minipage}
\end{ETCase}

\begin{ETCase}{(b) Alphabet soup and tomato sauce in the basket}{Success}
\noindent\begin{minipage}{\linewidth}

\ETHeading{(b) Alphabet soup and tomato sauce in the basket}{Success}
\ETMeta{\ModelGeminiFlash{} $\cdot$ Panda $\cdot$ rollout 1}
\noindent \ETView{.316}{Start}{43}\hfill\ETView{.316}{$1/4$}{44}\hfill\ETView{.316}{$1/2$}{45}\par\vspace{1pt}
\noindent \ETView{.316}{$3/4$}{46}\hfill\ETView{.316}{End}{47}\hfill\ETView{.316}{Wrist}{48}\par\vspace{1pt}
\end{minipage}\par
\vspace{1pt}\noindent\begin{minipage}{\linewidth}
\begin{lstlisting}[style=trajectory,xleftmargin=3pt,xrightmargin=3pt]
connection = connect(); robot = connection.robot
camera = connection.cameras['scene']
\end{lstlisting}
\end{minipage}\par\nointerlineskip
\begin{lstlisting}[style=trajectory,xleftmargin=3pt,xrightmargin=3pt]
obs = camera.observe(save_path='snapshots/retry')
objs = obs.value['objects']; blue_can_bbox = (239, 479, 338, 625)
basket_bbox = (770, 453, 1000, 776)
blue_can = find_obj_by_bbox(objs, blue_can_bbox)
basket = find_obj_by_bbox(objs, basket_bbox)
if blue_can:
    print(f'Picking blue can {blue_can.id}...')
    result = robot.pick(blue_can,
        camera=connection.cameras['wrist'])
    print('Pick blue can result:', result.status)
    if result.status == 'success':
        print('Placing blue can...')
        robot.place(basket)
\end{lstlisting}
\nointerlineskip\noindent\begin{minipage}{\linewidth}
\begin{lstlisting}[style=trajectory,xleftmargin=3pt,xrightmargin=3pt]
    else:
        print('Failed to pick blue can.')
\end{lstlisting}
\captionof{figure}{Fresh observations update object identities through bounding-box centroid matching; that helper and other object branches are omitted. Placement requires a successful pick. The pattern is reused across objects.}\label{fig:trajectory-libero_pro-02}

\end{minipage}
\end{ETCase}

\begin{ETCase}{(c) White mug on the plate; pudding to its right}{Success}
\noindent\begin{minipage}{\linewidth}

\ETHeading{(c) White mug on the plate; pudding to its right}{Success}
\ETMeta{\ModelDeepSeekFlash{} $\cdot$ Panda $\cdot$ rollout 2}
\noindent \ETView{.316}{Start}{49}\hfill\ETView{.316}{$1/4$}{50}\hfill\ETView{.316}{$1/2$}{51}\par\vspace{1pt}
\noindent \ETView{.316}{$3/4$}{52}\hfill\ETView{.316}{End}{53}\hfill\ETView{.316}{Wrist}{54}\par\vspace{1pt}
\end{minipage}\par
\vspace{1pt}\noindent\begin{minipage}{\linewidth}
\begin{lstlisting}[style=trajectory,xleftmargin=3pt,xrightmargin=3pt]
connection = connect(); robot = connection.robot
scene = connection.cameras['scene']
\end{lstlisting}
\end{minipage}\par\nointerlineskip
\begin{lstlisting}[style=trajectory,xleftmargin=3pt,xrightmargin=3pt]
wrist = connection.cameras['wrist']
plate = scene.get('scene_obj_087')
table = scene.get('scene_obj_073')
held_pudding = wrist.get('wrist_obj_021')
result = robot.place_beside(plate, table, held_pudding,
    direction=(0.0, 1.0), gap=0.03, clearance=0.03,
\end{lstlisting}
\nointerlineskip\noindent\begin{minipage}{\linewidth}
\begin{lstlisting}[style=trajectory,xleftmargin=3pt,xrightmargin=3pt]
    effector='arm', stop_on_stall=True)
print('place_beside', result)
\end{lstlisting}
\captionof{figure}{Scene-view plate and table measurements combine with the wrist-view carried object to ground relative placement. Later programs regrasp the mug by its handle and place it on the plate.}\label{fig:trajectory-libero_pro-03}

\end{minipage}
\end{ETCase}

\begin{ETCase}{(d) Black bowl in bottom drawer; close it}{Failure}
\noindent\begin{minipage}{\linewidth}

\ETHeading{(d) Black bowl in bottom drawer; close it}{Failure}
\ETMeta{\ModelGPTAstra{} $\cdot$ Panda $\cdot$ rollout 5}
\noindent \ETView{.316}{Start}{55}\hfill\ETView{.316}{$1/4$}{56}\hfill\ETView{.316}{$1/2$}{57}\par\vspace{1pt}
\noindent \ETView{.316}{$3/4$}{58}\hfill\ETView{.316}{End}{59}\hfill\ETView{.316}{Wrist}{60}\par\vspace{1pt}
\end{minipage}\par
\vspace{1pt}\noindent\begin{minipage}{\linewidth}
\begin{lstlisting}[style=trajectory,xleftmargin=3pt,xrightmargin=3pt]
c = connect(); cam = c.cameras['scene']
r = c.robot
\end{lstlisting}
\end{minipage}\par\nointerlineskip
\begin{lstlisting}[style=trajectory,xleftmargin=3pt,xrightmargin=3pt]
b = cam.segment_region((626, 435, 795, 580), seed=(671, 500),
    save_path='snapshots/carried2')
print('body', b)
s = cam.segment_region((595, 570, 817, 725), seed=(692, 665),
    save_path='snapshots/floor2')
print('floor', s)
print(r.place(s, carried=b, clearance=0.008, stop_on_stall=True))
print(r.state)
print(cam.observe(save_path='snapshots/placed').status)

c = connect(); r = c.robot
print(r.pull((0, 1, 0), distance=0.15, step=0.025,
    rotate_tool=False))
print(r.state)
\end{lstlisting}
\nointerlineskip\noindent\begin{minipage}{\linewidth}
\begin{lstlisting}[style=trajectory,xleftmargin=3pt,xrightmargin=3pt]
print(c.cameras['scene'].observe(
    save_path='snapshots/closing').status)
\end{lstlisting}
\captionof{figure}{Placement loses holding contact. After recovery and handle grasping, P20 achieves only 0.039 m of the requested 0.15 m motion. Drawer closure remains incomplete at the native horizon.}\label{fig:trajectory-libero_pro-04}

\end{minipage}
\end{ETCase}

\begin{ETCase}{(a) Rank blocks by size}{Success}
\noindent\begin{minipage}{\linewidth}
\subsection{RoboTwin 2.0}\label{app:trajectories-robotwin}
\ETHeading{(a) Rank blocks by size}{Success}
\ETMeta{\ModelGPTAstra{} $\cdot$ Aloha-AgileX $\cdot$ rollout 1}
\noindent \ETView{.316}{Start}{61}\hfill\ETView{.316}{$1/4$}{62}\hfill\ETView{.316}{$1/2$}{63}\par\vspace{1pt}
\noindent \ETView{.316}{$3/4$}{64}\hspace{.025\linewidth}\ETView{.316}{End}{65}\par\vspace{1pt}
\end{minipage}\par
\vspace{1pt}\noindent\begin{minipage}{\linewidth}
\begin{lstlisting}[style=trajectory,xleftmargin=3pt,xrightmargin=3pt]
c = connect(); cam = c.cameras['scene']
r = c.robot
\end{lstlisting}
\end{minipage}\par\nointerlineskip
\begin{lstlisting}[style=trajectory,xleftmargin=3pt,xrightmargin=3pt]
save_artifact('programs/blocks.json',
    {'large': cam.get('scene_obj_013'),
    'medium': cam.get('scene_obj_012'),
    'small': cam.get('scene_obj_014'),
    'table': cam.get('scene_obj_007')})
print(r.pick(cam.get('scene_obj_013'), effector='right',
    depth=0.025, stop_on_stall=True))
\end{lstlisting}
\nointerlineskip\noindent\begin{minipage}{\linewidth}
\begin{lstlisting}[style=trajectory,xleftmargin=3pt,xrightmargin=3pt]
print(r.state)
print(cam.observe(save_path='snapshots/large_pick'))
\end{lstlisting}
\captionof{figure}{Observed objects are assigned size roles and saved for later programs. The large block is picked with the right arm; subsequent programs reuse both arms to establish the requested order.}\label{fig:trajectory-robotwin-05}
\label{fig:trajectories-robotwin}
\end{minipage}
\end{ETCase}

\begin{ETCase}{(b) Place object A to the left of object B}{Success}
\noindent\begin{minipage}{\linewidth}

\ETHeading{(b) Place object A to the left of object B}{Success}
\ETMeta{\ModelGPTAstra{} $\cdot$ dual UR5-WSG $\cdot$ rollout 1}
\noindent \ETView{.235}{Start}{66}\hfill\ETView{.235}{$1/4$}{67}\hfill\ETView{.235}{$1/2$}{68}\hfill\ETView{.235}{$3/4$}{69}\par\vspace{1pt}
\noindent \ETView{.235}{End}{70}\hspace{.02\linewidth}\ETView{.235}{Left wrist}{71}\hspace{.02\linewidth}\ETView{.235}{Right wrist}{72}\par\vspace{1pt}
\end{minipage}\par
\vspace{1pt}\noindent\begin{minipage}{\linewidth}
\begin{lstlisting}[style=trajectory,xleftmargin=3pt,xrightmargin=3pt]
c = connect(); cam = c.cameras['scene']
held = load_artifact('programs/held_cube.json')
\end{lstlisting}
\end{minipage}\par\nointerlineskip
\begin{lstlisting}[style=trajectory,xleftmargin=3pt,xrightmargin=3pt]
print(c.robot.place_beside(cam.get('scene_obj_084'),
    cam.get('scene_obj_076'), held, direction=(-1, 0), gap=0.04,
    clearance=0.015, effector='left', stop_on_stall=True))
\end{lstlisting}
\nointerlineskip\noindent\begin{minipage}{\linewidth}
\begin{lstlisting}[style=trajectory,xleftmargin=3pt,xrightmargin=3pt]
print(c.robot.state.effectors['left'])
print(cam.observe(save_path='snapshots/placement_check'))
\end{lstlisting}
\captionof{figure}{After a resumed pick, P16 loads the held-object measurement and retrieves the reference and support from the current view. Direction and gap specify the left-of relation.}\label{fig:trajectory-robotwin-06}

\end{minipage}
\end{ETCase}

\begin{ETCase}{(c) Stack three bowls}{Success}
\noindent\begin{minipage}{\linewidth}

\ETHeading{(c) Stack three bowls}{Success}
\ETMeta{\ModelGPTAstra{} $\cdot$ dual UR5-WSG $\cdot$ rollout 2}
\noindent \ETView{.235}{Start}{73}\hfill\ETView{.235}{$1/4$}{74}\hfill\ETView{.235}{$1/2$}{75}\hfill\ETView{.235}{$3/4$}{76}\par\vspace{1pt}
\noindent \ETView{.235}{End}{77}\hspace{.02\linewidth}\ETView{.235}{Left wrist}{78}\hspace{.02\linewidth}\ETView{.235}{Right wrist}{79}\par\vspace{1pt}
\end{minipage}\par
\vspace{1pt}\noindent\begin{minipage}{\linewidth}
\begin{lstlisting}[style=trajectory,xleftmargin=3pt,xrightmargin=3pt]
c = connect(); r = c.robot
p = r.state.effectors['right'].pose
\end{lstlisting}
\end{minipage}\par\nointerlineskip
\begin{lstlisting}[style=trajectory,xleftmargin=3pt,xrightmargin=3pt]
q = r.move_to(Pose((p.position[0], p.position[1], 0.88),
    p.quaternion), effector='right', stop_on_stall=True)
print(q)
if q.success:
    print(r.set_gripper(1, effector='right'))
    print(r.move_to(Pose((p.position[0], p.position[1], 1.0),
        p.quaternion), effector='right', stop_on_stall=True))
\end{lstlisting}
\nointerlineskip\noindent\begin{minipage}{\linewidth}
\begin{lstlisting}[style=trajectory,xleftmargin=3pt,xrightmargin=3pt]
print(c.cameras['scene'].observe(
    save_path='snapshots/stack3').message)
\end{lstlisting}
\captionof{figure}{Earlier programs lift and transfer bowls with alternating arms. P15 preserves horizontal position and orientation while lowering the final bowl; release is conditional on reaching the target.}\label{fig:trajectory-robotwin-07}

\end{minipage}
\end{ETCase}

\begin{ETCase}{(d) Stack three bowls}{Failure}
\noindent\begin{minipage}{\linewidth}

\ETHeading{(d) Stack three bowls}{Failure}
\ETMeta{\ModelGPTAstra{} $\cdot$ dual UR5-WSG $\cdot$ rollout 3}
\noindent \ETView{.235}{Start}{80}\hfill\ETView{.235}{$1/4$}{81}\hfill\ETView{.235}{$1/2$}{82}\hfill\ETView{.235}{$3/4$}{83}\par\vspace{1pt}
\noindent \ETView{.235}{End}{84}\hspace{.02\linewidth}\ETView{.235}{Left wrist}{85}\hspace{.02\linewidth}\ETView{.235}{Right wrist}{86}\par\vspace{1pt}
\end{minipage}\par
\vspace{1pt}\noindent\begin{minipage}{\linewidth}
\begin{lstlisting}[style=trajectory,xleftmargin=3pt,xrightmargin=3pt]
c = connect(); r = c.robot
print(r.set_gripper(0, effector='left'))
\end{lstlisting}
\end{minipage}\par\nointerlineskip
\begin{lstlisting}[style=trajectory,xleftmargin=3pt,xrightmargin=3pt]
print(r.lift(0.1, effector='left'))
print(r.state.effectors['left'])
print(c.cameras['scene'].observe(
    save_path='snapshots/top_relift').message)
\end{lstlisting}
\nointerlineskip\noindent\begin{minipage}{\linewidth}
\begin{lstlisting}[style=trajectory,xleftmargin=3pt,xrightmargin=3pt]
print(c.cameras['left_wrist'].observe(
    save_path='snapshots/top_relift_wrist').message)
\end{lstlisting}
\captionof{figure}{With the same initialization as the preceding success, closure reports holding but the corrective lift loses contact. Further scene and wrist observations guide unsuccessful recovery after a bowl falls.}\label{fig:trajectory-robotwin-08}

\end{minipage}
\end{ETCase}

\begin{ETCase}{(a) CloseBlenderLid}{Success}
\noindent\begin{minipage}{\linewidth}
\subsection{RoboCasa365}\label{app:trajectories-robocasa}
\ETHeading{(a) CloseBlenderLid}{Success}
\ETMeta{\ModelGPTAstra{} $\cdot$ Panda-Omron $\cdot$ rollout 4}
\noindent \ETView{.316}{Start}{87}\hfill\ETView{.316}{$1/4$}{88}\hfill\ETView{.316}{$1/2$}{89}\par\vspace{1pt}
\noindent \ETView{.316}{$3/4$}{90}\hfill\ETView{.316}{End}{91}\hfill\ETView{.316}{Wrist}{92}\par\vspace{1pt}
\end{minipage}\par
\vspace{1pt}\noindent\begin{minipage}{\linewidth}
\begin{lstlisting}[style=trajectory,xleftmargin=3pt,xrightmargin=3pt]
c = connect(); cam = c.cameras['wrist']
print(c.robot.grasp(cam.get('wrist_obj_010'), depth=0.013,
\end{lstlisting}
\end{minipage}\par\nointerlineskip
\begin{lstlisting}[style=trajectory,xleftmargin=3pt,xrightmargin=3pt]
    clearance=0.1, approach=(0, 0, -1), closing_axis=(1, 0, 0),
    stop_on_stall=True))
print(c.robot.state.effectors['arm'])

c = connect()
print(c.robot.lift(0.28, step=0.04, stop_on_stall=True))
\end{lstlisting}
\nointerlineskip\noindent\begin{minipage}{\linewidth}
\begin{lstlisting}[style=trajectory,xleftmargin=3pt,xrightmargin=3pt]
r = c.cameras['scene'].observe(save_path='snapshots/lid_lifted')
print(r)
\end{lstlisting}
\captionof{figure}{Following an earlier grasp attempt, a wrist-view target yields confirmed holding. P11 lifts the lid incrementally and requests a new scene view before later transport and release.}\label{fig:trajectory-robocasa-09}
\label{fig:trajectories-robocasa}
\end{minipage}
\end{ETCase}

\begin{ETCase}{(b) BreadSelection}{Success}
\noindent\begin{minipage}{\linewidth}

\ETHeading{(b) BreadSelection}{Success}
\ETMeta{\ModelGPTAstra{} $\cdot$ Panda-Omron $\cdot$ rollout 5}
\noindent \ETView{.316}{Start}{93}\hfill\ETView{.316}{$1/4$}{94}\hfill\ETView{.316}{$1/2$}{95}\par\vspace{1pt}
\noindent \ETView{.316}{$3/4$}{96}\hfill\ETView{.316}{End}{97}\hfill\ETView{.316}{Wrist}{98}\par\vspace{1pt}
\end{minipage}\par
\vspace{1pt}\noindent\begin{minipage}{\linewidth}
\begin{lstlisting}[style=trajectory,xleftmargin=3pt,xrightmargin=3pt]
c = connect(); p = c.robot.state.effectors['arm'].pose
r = c.robot.move_to(Pose((4.1, -0.55, 1.1), p.quaternion),
\end{lstlisting}
\end{minipage}\par\nointerlineskip
\begin{lstlisting}[style=trajectory,xleftmargin=3pt,xrightmargin=3pt]
    stop_on_stall=True)
print('retreat', r)
if r.success:
\end{lstlisting}
\nointerlineskip\noindent\begin{minipage}{\linewidth}
\begin{lstlisting}[style=trajectory,xleftmargin=3pt,xrightmargin=3pt]
    print('elevate', c.robot.adjust_height(0.28))
print(c.robot.state)
\end{lstlisting}
\captionof{figure}{After placing the croissant, the arm retreats before the torso rises. Elevation requires successful retreat, preparing the robot to retrieve cabinet-stored jam and place it beside the croissant.}\label{fig:trajectory-robocasa-10}

\end{minipage}
\end{ETCase}

\begin{ETCase}{(c) NavigateKitchen}{Success}
\noindent\begin{minipage}{\linewidth}

\ETHeading{(c) NavigateKitchen}{Success}
\ETMeta{\ModelDeepSeekFlash{} $\cdot$ Panda-Omron $\cdot$ rollout 4}
\noindent \ETView{.316}{Start}{99}\hfill\ETView{.316}{$1/4$}{100}\hfill\ETView{.316}{$1/2$}{101}\par\vspace{1pt}
\noindent \ETView{.316}{$3/4$}{102}\hfill\ETView{.316}{End}{103}\hfill\ETView{.316}{Wrist}{104}\par\vspace{1pt}
\end{minipage}\par
\vspace{1pt}\noindent\begin{minipage}{\linewidth}
\begin{lstlisting}[style=trajectory,xleftmargin=3pt,xrightmargin=3pt]
connection = connect(); robot = connection.robot
camera = connection.cameras['scene']
\end{lstlisting}
\end{minipage}\par\nointerlineskip
\begin{lstlisting}[style=trajectory,xleftmargin=3pt,xrightmargin=3pt]
toaster = camera.get('scene_obj_426')
result = robot.align_base(toaster, direction=(-1.0, 0.0),
    standoff=0.55, clearance=0.3, timeout_s=20.0, effector='arm')
\end{lstlisting}
\nointerlineskip\noindent\begin{minipage}{\linewidth}
\begin{lstlisting}[style=trajectory,xleftmargin=3pt,xrightmargin=3pt]
print('align_base:', result)
print('after:', robot.actions.read_base())
\end{lstlisting}
\captionof{figure}{After bounded survey and turning attempts, a fresh observation identifies the toaster. Object-relative direction, standoff, and clearance then guide base alignment to task success.}\label{fig:trajectory-robocasa-11}

\end{minipage}
\end{ETCase}

\begin{ETCase}{(d) SlideDishwasherRack}{Failure}
\noindent\begin{minipage}{\linewidth}

\ETHeading{(d) SlideDishwasherRack}{Failure}
\ETMeta{\ModelGPTAstra{} $\cdot$ Panda-Omron $\cdot$ rollout 2}
\noindent \ETView{.316}{Start}{105}\hfill\ETView{.316}{$1/4$}{106}\hfill\ETView{.316}{$1/2$}{107}\par\vspace{1pt}
\noindent \ETView{.316}{$3/4$}{108}\hfill\ETView{.316}{End}{109}\hfill\ETView{.316}{Wrist}{110}\par\vspace{1pt}
\end{minipage}\par
\vspace{1pt}\noindent\begin{minipage}{\linewidth}
\begin{lstlisting}[style=trajectory,xleftmargin=3pt,xrightmargin=3pt]
c = connect(); cam = c.cameras['wrist']
r = c.robot
\end{lstlisting}
\end{minipage}\par\nointerlineskip
\begin{lstlisting}[style=trajectory,xleftmargin=3pt,xrightmargin=3pt]
p = cam.segment_region((445, 300, 515, 720), seed=(478, 510),
    save_path='snapshots/front_bar_same')
print(r.grasp(p, clearance=0.055, depth=0.009, approach=(1, 0, 0),
    closing_axis=(0, 0, 1), opening=0.4, stop_on_stall=True))
print(r.state)
print(cam.observe(save_path='snapshots/contact').status)

c = connect(); r = c.robot
print(r.set_gripper(0))
print(r.state)
\end{lstlisting}
\nointerlineskip\noindent\begin{minipage}{\linewidth}
\begin{lstlisting}[style=trajectory,xleftmargin=3pt,xrightmargin=3pt]
print(c.cameras['scene'].observe(
    save_path='snapshots/closed').status)
\end{lstlisting}
\captionof{figure}{Wrist-view segmentation localizes the rack bar, but grasp descent stalls. Closure leaves holding false. Subsequent pose and base adjustments consume the horizon without a successful pull.}\label{fig:trajectory-robocasa-12}

\end{minipage}
\end{ETCase}

\begin{ETCase}{(a) PickPlace}{Success}
\noindent\begin{minipage}{\linewidth}
\subsection{robosuite}\label{app:trajectories-robosuite}
\ETHeading{(a) PickPlace}{Success}
\ETMeta{\ModelGPTAstra{} $\cdot$ Sawyer $\cdot$ rollout 2}
\noindent \ETView{.316}{Start}{111}\hfill\ETView{.316}{$1/4$}{112}\hfill\ETView{.316}{$1/2$}{113}\par\vspace{1pt}
\noindent \ETView{.316}{$3/4$}{114}\hfill\ETView{.316}{End}{115}\hfill\ETView{.316}{Wrist}{116}\par\vspace{1pt}
\end{minipage}\par
\vspace{1pt}\noindent\begin{minipage}{\linewidth}
\begin{lstlisting}[style=trajectory,xleftmargin=3pt,xrightmargin=3pt]
c = connect(); r = c.robot
p = r.state.pose
\end{lstlisting}
\end{minipage}\par\nointerlineskip
\begin{lstlisting}[style=trajectory,xleftmargin=3pt,xrightmargin=3pt]
for pos in [(0.2, 0.415, 1.1), (0.2, 0.415, 0.9)]:
    res = r.move_to(Pose(pos, p.quaternion), stop_on_stall=True)
    print(res)
    if not res.success:
        break
else:
    print(r.set_gripper(1))
    print(r.move_to(Pose((0.2, 0.415, 1.15), p.quaternion),
\end{lstlisting}
\nointerlineskip\noindent\begin{minipage}{\linewidth}
\begin{lstlisting}[style=trajectory,xleftmargin=3pt,xrightmargin=3pt]
        stop_on_stall=True))
print(c.cameras['scene'].observe(save_path='snapshots/final'))
\end{lstlisting}
\captionof{figure}{After the can is picked, transport and lowering preserve its orientation. Failed waypoints prevent release. Analogous pick-and-transfer stages handle the other three object categories.}\label{fig:trajectory-robosuite-13}
\label{fig:trajectories-robosuite}
\end{minipage}
\end{ETCase}

\begin{ETCase}{(b) TwoArmLift}{Success}
\noindent\begin{minipage}{\linewidth}

\ETHeading{(b) TwoArmLift}{Success}
\ETMeta{\ModelGPTSol{} $\cdot$ Baxter $\cdot$ rollout 3}
\noindent \ETView{.316}{Start}{117}\hfill\ETView{.316}{$1/4$}{118}\hfill\ETView{.316}{$1/2$}{119}\par\vspace{1pt}
\noindent \ETView{.316}{$3/4$}{120}\hspace{.025\linewidth}\ETView{.316}{End}{121}\par\vspace{1pt}
\end{minipage}\par
\vspace{1pt}\noindent\begin{minipage}{\linewidth}
\begin{lstlisting}[style=trajectory,xleftmargin=3pt,xrightmargin=3pt]
c = connect(); targets = {}
for arm, s in c.robot.state.effectors.items():
\end{lstlisting}
\end{minipage}\par\nointerlineskip
\begin{lstlisting}[style=trajectory,xleftmargin=3pt,xrightmargin=3pt]
    p = s.pose
    targets[arm] = Pose((p.position[0], p.position[1],
        p.position[2] + 0.14), p.quaternion)
print(c.robot.actions.move_tcps(targets, timeout_s=12),
    flush=True)
print(c.robot.state.effectors, flush=True)
print(c.cameras['side'].observe(
    save_path='snapshots/lift2_side').status, flush=True)
\end{lstlisting}
\nointerlineskip\noindent\begin{minipage}{\linewidth}
\begin{lstlisting}[style=trajectory,xleftmargin=3pt,xrightmargin=3pt]
print(c.cameras['scene'].observe(
    save_path='snapshots/lift2_scene').status, flush=True)
\end{lstlisting}
\captionof{figure}{A guarded lift stops short after both handles are grasped. P11 adds 0.14 m to each measured arm pose while preserving orientation, then submits both targets in one synchronized native action.}\label{fig:trajectory-robosuite-14}

\end{minipage}
\end{ETCase}

\begin{ETCase}{(c) TwoArmHandover}{Success}
\noindent\begin{minipage}{\linewidth}

\ETHeading{(c) TwoArmHandover}{Success}
\ETMeta{\ModelGPTSol{} $\cdot$ Tiago $\cdot$ rollout 5}
\noindent \ETView{.316}{Start}{122}\hfill\ETView{.316}{$1/4$}{123}\hfill\ETView{.316}{$1/2$}{124}\par\vspace{1pt}
\noindent \ETView{.316}{$3/4$}{125}\hspace{.025\linewidth}\ETView{.316}{End}{126}\par\vspace{1pt}
\end{minipage}\par
\vspace{1pt}\noindent\begin{minipage}{\linewidth}
\begin{lstlisting}[style=trajectory,xleftmargin=3pt,xrightmargin=3pt]
c = connect()
q = load_grasp_candidate('programs/left_grasps.json',
\end{lstlisting}
\end{minipage}\par\nointerlineskip
\begin{lstlisting}[style=trajectory,xleftmargin=3pt,xrightmargin=3pt]
    'C7')['target'].quaternion
print(c.robot.set_gripper(1, effector='left'))
print(c.robot.move_to(Pose((-0.117, 0.045, 1.16), q),
    effector='left', stop_on_stall=True))
r = c.robot.actions.move_tcp(Pose((-0.108, -0.012, 1.117), q),
    effector='left', position_tolerance_m=0.005,
    stop_on_stall=False)
print(r)
print(c.robot.set_gripper(0, effector='left'))
print(c.robot.state)
print(c.cameras['scene'].observe(
    save_path='snapshots/left_mid_grip').status)

c = connect()
\end{lstlisting}
\nointerlineskip\noindent\begin{minipage}{\linewidth}
\begin{lstlisting}[style=trajectory,xleftmargin=3pt,xrightmargin=3pt]
print(c.robot.set_gripper(1, effector='right'))
print(c.robot.state)
\end{lstlisting}
\captionof{figure}{Following base repositioning and elevation, P34 uses a saved grasp orientation to establish the receiving grip and inspect contact. P35 then releases the sending right hand.}\label{fig:trajectory-robosuite-15}

\end{minipage}
\end{ETCase}

\begin{ETCase}{(d) TwoArmLift}{Failure}
\noindent\begin{minipage}{\linewidth}

\ETHeading{(d) TwoArmLift}{Failure}
\ETMeta{\ModelMiniMaxFlash{} $\cdot$ Tiago $\cdot$ rollout 3}
\noindent \ETView{.316}{Start}{127}\hfill\ETView{.316}{$1/4$}{128}\hfill\ETView{.316}{$1/2$}{129}\par\vspace{1pt}
\noindent \ETView{.316}{$3/4$}{130}\hspace{.025\linewidth}\ETView{.316}{End}{131}\par\vspace{1pt}
\end{minipage}\par
\vspace{1pt}\noindent\begin{minipage}{\linewidth}
\begin{lstlisting}[style=trajectory,xleftmargin=3pt,xrightmargin=3pt]
c = connect(); cam = c.cameras['scene']
cam.observe(save_path='snapshots/chk')
\end{lstlisting}
\end{minipage}\par\nointerlineskip
\begin{lstlisting}[style=trajectory,xleftmargin=3pt,xrightmargin=3pt]
f = cam.frame
for nm, eff in (('GREEN', 'right'), ('BLUE', 'left')):
    res = json.load(open('snapshots/prop_%s.json' % nm))
    for i, cand in enumerate(res['candidates']):
        t = _pose(cand['target']); p = _pose(cand['pregrasp'])
        r = c.robot.check_motion(Pose(tuple(t[0]), tuple(t[1])),
            via=[Pose(tuple(p[0]), tuple(p[1]))], frames=[f],
            opening=1.0, effector=eff)
        st = (r.get('stages') or [{}])[0]
        best = st.get('best', {})
        pe = best.get('position_error_m')
\end{lstlisting}
\nointerlineskip\noindent\begin{minipage}{\linewidth}
\begin{lstlisting}[style=trajectory,xleftmargin=3pt,xrightmargin=3pt]
        ok = r.get('status') == 'found' and st.get(
            'sampled_interpolation', {}).get('status') == 'clear'
\end{lstlisting}
\captionof{figure}{Candidate grasps are screened for both arms; the pose-conversion helper is omitted. Across the episode, 200 requests and 1,517 motion checks yield no physical action.}\label{fig:trajectory-robosuite-16}

\end{minipage}
\end{ETCase}

\begin{ETCase}[unbreakable]{(a) BuildTower}{Success}
\noindent\begin{minipage}{\linewidth}
\subsection{RoboDojo}\label{app:trajectories-robodojo}
\ETHeading{(a) BuildTower}{Success}
\ETMeta{\ModelGPTAstra{} $\cdot$ dual ARX-X5 $\cdot$ rollout 1}
\noindent \ETView{.316}{Start}{132}\hfill\ETView{.316}{$1/4$}{133}\hfill\ETView{.316}{$1/2$}{134}\par\vspace{1pt}
\noindent \ETView{.316}{$3/4$}{135}\hspace{.025\linewidth}\ETView{.316}{End}{136}\par\vspace{1pt}
\end{minipage}\par
\vspace{1pt}\noindent\begin{minipage}{\linewidth}
\begin{lstlisting}[style=trajectory,xleftmargin=3pt,xrightmargin=3pt]
c = connect(); cam = c.cameras['scene']
objects = {name: cam.get('scene_obj_' + str(i).zfill(3)) for (
\end{lstlisting}
\end{minipage}\par\nointerlineskip
\begin{lstlisting}[style=trajectory,xleftmargin=3pt,xrightmargin=3pt]
    name, i) in [('Lrear', 17), ('Lfront', 21), ('Rrear', 16),
    ('Rfront', 18), ('large', 26), ('middle', 30), ('top', 23),
    ('triangle', 19)]}
save_artifact('artifacts/initial.json', objects)
c.robot.pick(objects['Lrear'], effector='left', clearance=0.07,
\end{lstlisting}
\nointerlineskip\noindent\begin{minipage}{\linewidth}
\begin{lstlisting}[style=trajectory,xleftmargin=3pt,xrightmargin=3pt]
    depth=0.013, position_tolerance_m=0.003,
    rotation_tolerance_rad=0.04)
\end{lstlisting}
\captionof{figure}{P2 assigns observed pieces to roles and saves their Object records before picking a support. Later programs reuse the records to assemble all eight pieces. The native task succeeds at 939 steps.}\label{fig:trajectory-robodojo-17}
\label{fig:trajectories-robodojo}
\end{minipage}
\end{ETCase}

\begin{ETCase}[unbreakable]{(b) FastenScrews}{Success}
\noindent\begin{minipage}{\linewidth}

\ETHeading{(b) FastenScrews}{Success}
\ETMeta{\ModelGPTAstra{} $\cdot$ dual ARX-X5 $\cdot$ rollout 1}
\noindent \ETView{.235}{Start}{137}\hfill\ETView{.235}{$1/4$}{138}\hfill\ETView{.235}{$1/2$}{139}\hfill\ETView{.235}{$3/4$}{140}\par\vspace{1pt}
\noindent \ETView{.235}{End}{141}\hspace{.02\linewidth}\ETView{.235}{Left wrist}{142}\hspace{.02\linewidth}\ETView{.235}{Right wrist}{143}\par\vspace{1pt}
\end{minipage}\par
\vspace{1pt}\noindent\begin{minipage}{\linewidth}
\begin{lstlisting}[style=trajectory,xleftmargin=3pt,xrightmargin=3pt]
c = connect(); r = c.robot
arm = sys.argv[1]; angle = float(sys.argv[2])
\end{lstlisting}
\end{minipage}\par\nointerlineskip
\begin{lstlisting}[style=trajectory,xleftmargin=3pt,xrightmargin=3pt]
drop = float(sys.argv[3]); p = r.state.effectors[arm].pose
start = np.array(p.position)
rot = Rotation.from_quat(p.quaternion)
n = int(np.ceil(abs(angle) / 0.3))
for i in range(1, n + 1):
    xyz = start.copy()
    xyz[2] -= drop * i / n
    q = (Rotation.from_rotvec([0, 0,
        angle * i / n]) * rot).as_quat()
    result = r.move_to(Pose(tuple(xyz), tuple(q)), effector=arm,
        position_tolerance_m=0.0012, rotation_tolerance_rad=0.015,
        stop_on_stall=True)
\end{lstlisting}
\nointerlineskip\noindent\begin{minipage}{\linewidth}
\begin{lstlisting}[style=trajectory,xleftmargin=3pt,xrightmargin=3pt]
    if not result.success:
        break
\end{lstlisting}
\captionof{figure}{P50 combines incremental rotation and downward motion from the measured TCP pose, stopping on failed motion feedback. Repeated grasp--turn--release stages fasten all three nuts; native success occurs at 1,289 steps.}\label{fig:trajectory-robodojo-18}

\end{minipage}
\end{ETCase}

\begin{ETCase}[unbreakable]{(c) SweepBlocks}{Success}
\noindent\begin{minipage}{\linewidth}

\ETHeading{(c) SweepBlocks}{Success}
\ETMeta{\ModelGPTAstra{} $\cdot$ dual ARX-X5 $\cdot$ rollout 1}
\noindent \ETView{.235}{Start}{144}\hfill\ETView{.235}{$1/4$}{145}\hfill\ETView{.235}{$1/2$}{146}\hfill\ETView{.235}{$3/4$}{147}\par\vspace{1pt}
\noindent \ETView{.235}{End}{148}\hspace{.02\linewidth}\ETView{.235}{Left wrist}{149}\hspace{.02\linewidth}\ETView{.235}{Right wrist}{150}\par\vspace{1pt}
\end{minipage}\par
\vspace{1pt}\noindent\begin{minipage}{\linewidth}
\begin{lstlisting}[style=trajectory,xleftmargin=3pt,xrightmargin=3pt]
c = connect(); r = c.robot
p = r.state.effectors['right'].pose
\end{lstlisting}
\end{minipage}\par\nointerlineskip
\begin{lstlisting}[style=trajectory,xleftmargin=3pt,xrightmargin=3pt]
r.move_to(Pose((0.065, -0.19, 0.792), p.quaternion),
    effector='right', position_tolerance_m=0.003,
    stop_on_stall=True)
p = r.state.effectors['right'].pose
r.move_to(Pose((p.position[0], p.position[1], 0.9), p.quaternion),
    effector='right', position_tolerance_m=0.004,
    stop_on_stall=True)

c = connect(); r = c.robot
p = r.state.effectors['left'].pose
r.move_to(Pose((p.position[0] + 0.045, p.position[1],
\end{lstlisting}
\nointerlineskip\noindent\begin{minipage}{\linewidth}
\begin{lstlisting}[style=trajectory,xleftmargin=3pt,xrightmargin=3pt]
    p.position[2]), p.quaternion), effector='left',
    position_tolerance_m=0.003, stop_on_stall=True)
\end{lstlisting}
\captionof{figure}{P84 sweeps with the right-hand brush and lifts it away; P86 advances the left-hand dustpan. Each stage uses a newly measured pose. Subsequent adjustment and reset reach native success at 922 steps.}\label{fig:trajectory-robodojo-19}

\end{minipage}
\end{ETCase}

\begin{ETCase}[unbreakable]{(d) PressByNumber}{Success}
\noindent\begin{minipage}{\linewidth}

\ETHeading{(d) PressByNumber}{Success}
\ETMeta{\ModelGPTAstra{} $\cdot$ dual ARX-X5 $\cdot$ rollout 1}
\noindent \ETView{.316}{Start}{151}\hfill\ETView{.316}{$1/4$}{152}\hfill\ETView{.316}{$1/2$}{153}\par\vspace{1pt}
\noindent \ETView{.316}{$3/4$}{154}\hspace{.025\linewidth}\ETView{.316}{End}{155}\par\vspace{1pt}
\end{minipage}\par
\vspace{1pt}\noindent\begin{minipage}{\linewidth}
\begin{lstlisting}[style=trajectory,xleftmargin=3pt,xrightmargin=3pt]
c = connect(); r = c.robot
b = c.cameras['scene'].measure_region((475, 523, 530, 591),
\end{lstlisting}
\end{minipage}\par\nointerlineskip
\begin{lstlisting}[style=trajectory,xleftmargin=3pt,xrightmargin=3pt]
    seed=(500, 550))
r.set_motion_scene(contact=b, effector='left')
for i in range(2, 10):
    a = r.actions.move_tcp(Pose((0, -0.17, 0.785), (0, 1, 0, 0)),
        effector='left', position_tolerance_m=0.001)
    p = r.state.effectors['left'].pose.position
    if not (-0.008 < p[0] < 0.008 and -0.18 < p[1] < -0.158 and (
        0.78 < p[2] < 0.8)):
        raise RuntimeError('Unexpected depression position')
    a = r.actions.move_tcp(Pose((0, -0.17, 0.86), (0, 1, 0, 0)),
        effector='left', position_tolerance_m=0.002)
    if a.status != 'success':
\end{lstlisting}
\nointerlineskip\noindent\begin{minipage}{\linewidth}
\begin{lstlisting}[style=trajectory,xleftmargin=3pt,xrightmargin=3pt]
        raise RuntimeError('release failed')
c.cameras['scene'].observe(save_path='snapshots/middle_done9')
\end{lstlisting}
\captionof{figure}{After the first middle-button press, P17 completes presses 2--9. It checks the measured depression pose and requires successful release before the next cycle. Subsequent confirmation and reset complete the native task at 508 steps.}\label{fig:trajectory-robodojo-21}

\end{minipage}
\end{ETCase}

\begin{ETCase}[unbreakable]{(e) InsertTubes}{Failure}
\noindent\begin{minipage}{\linewidth}

\ETHeading{(e) InsertTubes}{Failure}
\ETMeta{\ModelGPTAstra{} $\cdot$ dual ARX-X5 $\cdot$ rollout 1}
\noindent \ETView{.235}{Start}{156}\hfill\ETView{.235}{$1/4$}{157}\hfill\ETView{.235}{$1/2$}{158}\hfill\ETView{.235}{$3/4$}{159}\par\vspace{1pt}
\noindent \ETView{.235}{End}{160}\hspace{.02\linewidth}\ETView{.235}{Left wrist}{161}\hspace{.02\linewidth}\ETView{.235}{Right wrist}{162}\par\vspace{1pt}
\end{minipage}\par
\vspace{1pt}\noindent\begin{minipage}{\linewidth}
\begin{lstlisting}[style=trajectory,xleftmargin=3pt,xrightmargin=3pt]
c = connect(); r = c.robot
cam = c.cameras['scene']; f = cam.frame
\end{lstlisting}
\end{minipage}\par\nointerlineskip
\begin{lstlisting}[style=trajectory,xleftmargin=3pt,xrightmargin=3pt]
p = cam.measure_region((602, 497, 610, 510))
cur = r.state.effectors['right'].pose
for (name, approach) in [('downward', (0, 0, -1)), ('inclined',
    (0.15, 0.65, -0.745))]:
    (pre, grip) = grasp_poses(p.region, cur,
        r.max_opening_m['right'], 0.045, 0.01, None,
        approach=approach, closing_axis=(0.9, -0.4, 0))
\end{lstlisting}
\nointerlineskip\noindent\begin{minipage}{\linewidth}
\begin{lstlisting}[style=trajectory,xleftmargin=3pt,xrightmargin=3pt]
    check = r.check_motion(grip, via=[pre], frames=[f], contact=p,
        opening=1, effector='right')
\end{lstlisting}
\captionof{figure}{After earlier insertion attempts, P26 checks additional grasp candidates but dispatches no further physical action. The agent ends at 477/500 steps with a native score of 40/100; the complete three-tube goal is not satisfied.}\label{fig:trajectory-robodojo-20}

\end{minipage}
\end{ETCase}

\begin{ETCase}[unbreakable]{(f) OrganizeTable}{Failure}
\noindent\begin{minipage}{\linewidth}

\ETHeading{(f) OrganizeTable}{Failure}
\ETMeta{\ModelGPTAstra{} $\cdot$ dual ARX-X5 $\cdot$ rollout 1}
\noindent \ETView{.316}{Start}{163}\hfill\ETView{.316}{$1/4$}{164}\hfill\ETView{.316}{$1/2$}{165}\par\vspace{1pt}
\noindent \ETView{.316}{$3/4$}{166}\hfill\ETView{.316}{End}{167}\hfill\ETView{.316}{Right wrist}{168}\par\vspace{1pt}
\end{minipage}\par
\vspace{1pt}\noindent\begin{minipage}{\linewidth}
\begin{lstlisting}[style=trajectory,xleftmargin=3pt,xrightmargin=3pt]
c = connect(); r = c.robot
cam = c.cameras['scene']
\end{lstlisting}
\end{minipage}\par\nointerlineskip
\begin{lstlisting}[style=trajectory,xleftmargin=3pt,xrightmargin=3pt]
o = cam.segment_region((28, 405, 224, 565), seed=(145, 491),
    save_path='snapshots/fig_body')
r.set_motion_scene(carried=o, effector='left')
q = r.state.effectors['left'].pose.quaternion
for p in [(-0.26, -0.02, 1.0), (-0.166, 0.109, 1.0), (-0.166,
    0.109, 0.905)]:
    a = r.move_to(Pose(p, q), effector='left', stop_on_stall=True)
    if not a.success:
        break
else:
\end{lstlisting}
\nointerlineskip\noindent\begin{minipage}{\linewidth}
\begin{lstlisting}[style=trajectory,xleftmargin=3pt,xrightmargin=3pt]
    r.set_gripper(1, effector='left')
cam.observe(save_path='snapshots/fig_stand')
\end{lstlisting}
\captionof{figure}{P69 binds an observed figurine estimate and attempts guarded transport. The first waypoint fails, so the loop withholds release. Later grasp and placement recoveries exhaust 200 requests at 991/1,000 steps, with the four-object arrangement still incomplete.}\label{fig:trajectory-robodojo-22}

\end{minipage}
\end{ETCase}

\label{app:execution-trajectories-end}
\endgroup

\end{document}